\PassOptionsToPackage{unicode}{hyperref}
\PassOptionsToPackage{hyphens}{url}
\PassOptionsToPackage{dvipsnames,svgnames,x11names}{xcolor}
\documentclass[
  letterpaper,
  9pt]{article}
\usepackage{amsmath,amssymb}
\usepackage{iftex}
\ifPDFTeX
  \usepackage[T1]{fontenc}
  \usepackage[utf8]{inputenc}
  \usepackage{textcomp} 
\else 
  \usepackage{unicode-math} 
  \defaultfontfeatures{Scale=MatchLowercase}
  \defaultfontfeatures[\rmfamily]{Ligatures=TeX,Scale=1}
\fi
\usepackage{lmodern}
\ifPDFTeX\else
\fi
\IfFileExists{upquote.sty}{\usepackage{upquote}}{}
\IfFileExists{microtype.sty}{
  \usepackage[]{microtype}
  \UseMicrotypeSet[protrusion]{basicmath} 
}{}
\makeatletter
\@ifundefined{KOMAClassName}{
  \IfFileExists{parskip.sty}{%
    \usepackage{parskip}
  }{
    \setlength{\parindent}{0pt}
    \setlength{\parskip}{6pt plus 2pt minus 1pt}}
}{
  \KOMAoptions{parskip=half}}
\makeatother
\usepackage{xcolor}
\usepackage[margin=0.72in]{geometry}
\usepackage{longtable,booktabs,array}
\usepackage{calc} 
\usepackage{etoolbox}
\makeatletter
\patchcmd\longtable{\par}{\if@noskipsec\mbox{}\fi\par}{}{}
\makeatother
\IfFileExists{footnotehyper.sty}{\usepackage{footnotehyper}}{\usepackage{footnote}}
\makesavenoteenv{longtable}
\usepackage{graphicx}
\makeatletter
\def\maxwidth{\ifdim\Gin@nat@width>\linewidth\linewidth\else\Gin@nat@width\fi}
\def\maxheight{\ifdim\Gin@nat@height>\textheight\textheight\else\Gin@nat@height\fi}
\makeatother
\setkeys{Gin}{width=\maxwidth,height=\maxheight,keepaspectratio}
\makeatletter
\def\fps@figure{htbp}
\makeatother
\providecommand{\tightlist}{%
  \setlength{\itemsep}{0pt}\setlength{\parskip}{0pt}}
\NewDocumentCommand\citeproctext{}{}

\makeatletter
 \let\@cite@ofmt\@firstofone
 \def\@biblabel#1{}
 \def\@cite#1#2{{#1\if@tempswa , #2\fi}}
\makeatother
\newlength{\cslhangindent}
\newlength{\csllabelwidth}
\newenvironment{CSLReferences}[2] 
 {\begin{list}{}{%
  \setlength{\itemindent}{0pt}
  \setlength{\leftmargin}{0pt}
  \setlength{\parsep}{0pt}
  \ifodd #1
   \setlength{\leftmargin}{\cslhangindent}
   \setlength{\itemindent}{-1\cslhangindent}
  \fi
  \setlength{\itemsep}{#2\baselineskip}}}
 {\end{list}}
\usepackage{calc}

\usepackage{fontspec}
\usepackage{microtype}
\usepackage{booktabs}
\usepackage{longtable}
\usepackage{array}
\usepackage{ragged2e}
\usepackage{enumitem}
\usepackage{fancyhdr}
\usepackage{titlesec}
\usepackage{xcolor}
\usepackage{graphicx}
\usepackage{amsmath,amssymb,mathtools}
\usepackage{caption}
\usepackage{float}
\usepackage{etoolbox}
\usepackage{xurl}
\usepackage{hyperref}

\definecolor{EurekaBlue}{HTML}{183B73}
\definecolor{EurekaGreen}{HTML}{57B649}
\definecolor{EurekaRule}{HTML}{262626}

\hypersetup{
  colorlinks=true,
  linkcolor=EurekaBlue,
  citecolor=EurekaBlue,
  urlcolor=EurekaBlue,
  pdftitle={Eureka: Task-Conditioned Meta-Agent Orchestration for Scientific Discovery},
  pdfauthor={ManXis},
  pdfsubject={Scientific Discovery Meta-Agent Architecture}
}
\setlist{nosep,leftmargin=1.5em}

\titleformat{\section}{\Large\bfseries\sffamily\color{EurekaBlue}}{}{0pt}{}
\titleformat{\subsection}{\large\bfseries\sffamily}{}{0pt}{}
\titleformat{\subsubsection}{\normalsize\bfseries\sffamily}{}{0pt}{}
\titlespacing*{\section}{0pt}{1.7em}{0.6em}
\titlespacing*{\subsection}{0pt}{1.4em}{0.45em}
\titlespacing*{\subsubsection}{0pt}{1.1em}{0.35em}

\fancypagestyle{eurekafirst}{%
  \fancyhf{}%
  \fancyhead[C]{%
    \raisebox{-3.5em}[0pt][0pt]{%
    \fcolorbox{EurekaGreen}{white}{%
      \parbox[c][12pt][c]{0.965\textwidth}{%
        \centering\fontsize{7.5}{8.5}\selectfont\sffamily
        \textbf{Product Architecture Note:} Eureka is a research sub-architecture of \textbf{Tanglang}, developed by ManXis.%
      }%
    }%
    }%
  }%
  \fancyhead[R]{\raisebox{-3.5em}[0pt][0pt]{\fontsize{7}{8}\selectfont\sffamily\thepage}}%
  \fancyfoot{}%
}

\newcommand{\EurekaTitleBlock}{%
  \thispagestyle{eurekafirst}%
  \vspace*{-0.65em}%
  \noindent
  \begin{minipage}[t]{0.78\textwidth}
    \vspace{0pt}
    {\raggedright\fontsize{21.5}{24.5}\selectfont\bfseries\sffamily\color{EurekaBlue}
      Eureka: Task-Conditioned Meta-Agent\\Orchestration for Scientific Discovery\par}
    \vspace{0.45em}
    {\fontsize{11}{12.5}\selectfont\bfseries\sffamily\color{EurekaGreen}ManXis\par}
    \vspace{0.35em}
    {\fontsize{7.7}{9.3}\selectfont\bfseries\sffamily
      Alizer Wong\textsuperscript{1,*}, Heng Cui\textsuperscript{1}, Yi Tan\textsuperscript{2}, Xiongchao Zhan\textsuperscript{3},\\
      Liang Lin\textsuperscript{4}, Yuxiang Guo\textsuperscript{5}, Zhaorong Dai\textsuperscript{6}, Zixin Zeng\textsuperscript{7}, Wenyuan Li\textsuperscript{8}\par}
  \end{minipage}\hfill
  \begin{minipage}[t]{0.18\textwidth}
    \vspace{-0.15em}
    \raggedleft\includegraphics[width=1.16in]{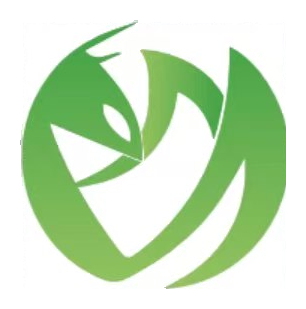}
  \end{minipage}

  \vspace{0.95em}
  {\fontsize{6.9}{8.2}\selectfont\sffamily
  \textsuperscript{1} ManXis; \textsuperscript{2} School of Information Engineering, Guangdong University of Technology;\\
  \textsuperscript{3} School of Automation, Guangdong University of Technology; \textsuperscript{4} School of Artificial Intelligence, South China Normal University;\\
  \textsuperscript{5} Shanghai Jiao Tong University; \textsuperscript{6} Pratt School of Engineering, Duke University;\\
  \textsuperscript{7} School of Computer Science and Technology, Guangdong University of Technology; \textsuperscript{8} Hokkaido University.\par}

  \vspace{0.65em}
  {\fontsize{6.3}{7.5}\selectfont\sffamily
  \textsuperscript{*} Corresponding author: Alizer Wong (contact@manxis.org).\\
  Author contacts: cuiheng2025@gmail.com; tyyeahhhhh@outlook.com; zxc857297353@outlook.com; linliang5618@gmail.com;\\
  yuxiang127@sjtu.edu.cn; zhaorong.dai@duke.edu; zengzixin@mails.gdut.edu.cn; wenyuan@lmd.ist.hokudai.ac.jp.\\
  Homepage: \href{https://manxis-website.netlify.app}{https://manxis-website.netlify.app}\quad Code: \href{https://github.com/manxis-contact/Eureka}{https://github.com/manxis-contact/Eureka}\par}

  \vspace{0.9em}
  \noindent\begin{minipage}[t]{0.49\textwidth}
    {\fontsize{7.8}{9}\selectfont\sffamily Research Report}
  \end{minipage}\hfill
  \begin{minipage}[t]{0.49\textwidth}
    \raggedleft{\fontsize{7.8}{9}\selectfont\sffamily August 2026}
  \end{minipage}
  \vspace{0.55em}
  \par\noindent{\color{EurekaGreen}\rule{\textwidth}{1.15pt}}
  \vspace{1.1em}
}
\ifLuaTeX
  \usepackage{selnolig}  
\fi
\usepackage{bookmark}
\IfFileExists{xurl.sty}{\usepackage{xurl}}{} 
\hypersetup{
  colorlinks=true,
  linkcolor={Maroon},
  filecolor={Maroon},
  citecolor={Blue},
  urlcolor={Blue},
  pdfcreator={LaTeX via pandoc}}

\author{}
\date{}

\begin{document}

\EurekaTitleBlock

\subsection{Abstract}\label{abstract}

Scientific discovery, open mathematical conjectures, and other
long-horizon tasks under substantial uncertainty impose requirements
that are difficult to satisfy with a single fixed agent architecture. A
fixed architecture must simultaneously perform task decomposition, state
maintenance, verification, tool use, and long-term adaptation, which
introduces architecture mismatch and orchestration overhead when task
structures are heterogeneous. We present \textbf{Eureka}, a
task-conditioned Meta-Agent architecture. Eureka compiles a long-horizon
task into a dynamic obligation graph with explicit acceptance semantics
and forms Macro-Agents with specialized state, memory, operators, tools,
verifiers, and local topology during execution through receding-horizon
planning, architecture promotion, and minimal sufficient architecture
compilation. When long-horizon execution exposes recurring bottlenecks,
Eureka further applies cost-benefit-gated governed evolution to update
the local architecture under explicit constraints. Theoretically, we
establish a collection of formal results concerning fixed-architecture
regret, planning invalidation, promotion and evolution amortization,
information-sufficient subtree interfaces, concurrency serializability,
and compositional verification correctness. Experimentally, Eureka
completes 170/170 recursive long-horizon tasks and produces 3,948
acceptance certificates, with no observed uncertified acceptance or
false terminal state. Compiled active context reduces the median
model-input context from 9,490 to 4,005 tokens; incremental dependency
processing avoids 65.38\% of repeated computation across 12,000
dependency-update tasks; and all 16,000 concurrent executions are
consistent with a valid serial execution. More importantly, the same
Eureka Meta-Agent forms a Theory-Discovery Agent and a Math/Conjecture
Agent under two distinct epistemic structures. The former yields
structural results in quantum-process and spacetime theory, including
full-rank conjunction interiorization, null-sector algebraic decoupling,
a global acted-set normal form, behavioural-interface equivalence
separation, and an operational intervention signature. The latter
identifies operator-access and representation bottlenecks in research on
the Riemann Hypothesis and advances a whole-vector positivity
certificate candidate for Suzuki's localized Weil quadratic form to
\(0<a\le 69/200=0.345\), reaching approximately 99.55\% of the
first-prime threshold \((\log 2)/2\). These results indicate that the
capability of a scientific agent depends not only on the underlying
model but also on whether an appropriate agent architecture can be
formed and maintained according to the cognitive structure of the task
itself.

\subsection{Key Findings at a Glance}\label{key-findings-at-a-glance}

Eureka forms two structurally distinct specialized scientific agents
from the same Meta-Agent architecture and produces verifiable progress
in both open-conjecture mathematics and theoretical-structure discovery.

\begin{longtable}[]{@{}
  >{\raggedright\arraybackslash}p{(\columnwidth - 6\tabcolsep) * \real{0.2500}}
  >{\raggedright\arraybackslash}p{(\columnwidth - 6\tabcolsep) * \real{0.2500}}
  >{\raggedright\arraybackslash}p{(\columnwidth - 6\tabcolsep) * \real{0.2500}}
  >{\raggedright\arraybackslash}p{(\columnwidth - 6\tabcolsep) * \real{0.2500}}@{}}
\toprule\noalign{}
\begin{minipage}[b]{\linewidth}\raggedright
Scientific track
\end{minipage} & \begin{minipage}[b]{\linewidth}\raggedright
Specialized agent formed by Eureka
\end{minipage} & \begin{minipage}[b]{\linewidth}\raggedright
Main advances
\end{minipage} & \begin{minipage}[b]{\linewidth}\raggedright
Current evidence status
\end{minipage} \\
\midrule\noalign{}
\endhead
\bottomrule\noalign{}
\endlastfoot
Riemann Hypothesis & \textbf{Math/Conjecture Agent} & operator-access
obstruction; finite/local Chebyshev cone separation; finite-cluster
interpolation; localized Weil positivity certificate candidate
\(\lambda_a>0\) for \(0<a\le69/200\) & analytic derivation + 1,010-cell
outward interval certificate candidate; not an RH proof \\
New Theoretical Structures & \textbf{Theory-Discovery Agent} & full-rank
conjunction interiorization; algebraic null-sector decoupling; global
acted-set normal form; behavioural-interface equivalence separation;
operational intervention signature & exact/internal certificates and
large-scale regression; some results still require external or formal
review \\
\end{longtable}

The current quantitative endpoint on the Riemann-Hypothesis track is

\[
\lambda_a>0,\qquad 0<a\le\frac{69}{200}=0.345,
\]

which expands the range of the same localized-Weil certificate family
relative to \(a\le1/4\) by \textbf{1.38\(\times\)} and reaches
approximately \textbf{99.55\%} of the first-prime structural threshold
\((\log2)/2\approx0.34657359028\). The result is not equivalent to a
proof of the Riemann Hypothesis and does not constitute a new record for
the proportion of zeros on the critical line.

For theoretical discovery, the full-rank two-setting/two-outcome QSOST
gluing candidate has explicit parameter \(t=1/2\), parent minimum
eigenvalue \(1/8\), single-setting domination cost \(4\), and joint dual
lower bound \(65/16>4\). Higher-level structural results further
distinguish closed behavioural equivalence from black-box interface
equivalence and use an operational intervention signature to fix
primitive, query, and resource semantics.

At the system level, Eureka completes \textbf{170/170} recursive
long-horizon tasks and produces \textbf{3,948} acceptance certificates.
Governed Evolution achieves both the lowest median total cost,
\textbf{2525.4}, and the highest success rate, \textbf{60.55\%}, among
four evaluated evolution policies. Compiled active context reduces the
median model-input context from \textbf{9,490} to \textbf{4,005} while
preserving the success rate. Across 12,000 incremental dependency tasks,
Eureka avoids \textbf{65.38\%} of repeated computation. All 16,000
concurrent-execution tasks are consistent with a valid serial execution,
with \textbf{0} unsafe commits.

\section{1 Introduction}\label{introduction}

Large-scale pretrained language models have progressively evolved from
task-specific systems for individual natural-language-processing
problems into general computational substrates capable of performing
multiple classes of cognitive tasks through a unified natural-language
interface. Earlier pretraining-and-fine-tuning paradigms typically
required task-specific data and parameter updates, whereas continued
scaling of model size, data size, and training compute substantially
changed this operating regime. Kaplan et al. (2020) observed stable
power-law relationships between language-model loss and model scale,
dataset scale, and training compute, indicating that scaling can yield
predictable performance improvements over a broad computational range.
Building on this development, Brown et al. (2020) showed that
sufficiently large language models can transfer to translation, question
answering, text generation, and selected reasoning tasks using only
natural-language task descriptions and a small number of in-context
examples, without task-specific gradient updates. Subsequent work on
instruction following further strengthened responses to open-ended
natural-language instructions (Ouyang et al. 2022). The key consequence
of these developments is not merely an increase in individual benchmark
scores; the mechanism of task adaptation has gradually shifted from
retraining a model for each task toward conditioning the behaviour of a
general cognitive substrate through context and external control
structures.

A unified task interface, however, does not imply that a complex problem
can be reduced to a single conditional generation. As large language
models have been applied to mathematical reasoning, program generation,
complex information retrieval, interactive decision making, and
scientific research, the limitations of a static input-output invocation
have become increasingly apparent. Complex tasks usually contain
multiple mutually dependent intermediate states, and correctness depends
on the ability to establish, preserve, and revise those states rather
than merely to generate locally plausible text. Wei et al. (2022) showed
that explicitly generating intermediate reasoning steps can
substantially improve performance on arithmetic, commonsense, and
symbolic reasoning tasks, demonstrating that the organization of the
computation trajectory is itself a significant determinant of model
capability. The capability boundary of a large language model therefore
extends beyond the amount of knowledge encoded in its parameters to the
ability to construct an intermediate computational process appropriate
for the current problem, providing the foundation for the subsequent
agent paradigm.

Internal reasoning trajectories alone remain insufficient for the
external information acquisition and environmental operations required
by real-world tasks. Information needed by many problems is absent from
the current context and may also be absent from the model parameters,
while certain operations cannot be executed reliably through language
generation alone, including real-time retrieval, exact computation, code
execution, database access, and modification of external environment
state. Schick et al. (2023) demonstrated that language models can learn
when to call external APIs, which API to select, how to construct
arguments, and how to exploit tool outputs. Yao et al. (2023) further
placed reasoning and acting in a single closed loop, enabling a language
model to generate an action from the current state, receive a new
observation from the environment, and update subsequent reasoning and
action plans accordingly. Systems of this form are no longer adequately
described as isolated text-generation models; they are more naturally
viewed as closed-loop cognitive and decision processes driven by
language models.

This transition further motivated LLM-based agents. The essential
distinction between an agent and a standalone language model is not the
use of a particular prompt, but the fact that task-solving capability is
jointly determined by model reasoning, external tools, environmental
observations, persistent state, memory, feedback, and control flow. For
example, Shinn et al. (2023) converted task outcomes into verbal
reflections stored in episodic memory, allowing subsequent trials to use
prior experience without changing the parameters of the base model.
Recent generalist multi-agent systems further demonstrate that
orchestration structure outside the model is a first-order variable in
complex task behaviour. Fourney et al. (2024) used a central
Orchestrator for planning, progress tracking, and replanning after
failure, while specialized agents provide browsing, file manipulation,
and code execution. As language models are transformed from static
generators into persistent task-execution entities, system performance
can no longer be explained by base-model capability alone. The same
underlying model can exhibit substantially different behaviour under
different state representations, tool configurations, memory mechanisms,
verification procedures, and control flows. Agent research consequently
expands from model capability to the joint design of the model and its
external cognitive architecture.

The importance of external cognitive architecture is amplified in
long-horizon complex tasks. Software engineering, complex investigation,
scientific research, and open mathematical problems commonly involve
tens or hundreds of mutually dependent operations, and the definition of
future steps may change continuously as intermediate results become
available. Such tasks combine hierarchical objectives, partially unknown
future steps, cross-stage state dependencies, and varying degrees of
parallelism. An intermediate result may determine not only whether a
later node remains valid but also how the remainder of the task should
be decomposed. A complete action sequence generated at the beginning of
the task therefore cannot be assumed to remain valid after new
information is acquired. At the opposite extreme, a purely iterative
process that generates one action, obtains an observation, and recalls
the model to select the next action incurs repeated context
transmission, frequent replanning, and serial execution overhead. The
central problem of long-horizon agents consequently shifts from
predicting the correct next action to constructing and maintaining a
task-computation structure that can evolve as new information arrives.

To address task-dependent complexity, Prasad, Koller, et al. (2023)
introduced as-needed decomposition, recursively decomposing a subtask
only when the current executor is unable to solve it and showing that
decomposition depth should adapt jointly to task complexity and executor
capability. In parallel, Kim et al. (2023) organized complex function
calling as an explicit dependency graph, with a planner constructing
task relationships, a task-fetching unit dispatching ready tasks whose
dependencies have been satisfied, and independent tasks executing in
parallel. These results indicate that long-horizon execution
increasingly resembles a compiler and task-scheduling system: the
language model is principally responsible for semantic decomposition and
reasoning that have not yet been determined, whereas dependency
resolution, readiness checks, parallel scheduling, and state maintenance
can be assigned to deterministic runtime components.

Recursive decomposition into ever smaller nodes nevertheless does not by
itself resolve persistent state and local autonomy. In many long-horizon
tasks, a group of adjacent subtasks is not a collection of independent
atomic calls but instead shares the same internal state, tool set,
verification mechanism, and local decision policy for an extended
period. A local problem may repeatedly access the same facts, invoke the
same reasoning operators, and update shared state through a common
verifier. Assigning every node to an isolated generic executor
repeatedly incurs state reload, context reconstruction, and
cross-executor coordination costs. Conversely, keeping all task state
inside a single agent causes unbounded context growth and constrains
parallel processing of independent work. Recent work has directly
quantified the resulting coordination cost. G. Zhang, Yue, et al. (2025)
represented LLM-based multi-agent collaboration as a spatiotemporal
message-passing graph and substantially reduced token consumption by
pruning redundant communication edges, demonstrating that unsuitable
agent boundaries and communication topology translate directly into
measurable inference cost. Task decomposition therefore raises a
higher-order question: which regions of the task graph should remain
collections of independent operations, and which regions should be
encapsulated as specialized agents with persistent state and local
autonomy?

Research on multi-agent systems and automated agent design has begun to
address this question. Y. Wang et al. (2024) dynamically decomposed
complex tasks according to execution requirements and generated
specialized subagents for individual subtasks, showing that agents can
be dynamically instantiated computational objects rather than execution
entities fixed before planning begins. Khattab et al. (2023) elevated
complex language-model pipelines from manually concatenated prompt
strings to compilable graphs of declarative modules and optimized those
modules using a compiler. In a related direction, Zhuge et al. (2024)
represented language agents as optimizable graphs of operation nodes and
information-flow edges, optimizing both node-level prompts and graph
connectivity. Jiayi Zhang et al. (2025) represented agentic workflows as
a code-level search space and used Monte Carlo Tree Search, execution
feedback, and tree-structured experience to iteratively modify the
workflow. Hu, Lu, et al. (2024) went further by formulating
agentic-system design itself as an automated search problem and using
Meta Agent Search to generate new agent programs composed of prompts,
tool use, control flow, and combinations of agentic building blocks.
Collectively, these studies show that the cognitive architecture above
the model can itself become an object of planning and optimization
rather than a fixed peripheral implementation.

Dynamic task decomposition and automated agent design, however, do not
become unified merely because each is individually feasible.
Architecture-search methods typically assume a relatively well-defined
evaluation task and search among complete candidate agents, whereas
dynamic task planning primarily concerns how a given complex task should
be decomposed and assigned. Recent work has begun to condition
architectures more directly on the input. G. Zhang, Niu, et al. (2025)
learned an agentic supernet containing multiple candidate structures and
sampled different multi-agent architectures and inference-resource
allocations for individual queries. Yue, Zhang, et al. (2025) used
cascaded controllers to determine collaboration mode, role allocation,
and model routing, allowing several dimensions of a multi-agent system
to vary with the input. For open, long-horizon tasks whose future
definition changes with intermediate results, the more fundamental
question concerns the coupling between dynamic orchestration and
architecture formation: when should an agent architecture be created,
which structural information in the task trajectory should determine the
location of the new boundary, and under what conditions is continued
generic execution preferable to synthesizing a new specialized agent?
Different task regions can exhibit fundamentally different persistence,
dependency density, verification semantics, parallelism, and future
planning depth. Imposing a fixed state representation and control
topology across such heterogeneous cognitive structures creates
structural overhead that cannot necessarily be removed by increasing the
scale of the base model.

Scientific discovery provides an especially stringent instance of this
problem because scientific research is not a single task category with a
uniform input-output structure. A complete research process may involve
literature retrieval, problem formalization, hypothesis generation,
theoretical derivation, counterexample search, experimental design,
evidence integration, exclusion of prior explanations, and final
validation, with substantially different dependency structures and
evidence requirements across research questions. Recent systems have
begun to demonstrate the potential of agents for automated science.
Gottweis et al. (2026) organized scientific hypothesis generation as a
multi-agent search process consisting of generation, reflection,
ranking, and evolution, while integrating search and specialized tools
to improve hypothesis quality and grounding. Yamada et al. (2025)
organized hypotheses, experiments, and analysis using progressive
agentic tree search while reducing dependence on manually authored code
templates. Schmidgall et al. (2025) connected literature review,
experimentation, and report writing in an end-to-end research workflow.
In another direction, Novikov et al. (2025) represented candidate
solutions as executable programs and combined automated evaluators with
evolutionary search to discover new algorithms and verifiable
mathematical or computational results. Together, these systems show that
agents are moving beyond information summarization toward automated
discovery systems that continually form, modify, and test candidate
scientific structures.

The epistemic structures of scientific tasks nevertheless differ
substantially. Open theoretical research typically cannot enumerate the
complete candidate space in advance; the system must maintain competing
explanations, unconfirmed assumptions, supporting and contradictory
evidence, and experiments or theoretical criteria that distinguish
alternative theories. Progress is not equivalent to executing more
steps, but to shrinking the feasible explanation space, strengthening
verifiable mechanisms, and eliminating alternatives. Rigorous
mathematics has a different structure: propositions, lemmas,
assumptions, and proof obligations form exact dependencies; intermediate
results must be reused reliably; and a local error can invalidate an
entire downstream proof chain. The AlphaProof/AlphaGeometry 2 line of
work emphasizes formal feedback and machine-verifiable proof state in
rigorous mathematical reasoning (Hubert et al. 2026). Ren et al. (2025)
further used a recursive theorem-proving pipeline to decompose difficult
theorems into subgoals and reorganize local proofs into long-horizon
formal-reasoning trajectories, highlighting the distinctive
decomposition-verification-composition structure required by
mathematical agents. Even with the same underlying language model,
hypothesis-evidence state for open theory discovery and fact-claim-proof
state for rigorous mathematics therefore obey fundamentally different
organizational principles. Task variation changes not only prompt
content but also which states must be maintained, which search operators
are admissible, what information must persist, and what evidence
suffices to certify an intermediate result.

These observations motivate a more fundamental question: \textbf{can a
fixed agent architecture efficiently cover long-horizon tasks with
heterogeneous epistemic structures simply by changing the task
instruction?} Evidence from automated agent design already suggests that
the strongest version of this assumption is untenable. Hu, Lu, et al.
(2024) places prompt, tool use, control flow, and combinations of
agentic building blocks in a common search space and shows that
automatically generated architectures can outperform multiple manually
designed systems. Prasad, Koller, et al. (2023) and Kim et al. (2023)
further show that decomposition depth and execution dependency structure
depend on task complexity and runtime state rather than being fully
specified by fixed control rules. Theoretical discovery and rigorous
mathematics provide a sharper example: the former benefits from
maintaining competing branches and independent falsification, whereas
the latter often benefits from continuity of proof state and exact
dependency tracking. A single fixed topology must therefore pay
avoidable coordination or representation costs on at least some such
tasks unless one architecture happens to be simultaneously optimal
across all relevant task structures.

We analyze this phenomenon from the perspective of joint
task-architecture optimization and identify a structural
\textbf{architecture mismatch}. Effective progress depends not only on
the reasoning capability of the base model but also on whether the task
state is represented in a sufficient and compact form, whether
subproblems are assigned appropriate autonomy boundaries, whether
operations receive suitable tools and verifiers, and whether strongly
dependent reasoning trajectories are unnecessarily split across isolated
sessions. Requirements can conflict across tasks. An architecture that
increases search branching for open theory discovery may increase
synchronization and state-replication cost in formal proof search; an
architecture optimized for uninterrupted proof-state continuity may
unnecessarily suppress parallel hypothesis exploration. This mismatch
cannot in general be removed by simply increasing context length or
model scale, because the resulting overhead may arise from repeated
state transfer, unnecessary serialization, missing verification
semantics, or inappropriate communication boundaries rather than
insufficient model capability.

A second difficulty arises even after architecture adaptation is
allowed: in a real long-horizon task, information required to determine
the appropriate local architecture may not exist at the beginning of
execution. The next phase of an open research program can depend on
whether a counterexample exists; subsequent decomposition of a proof can
depend on whether a critical lemma holds; an experimental result can
invalidate an entire downstream research branch. Fully expanding the
task graph at the outset therefore requires the planner to predict
unobserved intermediate outcomes. Incorrect predictions invalidate
downstream obligations, tool plans, and potentially already-created
agent state. The problem differs from uncertainty in a fixed planning
state space because execution may redefine the decomposition itself.

We formalize this phenomenon as \textbf{planning invalidation}. If the
correct form of a future subtask depends on upstream information that
has not yet been observed, constructing the subtask early does not
reduce the uncertainty of the actual scientific problem; instead, the
system spends computation on one branch of several possible futures.
When new observations invalidate that branch, the associated planning
tokens, context organization, dependency construction, and architecture
state become wasted computation. The probability of invalidation can
accumulate with planning depth because more unobserved intermediate
outcomes must be predicted correctly. Consequently, a more complete
initial plan is not necessarily a more efficient long-horizon plan,
particularly in scientific discovery where the task graph itself is
revealed through execution.

These observations motivate a planning regime between two extremes:
complete upfront planning and one-step-at-a-time replanning. Eureka
therefore organizes planning as a receding-horizon obligation process.
The planner expands only the portion of the task graph that current
information can determine reliably; unresolved future work remains
represented as deferred obligations with explicit information
boundaries. Once dependencies and acceptance conditions are known, ready
tasks can execute immediately. Semantic planning is paused while the
executor still possesses sufficient ready work and is reactivated when
the ready frontier is depleted, when the task state changes
semantically, or when a structural dependency emerges. Planning is
thereby governed by execution backpressure rather than by a fixed depth
schedule.

Based on these considerations, we propose \textbf{Eureka}, a
task-conditioned Meta-Agent architecture for open long-horizon tasks.
Eureka dynamically compiles the task into a recursive obligation
structure and generates, runs, and evolves task-specific specialized
agents during execution. Rather than maintaining a predefined library of
expert agents or immediately selecting a fixed template, Eureka first
compiles goals, constraints, and acceptance semantics into a dynamic
obligation graph. Receding-horizon planning expands only the local
structure that can be determined from current information, while
ready-frontier backpressure prevents persistent token expenditure on
distant nodes that do not yet have execution value. As execution
generates new state transitions, Eureka identifies architecture hotspots
from persistent state sharing, dependency density, operator recurrence,
verifier recurrence, and long-term planning demand, and performs
architecture promotion only when local autonomy is expected to reduce
long-term coordination and state-reconstruction cost.

Promoted regions receive more than a task-specific prompt. Eureka
compiles them into Macro-Agents with task-specific state
representations, operators, memory, tool bindings, verifiers, and
session topology. Internal complexity is encapsulated within the local
subtree; the parent orchestrator observes only verified exported
artifacts, explicit assumptions, unresolved debts, and reopen conditions
through a typed subtree interface. As local execution continues,
telemetry determines whether a recurring architectural bottleneck has a
sufficiently long remaining horizon to justify adaptation. Governed
evolution then modifies the cognitive structure only when the expected
benefit can amortize diagnosis, evaluation, and migration cost. Figure 1
provides an overview of the complete Eureka workflow, from task
initialization and dynamic obligation orchestration to specialized-agent
formation, governed evolution, long-horizon execution, and
certificate-driven scientific outputs.

\begin{figure}
\centering
\includegraphics[width=1\textwidth,height=\textheight]{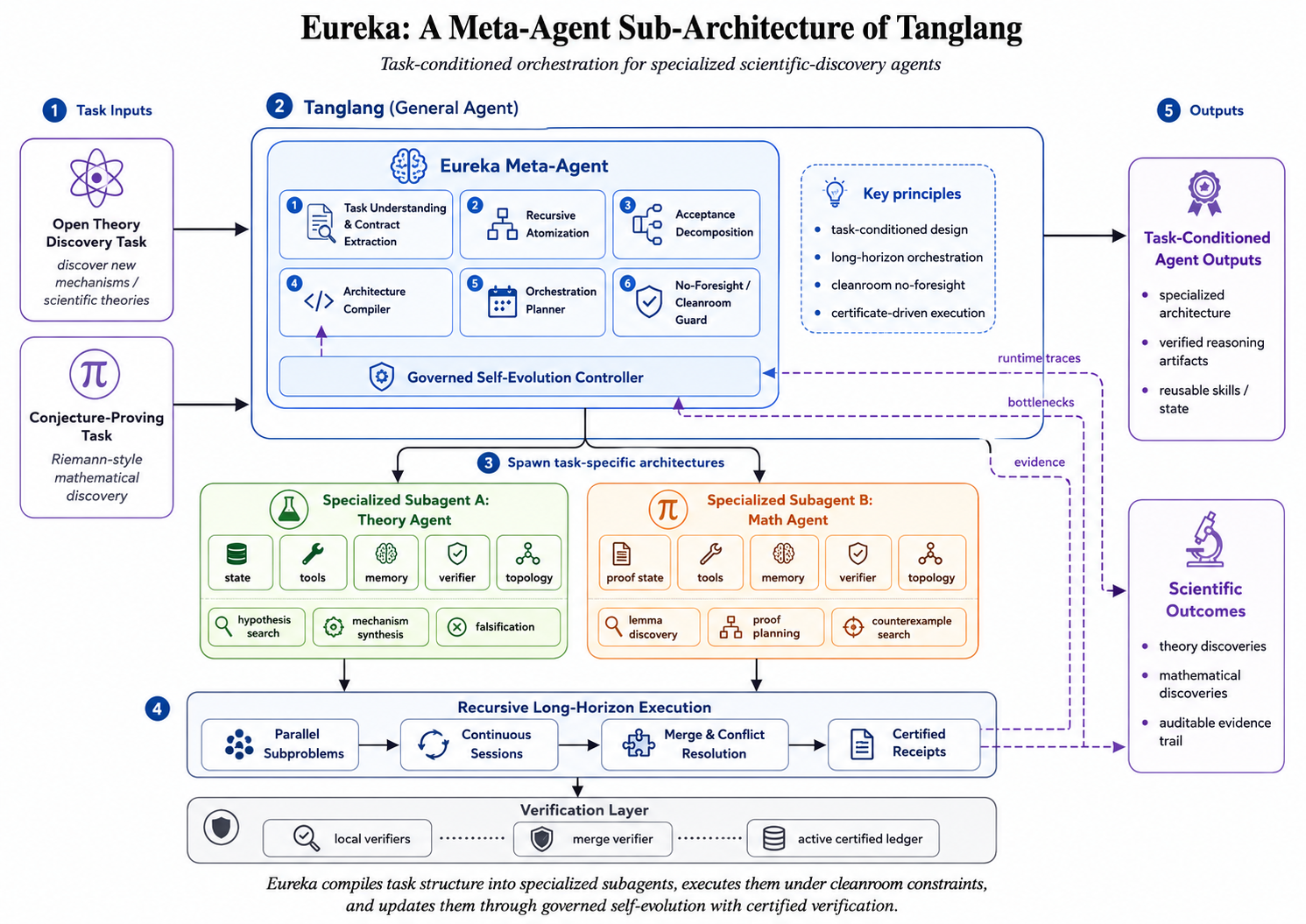}
\caption{Overview of Eureka. The Meta-Agent compiles a task into a
dynamic obligation graph, forms task-conditioned specialized agents when
architecture hotspots emerge, governs local evolution, and coordinates
recursive long-horizon execution under typed
verification.}\label{fig:eureka-framework}
\end{figure}

The central research question consequently contains two sequential and
separately testable stages. The first concerns \textbf{architecture
discovery}: can Eureka use structural information revealed by execution
to determine where a specialized agent should form and compile an
architecture matched to the local cognitive structure? The second
concerns \textbf{scientific discovery}: can the specialized agents
formed and governed by Eureka subsequently produce valuable new
theories, mathematical results, or other verifiable scientific
discoveries? This decomposition elevates scientific problem solving into
a more general systems question. A general agent must not only execute a
predefined cognitive workflow; it must also form the computational
organization that the current scientific problem requires and then
sustain knowledge discovery within that organization.

\section{2 Related Work}\label{related-work}

\subsection{Dynamic Task Orchestration and Automated Agent Architecture
Design}\label{dynamic-task-orchestration-and-automated-agent-architecture-design}

As LLM-based agents have expanded from single-round tool calls to
long-horizon tasks composed of many mutually dependent operations,
planning has evolved from selecting the next action into a joint problem
of decomposition granularity, execution dependency, and
computational-resource allocation. Kim et al. (2023) organized function
calling as a compiled execution framework with a planner, task-fetching
unit, and executor, allowing ready tasks to be dispatched as soon as
their dependencies are satisfied and independent operations to execute
in parallel. Y. Wang et al. (2024) combined dynamic task decomposition
with agent generation and produced corresponding subagents according to
execution-time task requirements. These results jointly show that
orchestration efficiency depends not only on local reasoning quality but
also on task-graph structure, decomposition granularity, and scheduling
policy.

Dynamic decomposition further exposes the limitations of fixed-role
multi-agent systems. Traditional frameworks commonly predefine planners,
researchers, critics, and executors before execution begins. When the
actual task state does not align with those boundaries, increasing the
number of agents can amplify context replication and inter-agent
communication. W. Chen et al. (2025) explored a more open organizational
regime through heterogeneous-agent integration, dynamic teaming, and
conversation-flow control. G. Zhang, Yue, et al. (2025) showed from the
perspective of communication graphs that many message edges in existing
multi-agent pipelines are removable, reinforcing that collaboration
topology itself is a structural variable for token efficiency.

A parallel research direction directly treats agent architecture as an
optimizable computational object. Zhuge et al. (2024) represents
language agents as recursively composable graphs and optimizes both
node-level prompts and graph connectivity. Hu, Lu, et al. (2024)
formulates agent-system design as automated search and uses Meta Agent
Search to generate prompts, tool-use patterns, and control flows in code
space. Shang et al. (2024) abstracts agents into standardized modules
for planning, reasoning, tool use, and memory, then searches for
improved architectures by module recombination and evolution. Jiayi
Zhang et al. (2025) represents workflows as code-level graphs and uses
Monte Carlo Tree Search, execution feedback, and tree-structured
experience to modify them. Agent scaffolds are consequently evolving
from manually fixed peripherals into searchable, recomposable
structures.

Recent work additionally considers input- or query-conditioned
architecture formation. G. Zhang, Niu, et al. (2025) learns an agentic
supernet and samples different multi-agent architectures and
inference-resource allocations for individual queries. Yue, Zhang, et
al. (2025) unifies collaboration mode, role allocation, and LLM routing
through cascaded controllers so that several structural dimensions of a
multi-agent system can vary with the input. Yaolun Zhang, Liu, and Xiao
(2025) automatically constructs finite-state-machine-controlled
multi-agent systems from task descriptions and further optimizes the
generated structure. These studies move automated agent design from
cross-task search for a single static architecture toward
task-conditioned system construction.

Eureka differs primarily in how task decomposition,
architecture-boundary discovery, and task-conditioned agent synthesis
are unified within the same online long-horizon execution trajectory.
Eureka neither searches for one complete specialized agent before
execution nor creates a new independent executor for every newly
generated subtask. The system recursively expands the obligation
structure that current information can determine, then uses state
sharing, dependency density, operator/verifier recurrence, and remaining
horizon observed in the actual execution trace to determine whether a
subtree has become an architecture hotspot. Agent synthesis is invoked
as a higher-order planning action only when persistent local autonomy
has positive amortized value, promoting the region into a Macro-Agent
with specialized state, memory, operators, verifiers, and internal
topology.

\subsection{Self-Improving and Self-Evolving Agent
Systems}\label{self-improving-and-self-evolving-agent-systems}

As agent performance increasingly depends on scaffold structure, memory,
tool interfaces, and control flow outside the language model, research
has shifted from single-trajectory optimization in a fixed agent toward
systems that modify their own computational structure using execution
experience. Robeyns et al. (2025) allows a coding agent to edit its own
implementation and selects changes using benchmark feedback. Jiaming
Zhang et al. (2025) combines code-level self-modification with
open-ended evolutionary search, maintaining an archive of self-modified
agents from which additional architectural variants can be generated.
These studies show that agent implementation can itself become a
continuing source of capability improvement even when the base model is
unchanged.

Recent work extends architecture evolution to complete harnesses and
persistent execution trajectories. Lee et al. (2026) treats harness code
governing what is stored, retrieved, and presented as an outer-loop
search object and allows an agentic proposer to access source code,
scores, and execution traces of prior candidates. Pan et al. (2026)
re-executes difficult tasks from historical trajectories and uses
self-validation, self-consistency, and pairwise self-preference to
produce harness updates. H. Zhang et al. (2026) uses a three-stage
process of Weakness Mining, Harness Proposal, and Proposal Validation to
convert observed failure modes into minimal, regression-tested harness
modifications. These studies collectively move self-improving agents
from local prompt refinement toward execution-trace-driven harness
optimization.

Most self-improving systems principally address how to generate better
agent variants. In long-horizon tasks, however, whether an evolution
event should be initiated at all, and which layer of the architecture
should be modified, are equally important determinants of total
efficiency. Persistent architecture search incurs diagnosis, candidate
generation, evaluation, and state-migration cost. When the remaining
horizon is short, even a mutation that improves future per-step
performance may fail to amortize its optimization overhead. Eureka
therefore models self-evolution as a Meta-Agent-governed planning
action: an evolution budget is allocated only when a bottleneck is
sufficiently recurrent, the remaining horizon is sufficiently long, and
expected verifiable gains exceed the cost of adaptation.

Eureka further classifies telemetry-derived bottlenecks into runtime,
prompt/operator, memory/skill, tool-interface, state/verifier, topology,
and model-capability levels, and searches first for the lowest-level
modification sufficient to explain and resolve the observed failure.
Local low-risk mutations can execute within a bounded EvolutionLease,
whereas structural changes affecting state semantics, verifier
contracts, or agent boundaries are escalated to the Meta-Agent.
Candidate evaluation progresses from inexpensive static checks and
micro-replays to more expensive shadow evaluation, so architecture
evolution is governed by the same cost-benefit discipline as
long-horizon execution rather than becoming an unconditional loop inside
every subagent.

\subsection{Agentic AI for Scientific and Mathematical
Discovery}\label{agentic-ai-for-scientific-and-mathematical-discovery}

Agentic AI is expanding from local research assistance, such as
literature retrieval, code generation, and experimental analysis, toward
complete scientific-discovery workflows. Gottweis et al. (2026)
organizes scientific hypothesis research through generation, reflection,
ranking, and evolution, using multi-agent search to generate, criticize,
and refine hypotheses. Schmidgall et al. (2025) advances a research idea
through literature review, experimentation, and report writing. Yamada
et al. (2025) reduces the dependence of the first AI Scientist system on
manually authored code templates and manages multiple experimental and
research branches through progressive agentic tree search. Collectively,
these systems demonstrate that scientific agents are becoming automated
research systems spanning multiple successive stages rather than
isolated research-assistance tools.

Different scientific tasks induce different cognitive architectures
because their search spaces and verification mechanisms differ. Novikov
et al. (2025) represents candidate objects as executable programs and
constructs a high-throughput generate-evaluate loop through evolutionary
search and automated evaluators. Wang and Luan (2026) explicitly argues
that scientific workflows in different disciplines have different
control-flow structures and supports multiple research paradigms through
a lightweight DAG kernel, editable workflows, full-text literature
indexing, and cross-run knowledge accumulation. These systems show that
workflow state structure and verification semantics are central design
variables in automated discovery rather than incidental implementation
details.

Rigorous mathematical discovery further sharpens these structural
differences. The AlphaProof/AlphaGeometry 2 line of work uses formal
environments to provide machine-checkable feedback, allowing proof
search to be organized around exact verification (Hubert et al. 2026).
Ren et al. (2025) recursively decomposes difficult theorems into
subgoals and reorganizes local proofs into formal reasoning
trajectories. Relative to open hypothesis search, mathematical agents
therefore require more stringent fact/claim/proof dependencies,
persistent proof state, and exact verification.

Long-term adaptation is also beginning to appear in AI Scientist
systems. Lyu et al. (2026) uses Researcher, Engineer, and Evolution
Manager agents and persistent ideation and experimentation memories to
extract reusable strategies from successful and failed prior research
trajectories. Such results show that scientific agents can improve
research behaviour through experience accumulated across iterations,
while the overall agent roles and research pipeline remain specified by
the system design. Eureka addresses a higher-level problem: whether the
system can determine from the emerging epistemic structure of execution
where a specialized scientific agent should form, which state,
operators, verifiers, and local topology the agent should maintain, and
how the resulting architecture should continue adapting during
discovery.

The evaluation object of Eureka therefore differs in level from that of
most existing AI-for-Science systems. Existing systems primarily test
what hypotheses, experiments, algorithms, or proofs can be produced once
a specialized scientific agent architecture has been provided. Eureka
jointly evaluates architecture discovery and scientific discovery:
whether the Meta-Agent can form a suitable cognitive architecture from
task requirements and runtime trajectories, and whether the specialized
agent produced by that architecture can subsequently generate
independently valuable and verifiable scientific results.

\section{3 Theoretical Analysis: From Dynamic Task Information to
Verifiable Recursive
Orchestration}\label{theoretical-analysis-from-dynamic-task-information-to-verifiable-recursive-orchestration}

This section provides the formal derivations for the first four
theoretical components of Eureka. To avoid converting engineering
intuition into mathematical claims, we state results only under
explicit, testable conditions; when a claim does not hold in full
generality, we also characterize the failure mode and state a corrected
result that is provable under reasonable assumptions. A unified notation
is used throughout: random variables are denoted by uppercase Latin
letters, individual obligations by lowercase \(o\), agent architectures
by calligraphic \(\mathcal A\), task instances by \(T\), and the task
distribution by \(\mu\). Unless stated otherwise, all sets are equipped
with their Borel \(\sigma\)-algebras, and all random variables and
policies are assumed measurable.

To eliminate ambiguity in the term \emph{efficiency}, Sections 3.1-3.4
use \textbf{expected total cost under a fixed reliability constraint} as
the sole optimization objective, rather than arbitrarily combining
tokens, success rate, and verification strength into one scalar score.
Verified progress per unit cost may be reported later as an auxiliary
experimental metric, but it does not enter the main theorems in this
section. The principal symbols are summarized below.

\begin{longtable}[]{@{}
  >{\raggedright\arraybackslash}p{(\columnwidth - 2\tabcolsep) * \real{0.5000}}
  >{\raggedright\arraybackslash}p{(\columnwidth - 2\tabcolsep) * \real{0.5000}}@{}}
\toprule\noalign{}
\begin{minipage}[b]{\linewidth}\raggedright
Symbol
\end{minipage} & \begin{minipage}[b]{\linewidth}\raggedright
Definition
\end{minipage} \\
\midrule\noalign{}
\endhead
\bottomrule\noalign{}
\endlastfoot
\(\Omega,\mathscr F,\mathbb P\) & underlying probability space for task
stochasticity \\
\(T\) & a task instance; \(\mu\) denotes the task distribution \\
\(X_t\) & latent task state at time \(t\), valued in a standard Borel
space \(\mathcal X\) \\
\(O_t\) & observation available before action \(A_t\), valued in
\(\mathcal O_{\mathrm{obs}}\) \\
\(A_t\) & orchestration/execution control action, valued in
\(\mathcal A_{\mathrm{ctrl}}\) \\
\(C_0\) & initial Task Contract, Acceptance Contract, budget,
permissions, and corpus cutoff \\
\(\mathcal F_t\) & filtration of information legally available at time
\(t\) \\
\(\tau\) & root-task termination time, required to be a stopping time \\
\(\kappa_t,K_\tau\) & step cost and cumulative total cost \\
\(S_\tau\) & indicator that the root Acceptance Contract is satisfied \\
\(\alpha\) & admissible failure probability; strict proof tasks may set
\(\alpha=0\) \\
\(\mathcal A\) & agent architecture; \(\mathfrak A\) is the candidate
architecture space \\
\(\Pi(\mathcal A)\) & admissible policies implementable by architecture
\(\mathcal A\) \\
\(G_t=(V_t,E_t)\) & dynamic obligation graph at time \(t\) \\
\(o\) & a single obligation; its goal, dependencies, read/write sets,
and acceptance condition are defined when introduced \\
\end{longtable}

\subsection{3.1 Dynamic Task Processes and Admissible
Information}\label{dynamic-task-processes-and-admissible-information}

\subsubsection{3.1.1 Probability Space, Task State, and Legally
Available
Information}\label{probability-space-task-state-and-legally-available-information}

We represent a discrete-time long-horizon task on the probability space
\((\Omega,\mathscr F,\mathbb P)\). The latent task state at time
\(t\in\mathbb N_0\) is \(X_t\in\mathcal X\); the information observed
before the \(t\)-th control action is
\(O_t\in\mathcal O_{\mathrm{obs}}\); and the control action is
\(A_t\in\mathcal A_{\mathrm{ctrl}}\). The initial Task Contract,
resource budget, tool permissions, public-corpus cutoff, and root
Acceptance Contract are collected in \(C_0\) and treated as information
determined at time zero. The filtration available before action \(A_t\)
is defined by

\[
\boxed{
\mathcal F_t
:=
\sigma\!\left(
C_0,
O_0,\ldots,O_t,
A_0,\ldots,A_{t-1}
\right),
\qquad t\in\mathbb N_0 .
}
\tag{1}
\]

The temporal ordering in (1) is substantive: \(A_t\) may depend on
\(O_t\), but not on an as-yet-unobserved \(O_{t+1}\) or any posteriorly
constructed discovery signal. Eureka's orchestration policy is
\(\pi=(\pi_t)_{t\ge0}\), where \(\pi_t(\cdot\mid\mathcal F_t)\) is a
stochastic kernel over \(\mathcal A_{\mathrm{ctrl}}\). Every admissible
action is therefore \(\mathcal F_t\)-adapted. The task termination time
is denoted by \(\tau\) and is required to be a stopping time with
respect to \((\mathcal F_t)_{t\ge0}\), so that the statement that the
task has completed cannot depend on future information that has not yet
been observed.

Execution cost is not collapsed to a token count. Instead, a nonnegative
step cost \(\kappa_t\) is used; after a fixed normalization,
\(\kappa_t\) may include token use, tool calls, concurrency
coordination, replanning, and architecture migration. The cumulative
cost is

\[
\boxed{
K_\tau
:=
\sum_{t=0}^{\tau-1}\kappa_t,
\qquad
\kappa_t\ge0,
\qquad
\mathbb E[K_\tau]<\infty .
}
\tag{2}
\]

Whether the root task satisfies its Acceptance Contract is represented
by \(S_\tau\in\{0,1\}\), which is required to be
\(\mathcal F_\tau\)-measurable. Given an admissible failure probability
\(\alpha\in[0,1)\), we first fix a reliability requirement and then
minimize cost rather than defining an arbitrarily weighted scalar
progress score. For task \(T\) and architecture \(\mathcal A\), let
\(\Pi(\mathcal A)\) denote all admissible policies implementable by that
architecture. The optimal expected cost under the reliability constraint
is

\[
\boxed{
\mathcal C_\alpha(\mathcal A;T)
:=
\inf_{\pi\in\Pi(\mathcal A)}
\left\{
\mathbb E_T^{\pi}[K_\tau]
\;:\;
\mathbb P_T^{\pi}(S_\tau=1)\ge1-\alpha
\right\}.
}
\tag{3}
\]

If the constraint set is empty, we define
\(\mathcal C_\alpha(\mathcal A;T)=+\infty\). Equation (3) explicitly
separates lower cost from sacrificing correctness: efficiency is
comparable only among candidate architectures that satisfy the same
Acceptance Contract. Formal proof tasks that require strict correctness
may set \(\alpha=0\).

\subsubsection{3.1.2 Obligation DAGs and Agent
Architectures}\label{obligation-dags-and-agent-architectures}

At any time, Eureka does not store the full task as a natural-language
plan. Instead, it maintains a dynamic obligation graph
\(G_t=(V_t,E_t)\). Each node \(o\in V_t\) carries five kinds of
information: goal semantics \(g_o\), known prerequisite dependencies
\(d_o\), persistent-state read set \(r_o\), state write set \(w_o\), and
Acceptance Contract \(\nu_o\). Edges in \(E_t\subseteq V_t\times V_t\)
represent semantic or data dependencies that must be satisfied first.
When the current graph is acyclic, the ready frontier is the set of
unfinished obligations whose predecessors have all been accepted. If
later evidence reveals a genuine dependency cycle, Section 3.4 specifies
the corresponding boundary treatment.

A local agent architecture is denoted by
\(\mathcal A=(\mathcal S,\mathcal M,\mathcal U,\mathcal V,\mathcal T,\mathcal P)\),
where \(\mathcal S\) is the persistent state representation,
\(\mathcal M\) the memory policy, \(\mathcal U\) the callable operator
family, \(\mathcal V\) the verifier family, \(\mathcal T\) the
tool/interface set, and \(\mathcal P\) the session and execution
topology. The definition deliberately excludes the simplification that
an agent is equivalent to a prompt, because the architecture-regret
results below require state, verification, and topology to vary
independently.

\subsubsection{Assumption 1 (Auditable
Acceptance)}\label{assumption-1-auditable-acceptance}

For every obligation \(o\) marked \texttt{DONE}, the runtime must retain
a replayable receipt from which the acceptance event under \(\nu_o\) can
be reconstructed from persistent state. A natural-language conclusion
without an acceptance receipt cannot be promoted to certified state.

\subsubsection{Assumption 2 (Finite Control
Cost)}\label{assumption-2-finite-control-cost}

Every individual planning, tool-execution, agent-synthesis,
verification, and architecture-migration action has finite conditional
expected cost. In addition, among policies that satisfy the root
Acceptance Contract, at least one policy must make (2) finite; otherwise
the task is regarded as infeasible under the current system resources.

\subsubsection{Assumption 3 (Sufficient Recording of Persistent
State)}\label{assumption-3-sufficient-recording-of-persistent-state}

All durable information that can affect a future policy decision or
Acceptance Contract must enter typed state or be losslessly recoverable
from immutable receipt references. The assumption does not require the
Meta-Agent to reread the complete history at every step; it requires
only that omitted information can be recovered without loss when needed.

\subsubsection{Proposition 1 (Structural Renaming
Invariance)}\label{proposition-1-structural-renaming-invariance}

Suppose tasks \(T\) and \(T'\) admit a graph isomorphism
\(\varphi:V_t\to V'_t\) preserving dependency edges, Acceptance
Contracts, state read/write relations, resource prices, and available
operator/tool contracts, and that the two tasks differ only in entity
names and natural-language surface form. If Eureka's decomposition,
promotion, and architecture-synthesis policies depend only on these
structural quantities and on information in \(\mathcal F_t\) associated
with the corresponding structural equivalence classes, then, under a
fixed random seed, the two control trajectories are isomorphic after
mapping by \(\varphi\).

\textbf{Proof.} At \(t=0\), the structural states are isomorphic by
assumption. Suppose the trajectories remain isomorphic through time
\(t\). The structured ControlCapsules received by the Meta-Agent are
then identical after mapping by \(\varphi\). Because the policy does not
read task names, the conditional action distributions agree.
Deterministic runtime transitions preserve graph isomorphism, and random
semantic actions produce corresponding outputs under a common fixed
seed. The states at time \(t+1\) therefore remain isomorphic. Induction
over time yields the result. \(\square\)

The proposition is not an automatic property of arbitrary
natural-language agents; it defines an architecture-invariance condition
that can be tested directly through task anonymization and
structural-consistency experiments.

\begin{center}\rule{0.5\linewidth}{0.5pt}\end{center}

\subsection{3.2 Structural Regret Lower Bound for Fixed Agent
Architectures}\label{structural-regret-lower-bound-for-fixed-agent-architectures}

\subsubsection{3.2.1 Formal Definition of Architecture
Regret}\label{formal-definition-of-architecture-regret}

Let \(\mathfrak A\) be the set of agent architectures under
consideration. For a task \(T\) at reliability threshold \(1-\alpha\),
define the globally optimal cost and the architecture regret of a fixed
architecture by

\[
\boxed{
\mathcal C_\alpha^\star(T)
:=
\inf_{\mathcal A\in\mathfrak A}
\mathcal C_\alpha(\mathcal A;T),
\qquad
\mathcal R_\alpha(\mathcal A;T)
:=
\mathcal C_\alpha(\mathcal A;T)-\mathcal C_\alpha^\star(T).
}
\tag{4}
\]

Because (3) already constrains the success probability,
\(\mathcal R_\alpha\) measures additional cost due to architecture
mismatch under the same reliability standard rather than misclassifying
a cheaper but less reliable system as more efficient. For any
\(\varepsilon\ge0\), define the \(\varepsilon\)-near-optimal
architecture set by
\(\mathfrak A_T(\varepsilon)=\{\mathcal A\in\mathfrak A:\mathcal R_\alpha(\mathcal A;T)\le\varepsilon\}\).

Disjoint exact minimizer sets alone do not imply a uniform positive
regret lower bound for fixed architectures. In a continuous architecture
space, there may exist a sequence of architectures that simultaneously
approaches the optimal value on both tasks arbitrarily closely without
attaining either optimum. The main result below therefore uses disjoint
near-optimal sets as the sufficient and testable structural condition.

\subsubsection{Lemma 1 (Mutually Exclusive Near-Optimal Sets Imply
Strict Loss on at Least One
Task)}\label{lemma-1-mutually-exclusive-near-optimal-sets-imply-strict-loss-on-at-least-one-task}

Let \(T_1,T_2\) be two tasks and let \(\varepsilon_1,\varepsilon_2>0\).
Suppose

\[
\boxed{
\mathfrak A_{T_1}(\varepsilon_1)
\cap
\mathfrak A_{T_2}(\varepsilon_2)
=
\varnothing .
}
\tag{5}
\]

Then, for every fixed architecture \(\mathcal A\in\mathfrak A\), at
least one index \(i\in\{1,2\}\) satisfies
\(\mathcal R_\alpha(\mathcal A;T_i)>\varepsilon_i\).

\textbf{Proof.} If \(\mathcal A\notin\mathfrak A_{T_1}(\varepsilon_1)\),
the definition of the near-optimal set gives
\(\mathcal R_\alpha(\mathcal A;T_1)>\varepsilon_1\). If
\(\mathcal A\in\mathfrak A_{T_1}(\varepsilon_1)\), (5) implies
\(\mathcal A\notin\mathfrak A_{T_2}(\varepsilon_2)\), hence
\(\mathcal R_\alpha(\mathcal A;T_2)>\varepsilon_2\). The two cases cover
every \(\mathcal A\). \(\square\)

\subsubsection{Theorem 1 (Fixed-Architecture Regret Lower
Bound)}\label{theorem-1-fixed-architecture-regret-lower-bound}

Let the task random variable \(T\) equal \(T_1\) with probability
\(p\in(0,1)\) and \(T_2\) with probability \(1-p\). If (5) holds, then
every fixed architecture \(\mathcal A\) satisfies

\[
\boxed{
\mathbb E\big[\mathcal R_\alpha(\mathcal A;T)\big]
=
p\,\mathcal R_\alpha(\mathcal A;T_1)
+(1-p)\,\mathcal R_\alpha(\mathcal A;T_2)
\ge
\min\!\left\{p\varepsilon_1,(1-p)\varepsilon_2\right\}
>0 .
}
\tag{6}
\]

\textbf{Proof.} By Lemma 1, every \(\mathcal A\) falls into one of two
cases. If \(\mathcal R_\alpha(\mathcal A;T_1)>\varepsilon_1\),
nonnegativity of regret gives

\[
\mathbb E[\mathcal R_\alpha(\mathcal A;T)]
\ge p\varepsilon_1.
\]

If the first case does not hold, Lemma 1 guarantees
\(\mathcal R_\alpha(\mathcal A;T_2)>\varepsilon_2\), so

\[
\mathbb E[\mathcal R_\alpha(\mathcal A;T)]
\ge (1-p)\varepsilon_2.
\]

Taking a common lower bound over the two cases yields (6). \(\square\)

Theorem 1 does not state unconditionally that every task requires a
distinct agent. It establishes a precise distinction: only when the
near-optimal architecture regions of different tasks are structurally
separated must every fixed architecture incur a strictly positive
average excess cost. Empirical work treating agent scaffolds as
optimization variables, including GPTSwarm (Zhuge et al. 2024), ADAS
(Hu, Lu, et al. 2024), and AgentSquare (Shang et al. 2024), motivates
this structural perspective; Equation (6), however, is derived
independently under the objective in (3).

\subsubsection{Corollary 1 (Task-Conditioned Architectures Can Remove
the Lower Bound When Structure Is
Identifiable)}\label{corollary-1-task-conditioned-architectures-can-remove-the-lower-bound-when-structure-is-identifiable}

Assume further that an \(\mathcal F_0\)-measurable structural variable
\(Z\) is observed before architecture selection and that \(Z=z_i\)
identifies \(T=T_i\) without error. If, for each task, there exists
\(\mathcal A_i\in\mathfrak A_{T_i}(\delta_i)\) with \(\delta_i\ge0\),
then the structure-conditioned policy \(g(Z)=\mathcal A_i\) satisfies

\[
\boxed{
\mathbb E\big[\mathcal R_\alpha(g(Z);T)\big]
\le
p\delta_1+(1-p)\delta_2 .
}
\tag{7}
\]

If the optimal values are attained for both tasks and
\(\delta_1=\delta_2=0\), (7) reduces to zero architecture regret. The
corollary explains why Eureka generates architectures from task
structure; it does not imply that an arbitrary task-conditioned selector
is necessarily optimal.

\subsubsection{Proposition 2 (Boundary Case in Which a Fixed
Architecture Is
Sufficient)}\label{proposition-2-boundary-case-in-which-a-fixed-architecture-is-sufficient}

If an architecture \(\mathcal A_0\in\mathfrak A\) satisfies
\(\mathcal C_\alpha(\mathcal A_0;T)=\mathcal C_\alpha^\star(T)\) for
every task \(T\) in the support of the task distribution, then the
expected architecture regret of \(\mathcal A_0\) is zero. Consequently,
on a task family admitting a universal optimal architecture, the
mutually exclusive near-optimal-set condition of Theorem 1 must fail,
and Eureka should not force promotion merely because task names differ.

\subsubsection{Proposition 3 (Scaling the Base Model Does Not
Automatically Eliminate Independent Architecture
Overhead)}\label{proposition-3-scaling-the-base-model-does-not-automatically-eliminate-independent-architecture-overhead}

Let \(m\in\mathcal M\) denote a backbone-model configuration and suppose
the reliability-constrained cost decomposes as

\[
\boxed{
\mathcal C_\alpha(\mathcal A;T,m)
=
B_\alpha(T,m)
+
H_\alpha(\mathcal A,T),
}
\tag{8}
\]

where \(B_\alpha\) is semantic-computation cost determined only by the
task and base model, and \(H_\alpha\) is architecture-specific overhead
from state replication, coordination, missing verification, or execution
topology and is invariant to \(m\). If the near-optimal sets induced by
\(H_\alpha\) for two tasks satisfy (5), then the lower bound of Theorem
1 holds for every \(m\in\mathcal M\).

\textbf{Proof.} For fixed \(T\), \(B_\alpha(T,m)\) does not depend on
the architecture and therefore cancels from
\(\mathcal C_\alpha(\mathcal A;T,m)-\inf_{\mathcal A'}\mathcal C_\alpha(\mathcal A';T,m)\).
Architecture regret is consequently determined entirely by differences
in \(H_\alpha\), so Lemma 1 and Theorem 1 apply unchanged. \(\square\)

Equation (8) is an explicit separability assumption rather than a
general fact. If a stronger model reduces state-recovery cost or changes
which verifiers are available, \(H_\alpha\) may itself depend on \(m\);
Proposition 3 then no longer applies and the dependence
\(H_\alpha(\mathcal A,T,m)\) must be measured directly.

\begin{center}\rule{0.5\linewidth}{0.5pt}\end{center}

\subsection{3.3 Endogenous Task Revelation and Planning
Invalidation}\label{endogenous-task-revelation-and-planning-invalidation}

\subsubsection{3.3.1 Marginal Value of Early
Planning}\label{marginal-value-of-early-planning}

Future obligations in long-horizon tasks often depend on upstream
observations that have not yet been resolved. To characterize whether
early planning is worthwhile, fix time \(t\), condition on
\(\mathcal F_t\), and consider one candidate future planning unit \(j\).
Let \(Z_j\in\{0,1\}\) indicate whether the unit constructed now remains
valid after the upstream information determining its semantics is
revealed, and define \(p_j:=\mathbb P(Z_j=1\mid\mathcal F_t)\).
Constructing the unit early incurs a conditionally determined cost
\(c_j^{\mathrm E}>0\). If planning is deferred until the upstream
information is revealed and the unit is constructed only when \(Z_j=1\),
the cost is \(c_j^{\mathrm D}\ge0\). If valid early planning reduces
subsequent latency, context reload, or coordination, let the
corresponding saving in the same cost units be \(b_j\ge0\). The main
result first adopts the conservative assumptions that an invalid early
plan has no reusable residual value and that early planning does not
change the environment state or the distribution of \(Z_j\).

Under these conditions, the conditional expected net value of early
planning relative to deferring until the required information arrives is
exactly

\[
\boxed{
\Delta_j
:=
\mathbb E\!\left[
C_j^{\mathrm{defer}}-C_j^{\mathrm{early}}
\mid\mathcal F_t
\right]
=
p_j\big(c_j^{\mathrm D}+b_j\big)-c_j^{\mathrm E}.
}
\tag{9}
\]

There is no hidden term in (9). The defer policy pays
\(c_j^{\mathrm D}\) only when \(Z_j=1\), so its expected planning cost
is \(p_jc_j^{\mathrm D}\). The early policy always pays
\(c_j^{\mathrm E}\), but gains the downstream saving \(b_j\) when
\(Z_j=1\). Their difference is therefore (9). The resulting single-node
criterion is not the informal rule that more distant work should never
be planned; it is an explicit trade-off among conditional survival
probability, early planning cost, and the parallelism or latency benefit
that early planning can provide.

\subsubsection{Lemma 2 (Optimal Early-Planning Decision for a Single
Planning
Unit)}\label{lemma-2-optimal-early-planning-decision-for-a-single-planning-unit}

Under the assumptions of (9), if planning unit \(j\) is separable from
other future units in both state and cost, then early planning weakly
dominates deferred planning if and only if

\[
\boxed{
\Delta_j\ge0
\quad\Longleftrightarrow\quad
p_j
\ge
\frac{c_j^{\mathrm E}}{c_j^{\mathrm D}+b_j}.
}
\tag{10}
\]

The ratio on the right is used only when \(c_j^{\mathrm D}+b_j>0\). If
the denominator is zero, \(c_j^{\mathrm E}>0\) makes early planning
strictly worse.

\textbf{Proof.} Every downstream cost common to the two decisions
cancels, so minimizing conditional expected cost is equivalent to
maximizing (9). The early policy is therefore optimal exactly when
\(\Delta_j\ge0\). Rearranging (9) gives (10). \(\square\)

\subsubsection{Theorem 2 (Optimal Receding Horizon Under Monotone
Marginal
Value)}\label{theorem-2-optimal-receding-horizon-under-monotone-marginal-value}

At time \(t\), consider finite planning units \(j=1,\ldots,H\), ordered
by forecast depth. Assume:

\begin{enumerate}
\def\labelenumi{\arabic{enumi}.}
\tightlist
\item
  conditioned on \(\mathcal F_t\), early/deferred choices are additive
  in expected cost, so early planning of one unit does not alter
  \(p_j,c_j^{\mathrm E},c_j^{\mathrm D},b_j\) for another unit;
\item
  the marginal value of early planning for each unit is given by (9);
  and
\item
  the sequence \(\Delta_1,\ldots,\Delta_H\) is nonincreasing with
  forecast depth.
\end{enumerate}

Define

\[
\boxed{
 h_t^\star
 :=
 \max\!\left(
 \{0\}
 \cup
 \left\{h\in\{1,\ldots,H\}:\Delta_h\ge0\right\}
 \right).
}
\tag{11}
\]

Among all policies choosing early or deferred construction independently
for each planning unit, the conditional expected cost is minimized by
constructing units at depths \(1,\ldots,h_t^\star\) early and retaining
all units deeper than \(h_t^\star\) as deferred obligations. Under
assumptions 1-3, the optimal early-planning set is therefore a prefix
and \(h_t^\star+1\) is an information boundary in the strict sense.

\textbf{Proof.} By assumption 1, the expected cost change of any choice
vector \(e=(e_1,\ldots,e_H)\in\{0,1\}^H\), relative to deferring every
unit, decomposes as

\[
\boxed{
\mathbb E[C(e)-C(0)\mid\mathcal F_t]
=
-\sum_{j=1}^{H}e_j\Delta_j .
}
\tag{12}
\]

Each \(e_j\) can therefore be optimized independently: choose \(e_j=1\)
if \(\Delta_j>0\), choose \(e_j=0\) if \(\Delta_j<0\), and either value
when \(\Delta_j=0\). Assumption 3 implies that all nonnegative
\(\Delta_j\) form a contiguous prefix beginning at depth 1, yielding
(11). \(\square\)

Theorem 2 does not claim that one scalar horizon is optimal for every
long-horizon planning system. Strong coupling among future nodes or
nonmonotone \(\Delta_j\) can destroy the prefix structure. Eureka must
then fall back to per-obligation marginal-value decisions rather than
enforcing one planning depth. The general receding-horizon principle is
closely related to model predictive control, where a finite-horizon
problem is repeatedly solved from the current state and only the
currently actionable portion is executed (Mayne et al. 2000). In LLM
agents, ADaPT provides empirical support for as-needed recursive
decomposition (Prasad, Koller, et al. 2023), while LLMCompiler shows
that ready tasks with resolved dependencies can execute in parallel
before the full future plan is known (Kim et al. 2023).

\subsubsection{Corollary 2 (Survival-Probability Threshold Under Equal
Costs)}\label{corollary-2-survival-probability-threshold-under-equal-costs}

Suppose \(c_j^{\mathrm E}=c_j^{\mathrm D}=c>0\) at every depth and the
benefit of valid early planning is a constant \(b\ge0\). Then the
necessary and sufficient condition for planning unit \(j\) early becomes

\[
\boxed{
 p_j
 \ge
 \frac{c}{c+b},
 \qquad
 h_t^\star
 =
 \max\left\{j:p_j\ge\frac{c}{c+b}\right\},
}
\tag{13}
\]

provided \(p_j\) is nonincreasing with depth. Equation (13) makes the
trade-off explicit. When early planning yields almost no parallelism or
latency benefit, \(b\to0\), only future units that are almost certain to
remain valid should be expanded early. When early planning hides
substantial downstream latency, the required survival probability
decreases accordingly.

\subsubsection{Proposition 4 (Modified Threshold with Reusable Residual
Value)}\label{proposition-4-modified-threshold-with-reusable-residual-value}

Suppose an invalid early plan still yields reusable value
\(s_j\in[0,c_j^{\mathrm E}]\), independent of other planning units.
Equation (9) becomes

\[
\boxed{
\Delta_j^{(s)}
=
p_j(c_j^{\mathrm D}+b_j)
+(1-p_j)s_j
-c_j^{\mathrm E}.
}
\tag{14}
\]

When \(c_j^{\mathrm D}+b_j>s_j\), early planning is optimal if and only
if

\[
\boxed{
 p_j
 \ge
 \frac{c_j^{\mathrm E}-s_j}
 {c_j^{\mathrm D}+b_j-s_j}.
}
\tag{15}
\]

Reusable abstract planning skeletons can therefore legitimately extend
the region in which early planning is useful. Eureka should not treat
every post-observation invalidation as 100\% wasted work.

\subsubsection{Proposition 5 (Threshold Structure of Ready-Frontier
Backpressure)}\label{proposition-5-threshold-structure-of-ready-frontier-backpressure}

Let \(q\in\mathbb N_0\) be the current number of ready obligations, let
one planner batch cost \(c_{\mathrm P}>0\), and suppose that the batch
creates an average of \(m\ge1\) additional ready obligations. Let
\(L(q)\) denote the conditional expected loss from executor starvation
during the next control period, and assume that \(L\) is nonincreasing
and convex. Define the starvation loss avoided by one planner batch as
\(B(q):=L(q)-L(q+m)\). Convexity makes \(B(q)\) nonincreasing in \(q\).
If a planner batch does not change any other cost, the optimal planning
decision has a threshold form: there exists
\(q^\star\in\mathbb N_0\cup\{-1,\infty\}\) such that

\[
\boxed{
 q\le q^\star
 \;\Longrightarrow\;
 B(q)\ge c_{\mathrm P}
 \;\Longrightarrow\;
 \text{activate planner},
\qquad
 q>q^\star
 \;\Longrightarrow\;
 \text{planner sleeps}.
}
\tag{16}
\]

\textbf{Proof.} For convex \(L\), the discrete increment \(L(q)-L(q+m)\)
is nonincreasing in \(q\). Hence \(\{q:B(q)\ge c_{\mathrm P}\}\) is
either empty or an interval beginning at zero, and its largest element
defines \(q^\star\). \(\square\)

\subsubsection{Proposition 6 (State Equivalence and Communication
Complexity of
PlanDelta)}\label{proposition-6-state-equivalence-and-communication-complexity-of-plandelta}

Let \(G_t\) encode the persistent state of the complete obligation
graph. Suppose a planner wake-up changes only \(m_t\) graph records,
while the full graph contains \(n_t\) records. If the runtime patch
operator \(\oplus\) is deterministic and \(\Delta_t\) contains every
insertion, deletion, and state update, then

\[
\boxed{
G_{t+1}
=
G_t\oplus\Delta_t
\equiv
\widetilde G_{t+1},
\qquad
\mathrm{size}(\Delta_t)=\Theta(m_t),
\qquad
\mathrm{size}(\widetilde G_{t+1})=\Theta(n_t),
}
\tag{17}
\]

where \(\widetilde G_{t+1}\) denotes the state obtained if the planner
emits the entire new graph. Whenever \(m_t=o(n_t)\), transmitting the
delta has strictly smaller asymptotic communication complexity than
rewriting the full plan. The proposition concerns only state encoding
and does not change the scientific action policy.

\begin{center}\rule{0.5\linewidth}{0.5pt}\end{center}

\subsection{3.4 Verifiable Recursive
Atomization}\label{verifiable-recursive-atomization}

\subsubsection{3.4.1 Obligation Semantics, Certificates, and Local
Decomposition}\label{obligation-semantics-certificates-and-local-decomposition}

To prove that recursive decomposition does not mistakenly equate the
completion of many subtasks with completion of the root task, we
explicitly distinguish the input instance, candidate artifact,
certificate, and true semantics of every obligation \(o\). Let
\(\mathcal I_o\) be the input space, \(\mathcal Y_o\) the
candidate-artifact space, and \(\mathcal C_o\) the certificate space.
Semantic correctness is represented by
\(\Phi_o:\mathcal I_o\times\mathcal Y_o\to\{0,1\}\), and the
machine-executable verifier is
\(V_o:\mathcal I_o\times\mathcal Y_o\times\mathcal C_o\to\{0,1\}\).
Verifier soundness means that for every \((i,y,c)\), \(V_o(i,y,c)=1\)
implies \(\Phi_o(i,y)=1\). Completeness is not required: a semantically
correct artifact for which no acceptable certificate has yet been found
may remain \texttt{INCONCLUSIVE}.

When a parent obligation \(o\) is decomposed into a finite collection of
child obligations \(o_1,\ldots,o_m\), the children may have a directed
acyclic dependency structure. Fix a topological order compatible with
the local dependency graph. The instance for child \(j\) is produced by
a measurable input constructor \(\psi_j\) from parent input \(i\) and
artifacts of predecessor children. After every child is accepted, the
parent artifact is produced by a composition operator \(\Gamma_o\),
while a merge verifier \(M_o\) checks cross-child consistency, interface
constraints, and the additional conditions required for semantic
composition.

\subsubsection{Assumption 4 (Leaf-Verifier
Soundness)}\label{assumption-4-leaf-verifier-soundness}

Every verifier for an atomic leaf obligation is sound: acceptance cannot
promote a semantically incorrect artifact to certified state.

\subsubsection{Assumption 5 (Soundness of Local Composition
Rules)}\label{assumption-5-soundness-of-local-composition-rules}

For every parent obligation \(o\) and every legal decomposition, if the
semantic predicates of all child obligations hold and the merge verifier
accepts, then the artifact generated by the composition operator must
satisfy the semantics of the parent. Formally, for every legal \(i\),
every topologically compatible collection \(y_1,\ldots,y_m\), and every
merge certificate \(c_M\),

\[
\boxed{
\left[
\bigwedge_{j=1}^{m}
\Phi_{o_j}\!\left(
\psi_j(i,y_{\mathrm{pred}(j)}),y_j
\right)
\right]
\land
\left[M_o(i,y_{1:m},c_M)=1\right]
\;\Longrightarrow\;
\Phi_o\!\left(i,\Gamma_o(i,y_{1:m})\right)=1 .
}
\tag{18}
\]

Equation (18) is the central condition for recursive atomization and
also one of the easiest conditions to omit in an engineering
implementation. Completion of every child is insufficient to establish
completion of the parent. Cross-child consistency, interface
compatibility, shared assumptions, and merge semantics must be covered
explicitly by \(M_o\) or by an equivalent verifiable composition rule.

\subsubsection{Lemma 3 (One-Level Semantic
Composition)}\label{lemma-3-one-level-semantic-composition}

Under Assumption 5, suppose every direct child artifact
\(y_1,\ldots,y_m\) of a parent obligation \(o\) satisfies its semantic
predicate and there exists a merge certificate \(c_M\) such that
\(M_o(i,y_{1:m},c_M)=1\). Then the composite artifact
\(y=\Gamma_o(i,y_{1:m})\) satisfies \(\Phi_o(i,y)=1\).

\textbf{Proof.} The statement is exactly Equation (18) instantiated at
the current \(i,y_{1:m},c_M\). Child semantics may be established either
by atomic verifiers or by deeper recursive composition, so Lemma 3 does
not require every internal node to be separately re-proved by an
additional child verifier. \(\square\)

\subsubsection{Theorem 3 (Recursive Decomposition
Soundness)}\label{theorem-3-recursive-decomposition-soundness}

Suppose the root obligation \(o_{\mathrm{root}}\) is recursively
decomposed a finite number of times into a finite DAG \(G=(V,E)\). Every
non-leaf node satisfies Assumption 5; every leaf verifier satisfies
Assumption 4 and has accepted its artifact; and every internal merge
verifier has accepted. Then the root artifact obtained by applying the
composition operators bottom-up satisfies

\[
\boxed{
\Phi_{o_{\mathrm{root}}}
\left(i_{\mathrm{root}},y_{\mathrm{root}}\right)
=1 .
}
\tag{19}
\]

\textbf{Proof.} Because \(G\) is a finite DAG, it admits a topological
order and a finite rank function \(r:V\to\mathbb N_0\): leaves have rank
zero, and every internal node has rank equal to one plus the maximum
rank of its direct children. We induct on the rank.

Base case: if \(r(o)=0\), then \(o\) is a leaf. Assumption 4 and
verifier acceptance imply \(\Phi_o=1\).

Inductive step: assume the semantics hold for all nodes with rank at
most \(k\). Let \(o\) have rank \(k+1\). Every direct child of \(o\) has
rank at most \(k\), so the induction hypothesis establishes all child
semantics. The merge verifier for \(o\) has accepted, hence Lemma 3
gives \(\Phi_o=1\). Finite induction yields the root result (19).
\(\square\)

Theorem 3 establishes semantic soundness of decomposition; it does not
guarantee that any particular verifier is sufficiently complete. A
genuinely correct leaf that cannot be certified may therefore keep the
root task unresolved, but cannot be accepted incorrectly. For
mathematical proof and scientific discovery, allowing incompleteness
while forbidding unsound acceptance is safer than forcing every
unresolved state into a binary decision.

\subsubsection{Corollary 3 (Acceptance Contracts and Execution Recursion
Can Share One
DAG)}\label{corollary-3-acceptance-contracts-and-execution-recursion-can-share-one-dag}

If every node stores its local \(\Phi_o/V_o\) semantics and every
decomposition stores a merge rule satisfying (18), a separate verifier
tree of the same scale as the work DAG is unnecessary. Parent acceptance
can be composed bottom-up along the same obligation DAG. Independent
verifier sessions can still audit the root, architecture-promotion
boundaries, cross-contract merges, or final acceptance, but such a
second tree is not a data-structure requirement of Theorem 3.

\subsubsection{Proposition 7 (Sufficient Condition for Termination of
Recursive
Atomization)}\label{proposition-7-sufficient-condition-for-termination-of-recursive-atomization}

Suppose there exists a complexity-rank function \(\rho:V\to\mathbb N_0\)
such that every \texttt{DECOMPOSE(o)} creates finitely many children and
every child \(o'\) satisfies \(\rho(o')<\rho(o)\). Then recursive
atomization from any finite-rank root terminates at finite depth and
produces a finite decomposition tree.

\textbf{Proof.} Along any root-to-leaf path, \(\rho\) forms a strictly
decreasing sequence of nonnegative integers, so the path length is at
most \(\rho(o_{\mathrm{root}})+1\). Finite branching together with
finite depth implies a finite total number of nodes. \(\square\)

The proposition is sufficient rather than necessary. A practical system
may use a more general well-founded order. Without any strictly
decreasing structural quantity, however, the instruction to continue
decomposing until an atom is reached does not itself guarantee
termination; an additional budget boundary or semantic stopping rule is
then required.

\subsubsection{Proposition 8 (Boundary Treatment for Cyclic
Dependencies)}\label{proposition-8-boundary-treatment-for-cyclic-dependencies}

If the current obligation graph \(G=(V,E)\) contains a directed cycle,
the DAG induction in Theorem 3 cannot be applied directly. Let
\(\mathrm{SCC}(G)\) be the partition into strongly connected components
and contract every SCC into one macro-obligation to obtain the
condensation graph \(G^\dagger\). The condensation graph of any finite
directed graph is a DAG, so Theorem 3 applies to \(G^\dagger\) provided
every nontrivial SCC has an independent sound acceptance rule. If an SCC
contains only mutually circular unverified claims and no additional
fixed-point semantics, invariant, or joint verifier, local mutual
support within the cycle is insufficient to mark the SCC as accepted.

\subsubsection{Proposition 9 (Decomposition Is Not Always
Beneficial)}\label{proposition-9-decomposition-is-not-always-beneficial}

Suppose an obligation \(o\) can be completed directly by a continuous
session with expected cost \(C_{\mathrm{dir}}\). A sound decomposition
incurs child-execution cost \(C_{\mathrm{sub}}\), context-switching cost
\(C_{\mathrm{ctx}}\), coordination cost \(C_{\mathrm{coord}}\), and
merge/verification cost \(C_{\mathrm{merge}}\). If

\[
\boxed{
C_{\mathrm{sub}}
+C_{\mathrm{ctx}}
+C_{\mathrm{coord}}
+C_{\mathrm{merge}}
\ge
C_{\mathrm{dir}},
}
\tag{20}
\]

and decomposition does not improve the success-probability constraint or
verifier strength, then the decomposition is no better than direct
execution under the objective in (3). Eureka's recursive atomization
must therefore retain a \texttt{DIRECT} branch. The paper does not
assume that complex tasks should be decomposed as finely as possible,
and Theorem 3 does not imply such a rule.

\begin{center}\rule{0.5\linewidth}{0.5pt}\end{center}

The notation introduced above is retained in the remaining theoretical
sections. Unless stated otherwise, the probability space remains
\((\Omega,\mathscr F,\mathbb P)\), the admissible information filtration
remains \((\mathcal F_t)_{t\ge0}\), the dynamic obligation graph remains
\(G_t=(V_t,E_t)\), the agent architecture remains
\(\mathcal A=(\mathcal S,\mathcal M,\mathcal U,\mathcal V,\mathcal T,\mathcal P)\),
and the optimal expected cost under the reliability constraint remains
\(\mathcal C_\alpha(\mathcal A;T)\). All preceding Acceptance-Contract,
verifier-soundness, and recursive-composition assumptions remain in
force.

To avoid introducing undefined quantities later, we use the following
additional notation.

\begin{longtable}[]{@{}
  >{\raggedright\arraybackslash}p{(\columnwidth - 2\tabcolsep) * \real{0.5000}}
  >{\raggedright\arraybackslash}p{(\columnwidth - 2\tabcolsep) * \real{0.5000}}@{}}
\toprule\noalign{}
\begin{minipage}[b]{\linewidth}\raggedright
Symbol
\end{minipage} & \begin{minipage}[b]{\linewidth}\raggedright
Definition
\end{minipage} \\
\midrule\noalign{}
\endhead
\bottomrule\noalign{}
\endlastfoot
\(S\subseteq V_t\) & connected local subgraph of the current obligation
graph, or a subgraph of its condensation DAG, under analysis \\
\(\tau_S\) & local stopping time at which subtree \(S\) is completed or
closed \\
\(N_S\) & number of future local service events generated by \(S\) from
the current time until \(\tau_S\) \\
\(F_S\) & one-time synthesis, state-migration, and
interface-installation cost of promoting \(S\) to a Macro-Agent \\
\(G_k,M_k\) & incremental cost of the \(k\)-th local service under
generic execution and Macro-Agent execution, respectively \\
\(\delta_S,\bar\delta_S\) & conservative lower and upper bounds on
per-service cost savings \\
\(\mathcal R_S\) & architecture requirements needed for subtree \(S\) to
satisfy its Acceptance Contracts \\
\(\mathcal B=\{\chi_1,\ldots,\chi_n\}\) & finite set of installable
architecture components \\
\(\omega_j\) & nonnegative installation/residency cost of component
\(\chi_j\) \\
\(B_{ij}\) & binary indicator that component \(\chi_j\) covers
requirement \(\varrho_i\) \\
\(P_{jk}\) & binary indicator that \(\chi_j\) requires \(\chi_k\) as a
prerequisite \\
\(Q_{jk}\) & binary indicator that \(\chi_j\) and \(\chi_k\) are
mutually incompatible \\
\(H_S\) & complete auditable internal history after Macro-Agent subtree
execution \\
\(Z_S\) & Subtree ABI compressed from \(H_S\) and exposed to the
parent \\
\(\psi_S\) & measurable map from the complete history to the Subtree
ABI \\
\(Y_k\) & \(k\)-th parent control state after subtree return \\
\(\Sigma\) & global durable key-value state maintained by the runtime \\
\(\mathcal K\) & key set of the durable state \\
\(R_i,W_i\) & finite read and write sets of lease \(L_i\) \\
\(\xi_i\) & local random seed or random variable for lease \(L_i\) \\
\(\Delta_i\) & keys modified by other committed leases after \(L_i\)
obtains its snapshot and before \(L_i\) commits \\
\end{longtable}

These quantities describe only currently observable task structure and
runtime state. Task names, historical discovery descriptions, and sealed
evaluator content do not enter the promotion, architecture-synthesis, or
parallelism criteria in the following sections.

\subsection{3.5 Architecture Hotspots and Macro-Agent
Promotion}\label{architecture-hotspots-and-macro-agent-promotion}

\subsubsection{3.5.1 Cost Semantics of Local
Promotion}\label{cost-semantics-of-local-promotion}

Fix time \(t\) and consider a local subgraph \(S\subseteq V_t\) of the
current obligation graph. Eureka has two admissible continuation
classes. The first retains the current generic-execution regime and is
denoted by \(\Pi_S^{\mathrm G}\). The second first pays a one-time
promotion cost, encapsulates \(S\) as a Macro-Agent with specialized
state, memory, operators, verifiers, and local topology, and then
executes the remaining work; this class is denoted by
\(\Pi_S^{\mathrm M}\). To make the efficiency of the two classes
comparable, both continuations are required to satisfy the same local
Acceptance Contract and the same upper bound \(\alpha_S\) on failure
probability.

Let \(N_S\in\mathbb N_0\) be the number of future local service events
from time \(t\) until \(S\) completes, and assume
\(\mathbb E[N_S\mid\mathcal F_t]<\infty\). Denote the incremental costs
of the \(k\)-th service under generic and Macro-Agent execution by
\(G_k\) and \(M_k\), respectively; if the \(k\)-th service never occurs,
the corresponding cost is extended by zero. The one-time promotion cost
\(F_S\) is \(\mathcal F_t\)-measurable and includes the unavoidable
fixed cost of architecture synthesis, durable-state migration,
tool/interface binding, and parent-child interface installation.

Rather than treating high state sharing or high dependency density as
unproved sufficient conditions for promotion, the main theorem below
uses only \textbf{per-service cost differences} that can be certified by
replay, profiling, or a conservative cost model. Structural statistics
are used to estimate these differences; they do not replace the
mathematical condition.

\subsubsection{Assumption 6 (Comparable Acceptance Before and After
Promotion)}\label{assumption-6-comparable-acceptance-before-and-after-promotion}

There exist \(\pi^{\mathrm G}\in\Pi_S^{\mathrm G}\) and
\(\pi^{\mathrm M}\in\Pi_S^{\mathrm M}\) that satisfy the same local
Acceptance Contract and obey
\(\mathbb P(S\text{ is accepted correctly}\mid\mathcal F_t)\ge1-\alpha_S\).
Accordingly, the following comparison concerns costs under the same
reliability requirement and does not allow lower verifier strength to be
exchanged for lower cost.

\subsubsection{Assumption 7 (Conditional Lower Bound on Per-Service
Savings)}\label{assumption-7-conditional-lower-bound-on-per-service-savings}

Let \(\mathcal G_{k-1}\) be the filtration generated by \(\mathcal F_t\)
and the local execution trace before the \(k\)-th service begins. There
exists an \(\mathcal F_t\)-measurable \(\delta_S>0\) such that, on
\(\{N_S\ge k\}\),

\[
\boxed{
\mathbb E\!\left[
G_k-M_k
\mid
\mathcal G_{k-1},\,N_S\ge k
\right]
\ge
\delta_S,
\qquad
\text{for every }k\ge1 .
}
\tag{21}
\]

Equation (21) does not require \(G_k-M_k\) to be independent or
identically distributed, nor does it require identical savings at every
service. The condition requires only a common conservative lower bound
on each future service that actually occurs.

\subsubsection{Lemma 4 (Conditional Lower Bound on Cumulative Local
Savings)}\label{lemma-4-conditional-lower-bound-on-cumulative-local-savings}

Under Assumption 7 and \(\mathbb E[N_S\mid\mathcal F_t]<\infty\),
cumulative runtime savings of the promoted Macro-Agent relative to
generic execution satisfy

\[
\boxed{
\mathbb E\!\left[
\sum_{k=1}^{N_S}(G_k-M_k)
\;\middle|\;
\mathcal F_t
\right]
\ge
\delta_S\,
\mathbb E[N_S\mid\mathcal F_t].
}
\tag{22}
\]

\textbf{Proof.} For the nonnegative truncation
\(N_S^{(m)}:=\min(N_S,m)\), finite summation and the tower property give

\[
\mathbb E\!\left[
\sum_{k=1}^{N_S^{(m)}}(G_k-M_k)
\middle|\mathcal F_t
\right]
=
\sum_{k=1}^{m}
\mathbb E\!\left[
\mathbf 1_{\{N_S\ge k\}}(G_k-M_k)
\middle|\mathcal F_t
\right].
\]

Condition the \(k\)-th term first on \(\mathcal G_{k-1}\) and apply (21)
on \(\{N_S\ge k\}\). The term is at least
\(\delta_S\mathbb P(N_S\ge k\mid\mathcal F_t)\). The conditional
expectation of the truncated sum is therefore at least
\(\delta_S\sum_{k=1}^{m}\mathbb P(N_S\ge k\mid\mathcal F_t)\). Letting
\(m\to\infty\) and using integrability together with the tail-sum
identity for an integer-valued random variable,
\(\sum_{k\ge1}\mathbb P(N_S\ge k\mid\mathcal F_t)=\mathbb E[N_S\mid\mathcal F_t]\),
yields (22). \(\square\)

\subsubsection{Theorem 4 (Conservative Amortization Threshold for
Macro-Agent
Promotion)}\label{theorem-4-conservative-amortization-threshold-for-macro-agent-promotion}

Under Assumptions 6-7, if the one-time promotion cost satisfies

\[
\boxed{
F_S
<
\delta_S\,\mathbb E[N_S\mid\mathcal F_t],
}
\tag{23}
\]

then, under the same local Acceptance Contract and reliability
threshold, the conditional expected total cost of immediately promoting
\(S\) to a Macro-Agent and then using \(\pi^{\mathrm M}\) is strictly
smaller than the conditional expected total cost of continuing with
\(\pi^{\mathrm G}\).

\textbf{Proof.} The total-cost difference between generic continuation
and Macro-Agent continuation is

\[
\boxed{
\mathbb E\!\left[
K_S^{\mathrm G}-K_S^{\mathrm M}
\mid\mathcal F_t
\right]
=
\mathbb E\!\left[
\sum_{k=1}^{N_S}(G_k-M_k)
\middle|\mathcal F_t
\right]
-F_S .
}
\tag{24}
\]

Lemma 4 lower-bounds the right-hand side by
\(\delta_S\mathbb E[N_S\mid\mathcal F_t]-F_S\), which is strictly
positive by (23). Hence the Macro-Agent continuation has strictly lower
expected cost. Assumption 6 ensures that the gain cannot be explained by
relaxing the reliability criterion. \(\square\)

\subsubsection{Corollary 4 (Break-Even Horizon for a Deterministic
Remaining Service
Count)}\label{corollary-4-break-even-horizon-for-a-deterministic-remaining-service-count}

If \(N_S=H_S^{\mathrm{rem}}\) is already determined under
\(\mathcal F_t\) and (21) holds, a sufficient promotion condition is

\[
\boxed{
H_S^{\mathrm{rem}}
>
H_S^\star
:=
\frac{F_S}{\delta_S}.
}
\tag{25}
\]

Even if every local service is strictly cheaper under the Macro-Agent,
promotion is not cost-optimal when the remaining horizon is too short to
cover the one-time synthesis and migration cost.

\subsubsection{Proposition 10 (A Provable Condition Under Which
Promotion Has No
Advantage)}\label{proposition-10-a-provable-condition-under-which-promotion-has-no-advantage}

Suppose there exists a finite \(\mathcal F_t\)-measurable
\(\bar\delta_S\ge0\) such that every local service that occurs satisfies

\[
\boxed{
\mathbb E\!\left[
G_k-M_k
\mid
\mathcal G_{k-1},\,N_S\ge k
\right]
\le
\bar\delta_S,
}
\tag{26}
\]

and \(F_S\ge\bar\delta_S\mathbb E[N_S\mid\mathcal F_t]\). Considering
only the recurring savings represented by (26), Macro-Agent promotion
cannot achieve a strictly positive conditional expected net saving.

\textbf{Proof.} The same tower-property argument as in Lemma 4 gives an
upper bound \(\bar\delta_S\mathbb E[N_S\mid\mathcal F_t]\) on cumulative
savings. Subtracting the fixed cost \(F_S\) makes the net saving
nonpositive. \(\square\)

Theorem 4 and Proposition 10 produce an identifiable three-region
decision rule. Promotion is safe when the conservative lower bound
already exceeds the fixed cost; generic execution is safe when even the
conservative upper bound cannot cover the fixed cost; only the
intermediate uncertainty region requires additional profiling or local
trial execution. Architecture routing need not depend on task names.

\subsubsection{Proposition 11 (Lower Bound on Generic-Execution Cost
from Shared-State
Reload)}\label{proposition-11-lower-bound-on-generic-execution-cost-from-shared-state-reload}

Let \(L_S>0\) be the serialized input length of the minimal sufficient
local state required to execute subtree \(S\), and let \(J_S\) be the
number of times generic execution must reactivate that state across
sessions that do not share it. Suppose the nonnegative cost per input
unit is \(\lambda_{\mathrm{in}}\), no exact cache, pointer dereference,
or persistent local memory can share the state losslessly across those
sessions, and a Macro-Agent can retain the state persistently after the
first load. Then the additional generic-only cost due to state
restoration is at least

\[
\boxed{
C_{\mathrm{reload}}(S)
\ge
\lambda_{\mathrm{in}}\,L_S\,
\mathbb E\!\left[(J_S-1)_+\mid\mathcal F_t\right].
}
\tag{27}
\]

\textbf{Proof.} Both execution regimes may require the first local-state
load, so the first load does not contribute to the relative excess cost.
Beginning with the second activation, every session that does not share
the state must receive a sufficient representation of length at least
\(L_S\). There are \((J_S-1)_+\) such additional restorations, each
costing at least \(\lambda_{\mathrm{in}}L_S\). Taking the conditional
expectation yields (27). \(\square\)

The assumptions of (27) are explicit and necessary. If the backend
already provides exact persistent state sharing, \(C_{\mathrm{reload}}\)
may be close to zero, and high state sharing cannot by itself imply that
promotion is advantageous. Empirically, AgentPrune shows that
multi-agent pipelines can contain substantial redundant communication
and token cost (G. Zhang, Yue, et al. 2025), while TDAG and ADAS
demonstrate the feasibility of dynamic subagent generation and automated
agent-architecture design (Y. Wang et al. 2024; Hu, Lu, et al. 2024).
These results provide empirical context; the promotion threshold in
Theorem 4 is determined independently from measurable cost and remaining
horizon in the current task trajectory.

\begin{center}\rule{0.5\linewidth}{0.5pt}\end{center}

\subsection{3.6 Minimal Sufficient Agent Architecture
Realization}\label{minimal-sufficient-agent-architecture-realization}

\subsubsection{3.6.1 From Task Requirements to a Finite
Component-Selection
Problem}\label{from-task-requirements-to-a-finite-component-selection-problem}

Once Macro-Agent promotion is justified, it still does not follow that
the system should install as many memory modules, tools, verifiers, and
planning components as possible. To formalize a minimal sufficient
architecture, fix a promoted subtree \(S\) and derive a finite
requirement set from its Acceptance Contracts, state reads/writes,
required operators, tool capabilities, and topology constraints:

\[
\mathcal R_S=\{\varrho_1,\ldots,\varrho_m\}.
\]

Let the candidate component set be
\(\mathcal B=\{\chi_1,\ldots,\chi_n\}\). Each component \(\chi_j\) has
strictly positive cost \(\omega_j>0\). Let \(B_{ij}=1\) mean that
\(\chi_j\) covers requirement \(\varrho_i\); \(P_{jk}=1\) mean that
installing \(\chi_j\) requires prerequisite \(\chi_k\); and \(Q_{jk}=1\)
mean that \(\chi_j\) and \(\chi_k\) are mutually incompatible in the
current architecture namespace. The binary variable \(x_j\in\{0,1\}\)
indicates whether \(\chi_j\) is installed.

The feasible architectures satisfying requirement coverage, prerequisite
constraints, and incompatibility constraints are

\[
\boxed{
\mathcal X_S
:=
\left\{
x\in\{0,1\}^{n}:
\begin{array}{l}
\displaystyle \sum_{j=1}^{n}B_{ij}x_j\ge1,\quad i=1,\ldots,m,\\[1.5mm]
x_j\le x_k,\quad \forall(j,k)\text{ with }P_{jk}=1,\\[1mm]
x_j+x_k\le1,\quad \forall(j,k)\text{ with }Q_{jk}=1
\end{array}
\right\}.
}
\tag{28}
\]

Equation (28) encodes only \textbf{necessary architectural
capabilities}. If no finite component combination can satisfy a
requirement, then \(\mathcal X_S=\varnothing\). The correct
interpretation is that the current component library cannot realize the
Macro-Agent; the system must design an additional component or return to
generic execution rather than assuming that a nonexistent architecture
is available.

\subsubsection{Assumption 8 (Finite
Realizability)}\label{assumption-8-finite-realizability}

The number of candidate components is finite, \(n<\infty\); every
component cost satisfies \(\omega_j>0\); and
\(\mathcal X_S\neq\varnothing\).

\subsubsection{Theorem 5 (Existence and Inclusion Minimality of a
Minimal Sufficient
Architecture)}\label{theorem-5-existence-and-inclusion-minimality-of-a-minimal-sufficient-architecture}

Under Assumption 8, the optimization problem

\[
\boxed{
x_S^\star
\in
\arg\min_{x\in\mathcal X_S}
\sum_{j=1}^{n}\omega_jx_j
}
\tag{29}
\]

has at least one optimal solution. Every optimal solution \(x_S^\star\)
is inclusion-minimal among feasible architectures: there is no
\(x'\in\mathcal X_S\) such that \(x'_j\le x_{S,j}^\star\) for every
\(j\) with at least one strict inequality.

\textbf{Proof.} \(\mathcal X_S\) is a nonempty subset of the finite set
\(\{0,1\}^n\), so the positive-valued objective attains its minimum.
Suppose an optimal solution \(x_S^\star\) were not inclusion-minimal.
Then a strict coordinatewise subset \(x'\in\mathcal X_S\) would exist.
Because every \(\omega_j>0\), deleting at least one installed component
yields \(\sum_j\omega_jx'_j<\sum_j\omega_jx_{S,j}^\star\), contradicting
optimality. \(\square\)

Theorem 5 does not assert uniqueness. Multiple component combinations
can satisfy the same requirements at the same minimum cost. A
deterministic implementation that requires a unique result must specify
a reproducible tie-breaking rule; it cannot declare an arbitrary optimal
component set to be theoretically unique.

\subsubsection{Lemma 5 (Necessity of a Forced
Component)}\label{lemma-5-necessity-of-a-forced-component}

Suppose there exist a requirement \(\varrho_i\) and a component
\(\chi_j\) such that \(B_{ij}=1\) and \(B_{ik}=0\) for every
\(k\neq j\). Then every feasible architecture \(x\in\mathcal X_S\) must
satisfy \(x_j=1\). Moreover, if \(\chi_k\) is reachable from \(\chi_j\)
along the prerequisite relation, every feasible architecture must also
install \(\chi_k\).

\textbf{Proof.} Requirement coverage imposes \(\sum_kB_{ik}x_k\ge1\). If
\(\chi_j\) is the unique covering component, the constraint reduces to
\(x_j\ge1\), hence \(x_j=1\). The conclusion then propagates along
prerequisite constraints \(x_j\le x_k\). \(\square\)

Lemma 5 provides basic causal provenance for an architecture component:
a component either directly covers an irreplaceable requirement or is a
prerequisite of a required component. Theorem 5 provides no reason to
retain an expensive component that has neither form of support.

\subsubsection{Proposition 12 (General Computational Complexity of
Minimal Sufficient Architecture
Search)}\label{proposition-12-general-computational-complexity-of-minimal-sufficient-architecture-search}

Even when prerequisite and incompatibility constraints are absent and
every \(\omega_j=1\), (29) contains the classical Set Cover problem as a
special case. Exact search cannot therefore be assumed to have a
polynomial-time algorithm in general unless additional task-specific
structure is exploited. Eureka's architecture compiler may require
heuristics, branch-and-bound, modular search, or approximation in large
component spaces; minimal sufficient architecture search cannot be
treated as a zero-cost primitive.

This optimization view is related to AgentSquare, which organizes
planning, reasoning, tool use, and memory into a modular agent-search
space (Shang et al. 2024), and to ADAS, which demonstrates that complete
agent programs can be objects of automated search (Hu, Lu, et al. 2024).
Eureka differs in that the requirement set in (28) is induced by the
task structure of the currently promoted subtree rather than by a fixed
benchmark defined in advance.

\subsubsection{3.6.2 Lazy Architecture
Extension}\label{lazy-architecture-extension}

At promotion time, some optional components may become necessary only
later. Fix a component \(\chi_j\) that is not currently required by
\(\mathcal R_S\) but may be requested by a future obligation. Let
\(Y_j\in\{0,1\}\) indicate whether the component is needed at least once
before \(S\) completes, and define
\(p_j:=\mathbb P(Y_j=1\mid\mathcal F_t)\). Let \(u_j>0\) be the cost of
immediate installation. If installation is deferred until the first
request, let \(d_j\ge0\) be the installation and safe state-migration
cost and \(\ell_j\ge0\) the latency or recovery cost introduced by the
temporary pause. If early residency also incurs cumulative overhead
\(R_j\ge0\) before first use or subtree completion, the conditional
expected costs of upfront and lazy installation are

\[
\boxed{
C_j^{\mathrm U}
=
u_j+\mathbb E[R_j\mid\mathcal F_t],
\qquad
C_j^{\mathrm L}
=
p_j(d_j+\ell_j).
}
\tag{30}
\]

The delayed installation in (30) must complete before the component is
actually used, so the semantics of the corresponding scientific operator
and verifier reliability are unchanged.

\subsubsection{Assumption 9 (Safe Monotone Architecture
Extension)}\label{assumption-9-safe-monotone-architecture-extension}

If \(\chi_j\) is installed only when first required, existing certified
state can be migrated to the extended architecture without loss, and all
Acceptance Contracts for already completed obligations remain valid
before and after extension. Installing \(\chi_j\) only enlarges the
future policy/capability set; it does not revoke a previously legal
operation.

\subsubsection{Theorem 6 (Exact Selection Condition for Lazy
Architecture
Extension)}\label{theorem-6-exact-selection-condition-for-lazy-architecture-extension}

Under Assumption 9, for a single future optional component \(\chi_j\),
lazy installation has conditional expected cost no greater than upfront
installation if and only if

\[
\boxed{
p_j
\le
\frac{
u_j+\mathbb E[R_j\mid\mathcal F_t]}{d_j+\ell_j},
\qquad
d_j+\ell_j>0 .
}
\tag{31}
\]

If \(d_j+\ell_j=0\), lazy installation weakly dominates upfront
installation.

\textbf{Proof.} After first use, the two strategies have the same
component installed, so all subsequent execution costs cancel. Comparing
the two expressions in (30) and rearranging
\(C_j^{\mathrm L}\le C_j^{\mathrm U}\) gives (31). If the denominator is
zero, the lazy cost is zero whereas the upfront cost is nonnegative.
\(\square\)

\subsubsection{Corollary 5 (Install-on-Demand Principle Without
Migration
Penalty)}\label{corollary-5-install-on-demand-principle-without-migration-penalty}

If \(d_j=u_j\), \(\ell_j=0\), and \(R_j=0\), then

\[
\boxed{
C_j^{\mathrm L}
=
p_ju_j
\le
u_j
=
C_j^{\mathrm U},
}
\tag{32}
\]

and lazy extension has strictly lower expected cost whenever \(p_j<1\).
Thus, when future use of a capability is uncertain and delayed
installation incurs no semantic or migration penalty, installing every
potential component upfront is not expected-cost optimal.

\begin{center}\rule{0.5\linewidth}{0.5pt}\end{center}

\subsection{3.7 Subtree ABI and Information-Sufficient
Compression}\label{subtree-abi-and-information-sufficient-compression}

\subsubsection{3.7.1 Information Sufficiency for Parent-Level
Decisions}\label{information-sufficiency-for-parent-level-decisions}

A Macro-Agent can generate an internal reasoning trajectory far longer
than the context budget available to its parent. To determine whether
the parent can receive only a compact Subtree ABI rather than the
complete internal transcript, the relevant criterion is
\textbf{sufficiency with respect to future parent decisions}, not
whether a summary appears complete to a reader.

Let \(H_S\) denote the complete internal history when subtree \(S\)
finishes, with values in a standard Borel space \(\mathscr H_S\). Define
the Subtree ABI as a measurable map
\(\psi_S:\mathscr H_S\to\mathscr Z_S\) and write \(Z_S=\psi_S(H_S)\). In
an implementation, \(Z_S\) may contain exported artifacts, assumptions
on which those exports remain valid, unresolved debts, verification
receipts, and reopen triggers. Mathematically, the only requirement at
this point is that \(Z_S\) be a measurable state visible to the parent.

After the subtree returns, consider a parent continuation decision
process of finite length \(L\). Let the \(k\)-th parent decision state
be \(Y_k\in\mathcal Y_k\), the admissible action set be
\(\mathcal D_k(y,h)\), the stage cost be \(c_k(y,h,a)\), the stochastic
transition kernel be \(P_k(\mathrm dy'\mid y,h,a)\), and the terminal
cost be \(g_L(y,h)\). All state spaces are assumed to be standard Borel
spaces, and all cost functions are bounded and measurable, ensuring that
the dynamic-programming integrals and conditional expectations below are
well defined.

\subsubsection{Definition (Decision-Sufficient Subtree
ABI)}\label{definition-decision-sufficient-subtree-abi}

Suppose there exist measurable objects
\(\bar{\mathcal D}_k,\bar c_k,\bar P_k,\bar g_L\) depending on the
internal history only through \(z=\psi_S(h)\), such that for every
internal history \(h\), parent state \(y\), and admissible action \(a\),

\[
\boxed{
\begin{aligned}
\mathcal D_k(y,h)
&=
\bar{\mathcal D}_k(y,\psi_S(h)),\\
c_k(y,h,a)
&=
\bar c_k(y,\psi_S(h),a),\\
P_k(\cdot\mid y,h,a)
&=
\bar P_k(\cdot\mid y,\psi_S(h),a),\\
g_L(y,h)
&=
\bar g_L(y,\psi_S(h)).
\end{aligned}
}
\tag{33}
\]

Then \(Z_S\) is \textbf{decision-sufficient} for the parent continuation
problem.

Equation (33) is stronger than requiring the summary to contain all
information that appears important. The influence of the complete
internal history on every future parent action set, stage cost, state
transition, and terminal acceptance criterion must factor entirely
through \(Z_S\). The definition is consistent with Blackwell's
decision-theoretic comparison of statistical experiments: if a
compressed observation preserves every achievable risk for the relevant
decision problem, the discarded information has no additional decision
value (Blackwell 1951, 1953). The results below follow directly from
(33) and do not require invoking Blackwell's theorem as an external
lemma.

\subsubsection{Theorem 7 (Lossless Subtree
Compression)}\label{theorem-7-lossless-subtree-compression}

Assume (33). When the complete history is visible, define the optimal
finite-horizon value function by

\[
\boxed{
\begin{aligned}
V_L(y,h)
&:=g_L(y,h),\\
V_k(y,h)
&:=
\inf_{a\in\mathcal D_k(y,h)}
\left[
c_k(y,h,a)
+
\int_{\mathcal Y_{k+1}}
V_{k+1}(y',h)\,
P_k(\mathrm dy'\mid y,h,a)
\right].
\end{aligned}
}
\tag{34}
\]

Let \(\bar V_k(y,z)\) be defined by the identical Bellman recursion over
\(\bar{\mathcal D}_k,\bar c_k,\bar P_k,\bar g_L\) when only the ABI is
observed. Then, for every \(k=0,\ldots,L\),

\[
\boxed{
V_k(y,h)
=
\bar V_k\!\left(y,\psi_S(h)\right).
}
\tag{35}
\]

Consequently, receiving only \(Z_S\) gives the parent the same optimal
continuation value as receiving the entire \(H_S\). The internal
trajectory can therefore be moved to cold storage without reducing
optimal decision capability for this parent decision class.

\textbf{Proof.} We use backward induction on \(k\). At the terminal time
\(k=L\), (33) gives
\(V_L(y,h)=g_L(y,h)=\bar g_L(y,\psi_S(h))=\bar V_L(y,\psi_S(h))\).
Assume the result holds at \(k+1\). Substituting (33) and the induction
hypothesis into (34) shows that the action set, immediate cost,
transition kernel, and next-step value under the complete history all
depend on \(h\) only through \(z=\psi_S(h)\). The Bellman infimum is
therefore identical to the ABI recursion, yielding
\(V_k(y,h)=\bar V_k(y,z)\). Backward induction establishes (35).
\(\square\)

\subsubsection{Corollary 6 (Hierarchical Bound on Parent
Context)}\label{corollary-6-hierarchical-bound-on-parent-context}

Suppose the parent maintains at most \(m_t\) active Macro-Agent
interfaces at any time, the serialized fixed parent control state has
length at most \(B_0\), and every decision-sufficient ABI has serialized
length at most \(B_{\max}\). If complete subtree histories are paged in
only for audit or reopen events, the ordinary parent decision context
satisfies

\[
\boxed{
B_t^{\mathrm{parent}}
\le
B_0+m_tB_{\max}.
}
\tag{36}
\]

The bound is independent of the total length \(\sum_S|H_S|\) of internal
transcripts across active subtrees. Thus, if both the number of active
Macro-Agents and ABI size remain controlled, parent context can be
decoupled from internal reasoning horizon.

\subsubsection{Proposition 13 (Decision Sufficiency Cannot Be
Omitted)}\label{proposition-13-decision-sufficiency-cannot-be-omitted}

Suppose two internal histories \(h,h'\in\mathscr H_S\) satisfy
\(\psi_S(h)=\psi_S(h')\), but there exists a parent state \(y\) and
action \(a\) for which \(c_k(y,h,a)\neq c_k(y,h',a)\). Then \(Z_S\)
cannot be a lossless interface for a parent problem containing that
decision step.

\textbf{Proof.} The two histories are compressed to the same \(z\), so a
parent policy observing only \(z\) cannot distinguish \(h\) from \(h'\).
Yet the stage cost of taking the same action \(a\) differs under the
complete histories. No single \(\bar c_k(y,z,a)\) can therefore satisfy
(33) for both histories, and the premise of Theorem 7 fails. \(\square\)

The same conclusion holds if two internal histories with the same ABI
induce different admissible action sets, different future
state-transition kernels, or different terminal Acceptance Contracts.
Eureka must consequently construct sufficiently expressive interfaces
using typed exports, assumptions, verification receipts, and reopen
triggers rather than relying on free-form summaries.

\subsubsection{3.7.2 Completeness of Reopen
Triggers}\label{completeness-of-reopen-triggers}

Let \(U_S\) denote the certified exports that a Macro-Agent exposes to
its parent. Let \(\Lambda_S(y,h)\in\{0,1\}\) indicate whether these
exports remain valid under current parent state \(y\) and internal
history \(h\). Let \(\rho_S(y,z)\in\{0,1\}\) be an ABI-detectable reopen
trigger.

\subsubsection{Assumption 10 (Reopen
Completeness)}\label{assumption-10-reopen-completeness}

Whenever an accepted export changes from valid to invalid, the
corresponding ABI must trigger a reopen event. For every parent-state
transition \(y\to y'\),

\[
\boxed{
\Lambda_S(y,h)=1,\;
\Lambda_S(y',h)=0
\quad\Longrightarrow\quad
\rho_S(y',\psi_S(h))=1 .
}
\tag{37}
\]

\subsubsection{Proposition 14 (Export Preservation in the Absence of a
Reopen
Trigger)}\label{proposition-14-export-preservation-in-the-absence-of-a-reopen-trigger}

Under Assumption 10, suppose an export is valid at parent state \(y_0\)
and \(\rho_S(y_j,Z_S)=0\) throughout the state sequence
\(y_0,\ldots,y_m\). Then the export remains valid at every \(y_j\).

\textbf{Proof.} Assume for contradiction that \(j\) is the smallest
index at which the export becomes invalid. By minimality, the export is
valid at \(y_{j-1}\) and invalid at \(y_j\). Assumption 10 then forces
\(\rho_S(y_j,Z_S)=1\), contradicting the stated condition. \(\square\)

\begin{center}\rule{0.5\linewidth}{0.5pt}\end{center}

\subsection{3.8 Parallel Execution, Isolation, and Merge
Safety}\label{parallel-execution-isolation-and-merge-safety}

\subsubsection{3.8.1 Typed Durable State and Lease
Semantics}\label{typed-durable-state-and-lease-semantics}

Eureka permits parallel execution only when state dependencies are
explicit and auditable. Let \(\mathcal K\) denote the key set of durable
state, and let every key \(q\in\mathcal K\) take values in a standard
Borel space \(\mathfrak S_q\). The global durable state lies in the
product space

\[
\boxed{
\mathfrak S
:=
\prod_{q\in\mathcal K}\mathfrak S_q,
\qquad
\Sigma\in\mathfrak S .
}
\tag{38}
\]

Each concurrent lease \(L_i\) starts from an immutable snapshot
\(\Sigma^{(v_i)}\) and declares a finite read set
\(R_i\subset\mathcal K\) and write set \(W_i\subset\mathcal K\). Given a
local random seed \(\xi_i\), the lease produces a measurable update on
its write set,

\[
\boxed{
f_i:
\left(\prod_{q\in R_i}\mathfrak S_q\right)\times\Xi_i
\longrightarrow
\prod_{q\in W_i}\mathfrak S_q .
}
\tag{39}
\]

Equation (39) includes a critical \textbf{complete-read-set condition}:
every durable-state key that can influence the output of \(L_i\) must be
included in \(R_i\). If a dependency exists only in model context and is
not captured by runtime instrumentation, the serializability results
below do not apply.

\subsubsection{Assumption 11 (Isolation of Side Effects Before
Commit)}\label{assumption-11-isolation-of-side-effects-before-commit}

Before validation succeeds, a lease may not make an irreversible
external-world modification take effect directly. An external tool
effect must satisfy at least one of the following conditions: it can be
delayed until commit; it has a verifiable idempotency key; an exact
compensation operation exists; or execution is forced to be serial. The
assumption prevents a lease that eventually aborts from leaving an
irreversible side effect outside durable state.

\subsubsection{Lemma 6 (Commutativity of Conflict-Free
Leases)}\label{lemma-6-commutativity-of-conflict-free-leases}

Fix two leases \(L_i,L_j\) and their local random seeds. If

\[
\boxed{
W_i\cap(R_j\cup W_j)=\varnothing,
\qquad
W_j\cap(R_i\cup W_i)=\varnothing,
}
\tag{40}
\]

then their state transformations commute. For every valid initial state
\(\Sigma\),

\[
\boxed{
F_i\!\left(F_j(\Sigma)\right)
=
F_j\!\left(F_i(\Sigma)\right),
}
\tag{41}
\]

where \(F_i,F_j:\mathfrak S\to\mathfrak S\) are the global state
transformations obtained by writing the outputs of (39) back to the
corresponding write sets.

\textbf{Proof.} Consider an arbitrary key \(q\in\mathcal K\). If
\(q\notin W_i\cup W_j\), both orders preserve its original value. If
\(q\in W_i\), (40) guarantees \(q\notin W_j\), so its final value is
written only by \(F_i\). In addition, \(W_j\cap R_i=\varnothing\), so
executing \(F_j\) first cannot change any input read by \(F_i\); \(F_i\)
computes the same value in either order. The case \(q\in W_j\) is
symmetric. Every key therefore has the same final value, proving (41).
\(\square\)

\subsubsection{Corollary 7 (Safe Parallelism for Pairwise Conflict-Free
Fanout)}\label{corollary-7-safe-parallelism-for-pairwise-conflict-free-fanout}

For leases \(\{L_1,\ldots,L_m\}\), if every distinct pair \(i,j\)
satisfies (40), all execution orders yield the same durable final state.
The leases may therefore execute in parallel and their typed deltas may
be merged in any order without changing the final state.

\subsubsection{3.8.2 Optimistic Validation and
Serializability}\label{optimistic-validation-and-serializability}

In a real system, a lease may execute from an older snapshot while other
leases commit. For a lease \(L_i\) attempting to commit, let
\(\Delta_i\subseteq\mathcal K\) be the union of keys modified by all
previously committed leases after snapshot \(\Sigma^{(v_i)}\) was
obtained and before the current commit point.

Eureka intentionally uses a validation rule more conservative than
ordinary Snapshot Isolation:

\[
\boxed{
(R_i\cup W_i)\cap\Delta_i
=
\varnothing .
}
\tag{42}
\]

Equation (42) rejects both read-write and write-write interference.
Classical optimistic concurrency control traces to Kung and Robinson
(Kung and Robinson 1981). Snapshot Isolation based only on snapshot
reads and write-write conflict detection does not in general imply
serializability; Berenson et al. (Berenson et al. 1995) systematically
documented anomalies including write skew. The theorem below therefore
depends on complete read/write validation in (42), not on Snapshot
Isolation alone.

\subsubsection{Theorem 8 (Optimistic Lease
Serializability)}\label{theorem-8-optimistic-lease-serializability}

Consider a collection of successfully committed leases
\(L_1,\ldots,L_m\), indexed by actual commit order. Assume every lease
satisfies the complete-read-set condition of (39), Assumption 11, and
validation rule (42) at commit. Then the durable final state produced by
concurrent execution is exactly the state obtained by executing
\(L_1,\ldots,L_m\) serially from the common initial state in commit
order.

\textbf{Proof.} Induct on the commit index. The first committed lease
has no earlier committed modification, so its snapshot agrees with the
serial initial state on every declared read/write key and its commit
agrees with serial execution.

Assume the durable state after the first \(i-1\) actual commits equals
the serial state \(\Sigma_{i-1}^{\mathrm{ser}}\). Consider \(L_i\). The
set \(\Delta_i\) contains every key modified by an earlier successful
commit after \(L_i\) obtained its snapshot. Equation (42) gives
\(R_i\cap\Delta_i=\varnothing\), so every key read by \(L_i\) has the
same value in its snapshot as in the current
\(\Sigma_{i-1}^{\mathrm{ser}}\). The complete-read-set condition
guarantees that the result of \(L_i\) does not depend on an undeclared
key. Equation (42) also gives \(W_i\cap\Delta_i=\varnothing\), so no
unresolved concurrent write-write overwrite exists. With the local
random seed fixed, the typed delta computed during actual concurrent
execution is therefore identical to the delta that would be computed by
executing \(L_i\) from \(\Sigma_{i-1}^{\mathrm{ser}}\). Committing that
delta makes the actual state equal \(\Sigma_i^{\mathrm{ser}}\).
Induction through \(m\) proves the result. \(\square\)

\subsubsection{Corollary 8 (Default Serial Boundary for Shared Proof
State)}\label{corollary-8-default-serial-boundary-for-shared-proof-state}

If two leases have a genuine read-after-write, write-after-read, or
write-after-write dependency on the same proof or theory state, (40)
fails. If the leases also execute from different snapshots and attempt
to commit concurrently, at least the later commit may violate (42).
Unless a specialized commutative update law or merge algebra is proved,
Eureka should therefore not split the same tightly coupled proof state
across parallel sessions.

\subsubsection{Proposition 15 (Snapshot Isolation Alone Is Insufficient
for Theorem
8)}\label{proposition-15-snapshot-isolation-alone-is-insufficient-for-theorem-8}

Suppose validation rejects only \(W_i\cap\Delta_i\neq\varnothing\) and
does not check \(R_i\cap\Delta_i\). Then two leases can have disjoint
write sets, each read a state key that the other will modify, both
commit, and produce a final state that is not equivalent to any serial
order. The conclusion of Theorem 8 therefore fails under this weaker
validation rule.

\textbf{Construction.} Take Boolean keys \(x,y\) with initial state
\(x=y=1\). Lease \(L_1\) reads \(y\) and writes \(x:=0\) when \(y=1\).
Lease \(L_2\) reads \(x\) and writes \(y:=0\) when \(x=1\). Both leases
read \(x=y=1\) from the same snapshot, and their write sets are
\(\{x\}\) and \(\{y\}\), so write-write-only validation permits both
commits and yields \(x=y=0\). If \(L_1\) executes before \(L_2\)
serially, \(L_2\) reads \(x=0\) and does not set \(y:=0\); the reverse
order is symmetric. The concurrent result is therefore not equivalent to
either serial order. \(\square\)

\subsubsection{Proposition 16 (Parallelism Boundary for Irreversible
External Side
Effects)}\label{proposition-16-parallelism-boundary-for-irreversible-external-side-effects}

Suppose lease \(L_i\) performs an irreversible external operation
\(e_i\) before validation, and no delayed commit, idempotent
deduplication, or compensation map can restore the external state after
an abort. Even if durable state satisfies (42), the lease cannot obtain
an end-to-end serializability guarantee.

\textbf{Reason.} Theorem 8 controls only durable-state transitions on
\(\Sigma\). If a lease that ultimately fails validation has already
changed external state and that modification cannot be reversed,
discarding the typed delta alone cannot restore a state corresponding to
the serial history. Assumption 11 is therefore a necessary runtime
boundary for end-to-end parallel safety rather than an optional
implementation detail.

\subsubsection{3.8.3 Semantic Transparency of Event
Coalescing}\label{semantic-transparency-of-event-coalescing}

Let \(\mathfrak C\) be the runtime control-state space. For a
non-hard-interrupt event \(e\), let \(\phi_e:\mathfrak C\to\mathfrak C\)
be its deterministic effect on control state. If a set of events
\(\{e_1,\ldots,e_m\}\) commutes pairwise within the current control
epoch and none of the events triggers a hard interrupt that requires the
Meta-Agent to wake immediately, processing the events in a batch does
not change the final control state.

\subsubsection{Proposition 17 (Equivalence of Coalescing Commutative
Control
Events)}\label{proposition-17-equivalence-of-coalescing-commutative-control-events}

If, for every \(i,j\),
\(\phi_{e_i}\circ\phi_{e_j}=\phi_{e_j}\circ\phi_{e_i}\), then for any
two permutations \(\sigma,\pi\),

\[
\boxed{
\phi_{e_{\sigma(m)}}\circ\cdots\circ\phi_{e_{\sigma(1)}}(c)
=
\phi_{e_{\pi(m)}}\circ\cdots\circ\phi_{e_{\pi(1)}}(c),
\qquad
\forall c\in\mathfrak C .
}
\tag{43}
\]

The runtime can therefore coalesce multiple worker-completion events,
ready-count updates, and conflict-free receipt insertions
programmatically and wake the Meta-Agent once with a single
ControlCapsule.

\textbf{Proof.} Any finite permutation can be transformed into any other
by a sequence of adjacent transpositions. Pairwise commutativity makes
every adjacent swap preserve the composition result, so all permutations
yield the same final state. \(\square\)

If a contradiction, verifier mismatch, budget exhaustion, or acceptance
boundary changes the set of legally available subsequent actions, the
corresponding event cannot be delayed and coalesced with ordinary
events; it must be sent to the upper-level controller as a hard
interrupt. Equation (43) therefore defines the exact safety boundary for
event coalescing rather than asserting unconditionally that fewer
Meta-Agent wake-ups are always correct.

The remaining theoretical sections retain all notation, assumptions, and
numbering introduced through Sections 3.1-3.8. The probability space
remains \((\Omega,\mathscr F,\mathbb P)\), the admissible information
filtration remains \((\mathcal F_t)_{t\ge0}\), the task stopping time is
\(\tau\), the dynamic obligation graph is \(G_t=(V_t,E_t)\), the agent
architecture is
\(\mathcal A=(\mathcal S,\mathcal M,\mathcal U,\mathcal V,\mathcal T,\mathcal P)\),
and the optimal expected cost under the reliability constraint remains
\(\mathcal C_\alpha(\mathcal A;T)\). For a promoted local subtree
\(S\subseteq V_t\), the remaining local service count \(N_S\),
Macro-Agent service cost \(M_k\), and local stopping time \(\tau_S\)
defined in Section 3.5 continue to be used. For an obligation \(o\), the
input space \(\mathcal I_o\), artifact space \(\mathcal Y_o\),
certificate space \(\mathcal C_o\), semantic predicate \(\Phi_o\), and
acceptance verifier \(V_o\) defined in Section 3.4 also remain
unchanged.

The additional symbols below are used in Sections 3.9-3.12.

\begin{longtable}[]{@{}
  >{\raggedright\arraybackslash}p{(\columnwidth - 2\tabcolsep) * \real{0.5000}}
  >{\raggedright\arraybackslash}p{(\columnwidth - 2\tabcolsep) * \real{0.5000}}@{}}
\toprule\noalign{}
\begin{minipage}[b]{\linewidth}\raggedright
Symbol
\end{minipage} & \begin{minipage}[b]{\linewidth}\raggedright
Definition
\end{minipage} \\
\midrule\noalign{}
\endhead
\bottomrule\noalign{}
\endlastfoot
\(\mathfrak M_t(S)\) & finite set of admissible mutations that may be
proposed for Macro-Agent subtree \(S\) at time \(t\) \\
\(m\in\mathfrak M_t(S)\) & a specific architecture mutation \\
\(\mathcal A_S^{(m)}\) & candidate architecture obtained by applying
mutation \(m\) to current local architecture \(\mathcal A_S\) \\
\(C_m^{\mathrm{evo}}\) & fixed diagnosis, generation, evaluation,
migration, and deployment cost of mutation \(m\) \\
\(M_k^{(m)}\) & incremental cost of the \(k\)-th future local service
under candidate architecture \(\mathcal A_S^{(m)}\) \\
\(\gamma_m,\bar\gamma_m\) & conservative lower and upper bounds on
per-service savings produced by mutation \(m\) \\
\(D_n^{(m)}\) & observed incumbent-minus-mutation cost difference in
paired micro-evaluation \(n\) \\
\(\mu_m\) & conditional mean of \(D_n^{(m)}\), i.e., average per-service
saving of mutation \(m\) \\
\([\ell_n^{(m)},u_n^{(m)}]\) & time-uniform confidence sequence for
\(\mu_m\) \\
\(\beta_m\) & allowable statistical error probability in mutation
evaluation \\
\(h_S\) & \(\mathcal F_t\)-measurable conservative lower bound on
\(\mathbb E[N_S\mid\mathcal F_t]\) \\
\(\mathfrak L_t\) & append-only ledger of certified receipts through
time \(t\) \\
\(\mathfrak L_t^{\mathrm{act}}\) & active certified receipts whose
versioned dependencies remain valid at time \(t\) \\
\(\operatorname{dep}(r)\) & direct provenance dependencies of certified
receipt \(r\) \\
\(I_t\) & seed receipts or versioned state references explicitly
invalidated at time \(t\) \\
\(\operatorname{cl}_{\mathrm{dep}}(I_t)\) & closure of invalid
descendants reachable from \(I_t\) in the provenance dependency graph \\
\(\mathcal C_o^{-}\) & refutation-certificate space for obligation
\(o\) \\
\(W_o\) & sound refutation verifier for obligation \(o\) \\
\(\theta_o\) & scalar parameter decided by a statistical verifier \\
\(\theta_o^\star\) & statistical threshold specified by the Acceptance
Contract \\
\(Y^\star\) & sealed evaluation content related to benchmark evaluation
or final scientific results that must remain isolated during
production \\
\(\mathcal E\) & possible value space of \(Y^\star\) \\
\(\mathcal H_t^{\mathrm{prod}}\) & production-only control history
through time \(t\) \\
\(Q_t^e\) & legal observation kernel when the sealed evaluator takes
value \(e\in\mathcal E\) \\
\(\tau_{\mathrm f}\) & freeze stopping time at which production
architecture/evolution policy is sealed \\
\(\tau_{\mathcal A}\) & stopping time at which a specialized Macro-Agent
architecture is first fixed \\
\(\tau_{\mathcal D}\) & stopping time at which the associated scientific
artifact is submitted, with \(\tau_{\mathcal A}\le\tau_{\mathcal D}\) \\
\(\mathfrak J_S\) & structured architecture instance defined by subtree
requirements, component coverage, prerequisites, conflicts, and
component costs \\
\(\mathfrak C\) & canonical architecture compiler that reads only
\(\mathfrak J_S\) \\
\(\mathscr D\) & measurable output space of scientific-discovery
artifacts \\
\end{longtable}

All newly introduced quantities are obtained from the current task
trajectory, versioned receipts, structured requirements, or
target-independent evaluation. The sealed evaluation variable
\(Y^\star\) is excluded from mutation generation, architecture
compilation, planning, and verifier design.

\begin{center}\rule{0.5\linewidth}{0.5pt}\end{center}

\subsection{3.9 Amortized Theory of Governed
Self-Evolution}\label{amortized-theory-of-governed-self-evolution}

\subsubsection{3.9.1 Cost Semantics of an Evolution
Intervention}\label{cost-semantics-of-an-evolution-intervention}

Suppose subtree \(S\) has already been promoted to a Macro-Agent
according to Theorem 4 and executes with current architecture
\(\mathcal A_S\). A mutation \(m\in\mathfrak M_t(S)\) may modify a
governed subset of runtime, prompt/operator, memory/skill, tool
interface, state/verifier, or topology, but it may not modify the root
Task Contract, frozen Acceptance Contracts, source cutoff, or sealed
evaluator boundary. The candidate architecture after applying the
mutation is denoted by \(\mathcal A_S^{(m)}\).

If the incumbent is retained, the cost of the \(k\)-th future local
service remains \(M_k\), as defined in Section 3.5; after mutation, the
corresponding cost is \(M_k^{(m)}\). The fixed cost
\(C_m^{\mathrm{evo}}\ge0\) of an evolution event includes bottleneck
diagnosis, mutation generation, micro-evaluation, required state
migration, and deployment. We do not define improvement through an
ambiguous scalar score. The reliability constraint used throughout the
paper remains in force: incumbent and mutated architecture are compared
only when both satisfy the same local Acceptance Contract and the same
failure-probability bound.

\subsubsection{Assumption 12 (Evolution Contract
Invariance)}\label{assumption-12-evolution-contract-invariance}

For candidate mutation \(m\), there exist an admissible policy
\(\pi^{(m)}\in\Pi(\mathcal A_S^{(m)})\) and the incumbent policy
\(\pi\in\Pi(\mathcal A_S)\) such that both satisfy the same local
Acceptance Contract and both obey
\(\mathbb P(S\text{ is accepted correctly}\mid\mathcal F_t)\ge1-\alpha_S\).
If a mutation changes verifier semantics or lowers the reliability
criterion, the cost theorems in this section cannot be used for
admission; the mutation must return to Meta-Agent contract-level review.

\subsubsection{Assumption 13 (Conservative Gain Lower Bound for a
Recurring
Bottleneck)}\label{assumption-13-conservative-gain-lower-bound-for-a-recurring-bottleneck}

Let \(\mathcal G_{k-1}\) again denote the filtration before the \(k\)-th
local service. There exists an \(\mathcal F_t\)-measurable
\(\gamma_m>0\) such that every future local service that actually occurs
satisfies

\[
\boxed{
\mathbb E\!\left[
M_k-M_k^{(m)}
\mid
\mathcal G_{k-1},\,N_S\ge k
\right]
\ge
\gamma_m,
\qquad k\ge1 .
}
\tag{44}
\]

Assumption 13 does not require the mutation to be strictly better at
every service and does not require the cost differences to be
independent or identically distributed. It requires only a conservative
conditional lower bound on future benefit arising from the same
recurring bottleneck. A one-off anomaly generally cannot support
\(\gamma_m>0\) and therefore does not automatically satisfy the
evolution-admission condition.

\subsubsection{Lemma 7 (Cumulative Conservative Benefit of a
Mutation)}\label{lemma-7-cumulative-conservative-benefit-of-a-mutation}

Under Assumption 13 and \(\mathbb E[N_S\mid\mathcal F_t]<\infty\),
cumulative expected runtime savings of the mutation over the remaining
local task satisfy

\[
\boxed{
\mathbb E\!\left[
\sum_{k=1}^{N_S}
\left(M_k-M_k^{(m)}\right)
\middle|\mathcal F_t
\right]
\ge
\gamma_m\,
\mathbb E[N_S\mid\mathcal F_t].
}
\tag{45}
\]

\textbf{Proof.} Truncate at \(N_S^{(q)}=\min(N_S,q)\) and rewrite the
finite sum as \(\sum_{k=1}^{q}\mathbf 1_{\{N_S\ge k\}}(M_k-M_k^{(m)})\).
Condition each term on \(\mathcal G_{k-1}\); Assumption 13 gives the
lower bound \(\gamma_m\mathbb P(N_S\ge k\mid\mathcal F_t)\). Sum over
\(k\), let \(q\to\infty\), and use the integer-valued tail-sum identity
\(\sum_{k\ge1}\mathbb P(N_S\ge k\mid\mathcal F_t)=\mathbb E[N_S\mid\mathcal F_t]\)
to obtain (45). The probability structure is the same as in Lemma 4,
with the comparison changed from generic-versus-Macro-Agent to
incumbent-versus-mutated-Macro-Agent. \(\square\)

\subsubsection{Theorem 9 (Governed Evolution Admission
Threshold)}\label{theorem-9-governed-evolution-admission-threshold}

Under Assumptions 12-13, if

\[
\boxed{
C_m^{\mathrm{evo}}
<
\gamma_m\,
\mathbb E[N_S\mid\mathcal F_t],
}
\tag{46}
\]

then immediately paying \(C_m^{\mathrm{evo}}\) and switching to
\(\mathcal A_S^{(m)}\) yields a strictly lower conditional expected
remaining total cost under the same local acceptance reliability.

\textbf{Proof.} Let \(K_{S,t}^{(0)}\) be the remaining total cost from
time \(t\) when the incumbent is retained, and let \(K_{S,t}^{(m)}\) be
the remaining total cost after mutation. External task costs common to
both continuations cancel, giving

\[
\boxed{
\mathbb E\!\left[
K_{S,t}^{(0)}-K_{S,t}^{(m)}
\mid\mathcal F_t
\right]
=
\mathbb E\!\left[
\sum_{k=1}^{N_S}
\left(M_k-M_k^{(m)}\right)
\middle|\mathcal F_t
\right]
-
C_m^{\mathrm{evo}} .
}
\tag{47}
\]

Lemma 7 lower-bounds the right-hand side by
\(\gamma_m\mathbb E[N_S\mid\mathcal F_t]-C_m^{\mathrm{evo}}\), which is
strictly positive by (46). The mutated architecture therefore has
strictly lower conditional expected cost. Assumption 12 excludes gains
obtained by lowering correctness. \(\square\)

\subsubsection{Corollary 9 (Break-Even Horizon for
Evolution)}\label{corollary-9-break-even-horizon-for-evolution}

If the remaining service count is determined by the current obligation
graph, \(N_S=H_S^{\mathrm{rem}}\), the sufficient condition of Theorem 9
becomes \(H_S^{\mathrm{rem}}>C_m^{\mathrm{evo}}/\gamma_m\). Thus, when
the remaining horizon is short, immediate evolution can remain inferior
to the incumbent even when the candidate mutation has lower steady-state
per-step cost, because diagnosis, evaluation, and migration impose fixed
overhead.

\subsubsection{Proposition 18 (Weak Dominance of the Lowest Sufficient
Mutation)}\label{proposition-18-weak-dominance-of-the-lowest-sufficient-mutation}

Suppose \(m_1,m_2\in\mathfrak M_t(S)\) both satisfy Assumption 12 and,
under the same current state and the same future legal observation
kernels, induce identical conditional distributions over all future
durable states, Acceptance events, and stepwise runtime costs. If
\(C_{m_1}^{\mathrm{evo}}<C_{m_2}^{\mathrm{evo}}\), then \(m_1\) strictly
cost-dominates \(m_2\).

\textbf{Proof.} The two mutations have identical continuation laws after
deployment, so future runtime cost and task reliability have the same
conditional distribution. Their only difference in total cost is the
one-time evolution cost, which is strictly smaller for \(m_1\). Any
admissible policy selecting \(m_2\) can therefore be replaced by one
selecting \(m_1\) with lower total cost. \(\square\)

The proposition provides a strict version of the \emph{lowest sufficient
mutation level} principle. A lower-level modification theoretically
dominates a higher-level modification only when it can induce the same
future behaviour distribution. If a state-level or topology-level
mutation enables behaviour that no lower-level mutation can realize,
smaller scope alone does not justify choosing the lower-level mutation.

Recent systems show that execution-trace-driven harness evolution is
practically feasible. Meta-Harness treats complete harness code as an
outer-loop optimization object and uses source code and execution traces
of prior candidates to propose new harnesses (Lee et al. 2026).
Retrospective Harness Optimization uses previous trajectories,
self-validation, and self-preference to produce harness updates (Pan et
al. 2026), while Self-Harness explicitly combines weakness mining,
minimal harness proposals, and regression validation (H. Zhang et al.
2026). Theorem 9 addresses a complementary question: when such evolution
has positive amortized value.

\subsubsection{3.9.2 Sequential Micro-Evaluation and Safety Under
Optional
Stopping}\label{sequential-micro-evaluation-and-safety-under-optional-stopping}

Evolution admission must also avoid the statistical bias created by
stopping as soon as a small number of favourable samples are observed.
For fixed mutation \(m\), let \(D_1^{(m)},D_2^{(m)},\ldots\) denote
paired cost differences on target-independent micro-evaluation tasks,
with positive values indicating that the mutation is cheaper than the
incumbent. Conditioned on \(\mathcal F_t\), assume that the observations
have common conditional mean \(\mu_m\) and satisfy an explicit bounded,
sub-Gaussian, or other sequential-inference condition sufficient to
construct a confidence sequence. Let \([\ell_n^{(m)},u_n^{(m)}]\) be a
time-uniform confidence sequence with confidence level \(1-\beta_m\):

\[
\boxed{
\mathbb P\!\left(
\mu_m\in
[\ell_n^{(m)},u_n^{(m)}]
\ \text{for all }n\ge1
\;\middle|\;
\mathcal F_t
\right)
\ge
1-\beta_m .
}
\tag{48}
\]

Time-uniform confidence sequences preserve coverage at data-dependent
stopping times without requiring the evaluation sample size to be fixed
in advance (Howard et al. 2021).

Assume the runtime also has an \(\mathcal F_t\)-measurable conservative
value \(h_S\ge0\) satisfying \(h_S\le\mathbb E[N_S\mid\mathcal F_t]\).
Define the first evaluation stopping time that certifies positive
amortized benefit by

\[
\boxed{
\tau_m^{\mathrm{val}}
:=
\inf\!\left\{
n\ge1:
h_S\,\ell_n^{(m)}
>
C_m^{\mathrm{evo}}
\right\},
}
\tag{49}
\]

with \(\tau_m^{\mathrm{val}}=\infty\) if the set is empty.

\subsubsection{Proposition 19 (Reliability of Evolution Admission Under
Optional
Stopping)}\label{proposition-19-reliability-of-evolution-admission-under-optional-stopping}

On the coverage event in (48), if \(\tau_m^{\mathrm{val}}<\infty\), then
\(C_m^{\mathrm{evo}}<\mu_m\mathbb E[N_S\mid\mathcal F_t]\).
Consequently, the probability of incorrectly certifying positive
amortized value because of micro-evaluation sampling error is at most
\(\beta_m\).

\textbf{Proof.} On the simultaneous coverage event in (48),
\(\ell_n^{(m)}\le\mu_m\) for every \(n\). When the stopping condition
first holds,
\(C_m^{\mathrm{evo}}<h_S\ell_n^{(m)}\le h_S\mu_m\le\mu_m\mathbb E[N_S\mid\mathcal F_t]\).
An erroneous admission can occur only when simultaneous coverage fails,
whose conditional probability is at most \(\beta_m\). \(\square\)

\begin{center}\rule{0.5\linewidth}{0.5pt}\end{center}

\subsection{3.10 Verification Semantics and Monotone Scientific
Progress}\label{verification-semantics-and-monotone-scientific-progress}

\subsubsection{3.10.1 Append-Only Receipt Ledger and Explicit
Invalidation
Propagation}\label{append-only-receipt-ledger-and-explicit-invalidation-propagation}

Section 3.4 distinguishes semantic correctness \(\Phi_o\) from the
acceptance verifier \(V_o\). We now further distinguish between a result
that has historically obtained a valid certificate and a result that
remains admissible as an active premise under the current versioned
dependency state.

A certified receipt \(r\in\mathfrak L_t\) contains at least an
obligation identifier, exact input \(i_r\in\mathcal I_o\), artifact
\(y_r\in\mathcal Y_o\), certificate \(c_r\in\mathcal C_o\), direct
dependencies \(\operatorname{dep}(r)\), and all versioned state
references. The historical ledger is append-only and never overwritten.
Let \(A_t^{+}\) denote the new set of accepted receipts. Then

\[
\boxed{
\mathfrak L_{t+1}
=
\mathfrak L_t
\cup
A_t^{+},
\qquad
A_t^{+}
\subseteq
\left\{
r:
V_o(i_r,y_r,c_r)=1
\right\}.
}
\tag{50}
\]

If a premise, source version, assumption, or state reference is
explicitly invalidated, let the corresponding seeds be \(I_t\). In the
receipt dependency graph, let \(\operatorname{cl}_{\mathrm{dep}}(I_t)\)
be the transitive closure obtained by following the direction in which
one receipt is depended on by another. The active certified view is
updated by

\[
\boxed{
\mathfrak L_{t+1}^{\mathrm{act}}
=
\left(
\mathfrak L_t^{\mathrm{act}}
\setminus
\operatorname{cl}_{\mathrm{dep}}(I_t)
\right)
\cup
A_t^{+}.
}
\tag{51}
\]

Historical monotonicity of the ledger is thus separated from
revocability of the active view. A certificate obtained in the past is
not silently deleted or rewritten by language-model output, but its
eligibility as a \textbf{current premise} is explicitly revoked when a
dependency is invalidated.

\subsubsection{Assumption 14 (Provenance Completeness and Version
Consistency)}\label{assumption-14-provenance-completeness-and-version-consistency}

For every active receipt \(r\), all prerequisite artifacts, assumptions,
source snapshots, and durable-state versions that can affect the
certified conclusion are included in \(\operatorname{dep}(r)\) or in the
receipt's versioned references. Whenever a relevant dependency changes,
the corresponding object is added to the invalidation seeds. Formal
study of database provenance as a record of which inputs contribute to
an output can be traced to Green, Karvounarakis, and Tannen (Green,
Karvounarakis, and Tannen 2007). Eureka uses only the
dependency-tracking principle and does not assume that scientific claims
possess relational-algebra structure.

\subsubsection{Lemma 8 (Semantic Soundness of Receipt
Addition)}\label{lemma-8-semantic-soundness-of-receipt-addition}

Under the verifier soundness of Assumption 4 in Section 3.4, every
\(r\in A_t^{+}\) satisfies the true semantics on its exact input:
\(\Phi_o(i_r,y_r)=1\).

\textbf{Proof.} Equation (50) requires every new receipt to satisfy
\(V_o(i_r,y_r,c_r)=1\). Verifier soundness implies
\(\Phi_o(i_r,y_r)=1\). \(\square\)

\subsubsection{Theorem 10 (Soundness of the Active Certified
Ledger)}\label{theorem-10-soundness-of-the-active-certified-ledger}

Assume that every receipt in the initial active ledger
\(\mathfrak L_0^{\mathrm{act}}\) has a sound certificate, Assumption 14
holds, every added receipt satisfies (50), and every dependency
invalidation propagates to all dependent descendants according to (51).
Then, for every \(t\ge0\), each \(r\in\mathfrak L_t^{\mathrm{act}}\)
simultaneously satisfies:

\begin{enumerate}
\def\labelenumi{\arabic{enumi}.}
\tightlist
\item
  \(\Phi_o(i_r,y_r)=1\);
\item
  every versioned dependency declared by the receipt is still current at
  time \(t\); and
\item
  no provenance path exists from any known invalidation seed to \(r\).
\end{enumerate}

\textbf{Proof.} Induct on \(t\). The base case holds by assumption.
Suppose the result holds at time \(t\). Active receipts at \(t+1\) fall
into two classes. The first class comes from
\(\mathfrak L_t^{\mathrm{act}}\setminus\operatorname{cl}_{\mathrm{dep}}(I_t)\).
The induction hypothesis establishes their previous soundness; Equation
(51) removes every dependency descendant reachable from the invalidation
seeds; and Assumption 14 guarantees that every relevant dependency
change appears among the invalidation seeds. The remaining receipts
therefore still have current dependency versions and no known
invalidation path. The second class consists of \(A_t^{+}\). Lemma 8
establishes semantic soundness, and the construction in (50) records
current versioned dependencies. The new receipts also satisfy all three
properties. Induction completes the proof. \(\square\)

\subsubsection{Corollary 10 (Certified State Cannot Be Silently
Overwritten by Unverified
Text)}\label{corollary-10-certified-state-cannot-be-silently-overwritten-by-unverified-text}

If the language model generates a natural-language claim that conflicts
with an active receipt but has no valid refutation or invalidation
certificate, the claim cannot enter the certified ledger through (50)
and cannot remove the existing active receipt through (51). Under
Theorem 10, unverified output therefore cannot silently rewrite
certified scientific state.

\subsubsection{3.10.2 Three-Valued Semantics: PASS, FAIL, and
INCONCLUSIVE}\label{three-valued-semantics-pass-fail-and-inconclusive}

An acceptance verifier alone is insufficient to interpret inability to
prove a claim as a refutation. For obligation \(o\), additionally define
a refutation-certificate space \(\mathcal C_o^{-}\) and an executable
refutation verifier

\[
W_o:
\mathcal I_o\times
\mathcal Y_o\times
\mathcal C_o^{-}
\longrightarrow
\{0,1\}.
\]

\subsubsection{Assumption 15 (Refutation-Verifier
Soundness)}\label{assumption-15-refutation-verifier-soundness}

For every \((i,y,c^{-})\), if \(W_o(i,y,c^{-})=1\), then
\(\Phi_o(i,y)=0\).

Given the available certificate-search results, Eureka defines the
three-valued decision rule

\[
\boxed{
\mathsf D_o(i,y)
=
\begin{cases}
\mathsf P,
&
\exists c^{+}\in\mathcal C_o:
V_o(i,y,c^{+})=1,\\[1mm]
\mathsf F,
&
\exists c^{-}\in\mathcal C_o^{-}:
W_o(i,y,c^{-})=1,\\[1mm]
\mathsf U,
&
\text{otherwise},
\end{cases}
}
\tag{52}
\]

where \(\mathsf P\), \(\mathsf F\), and \(\mathsf U\) denote certified
pass, certified fail, and inconclusive, respectively. Because both
verifiers are sound, the first two cases cannot both be semantically
correct. If an implementation observes both verifiers accepting, the
event is treated as a verifier inconsistency and triggers a hard
interrupt rather than an arbitrary choice between outcomes.

\subsubsection{Proposition 20 (Forcing INCONCLUSIVE to FAIL Is Unsound
in
General)}\label{proposition-20-forcing-inconclusive-to-fail-is-unsound-in-general}

Suppose the acceptance verifier \(V_o\) is incomplete: there exists
\((i^\star,y^\star)\) such that \(\Phi_o(i^\star,y^\star)=1\), but no
\(c\in\mathcal C_o\) satisfies \(V_o(i^\star,y^\star,c)=1\), and no
sound refutation certificate exists. Equation (52) then returns
\(\mathsf U\). Any binary rule that maps every \(\mathsf U\) to
\(\mathsf F\) produces a false negative at \((i^\star,y^\star)\) and
therefore cannot preserve semantic soundness.

\subsubsection{3.10.3 Anytime Validity of Statistical
Verifiers}\label{anytime-validity-of-statistical-verifiers}

For an obligation that cannot produce an exact certificate but whose
Acceptance Contract can be expressed as a threshold on a statistical
parameter, let the observation sequence determine an unknown parameter
\(\theta_o\in\Theta_o\subseteq\mathbb R\), and let \(\theta_o^\star\) be
the threshold defining the target. Let \([\ell_n,u_n]\) be a
time-uniform confidence sequence with confidence level \(1-\beta_o\):

\[
\boxed{
\mathbb P\!\left(
\theta_o\in[\ell_n,u_n]
\ \text{for all }n\ge1
\right)
\ge
1-\beta_o .
}
\tag{53}
\]

Such confidence sequences preserve coverage at arbitrary data-dependent
stopping times (Howard et al. 2021).

For any stopping time \(\tau_v\) with respect to the observation
filtration, define the statistical verifier

\[
\boxed{
\mathsf D_o^{\mathrm{stat}}
=
\begin{cases}
\mathsf P,
&
\ell_{\tau_v}\ge\theta_o^\star,\\[1mm]
\mathsf F,
&u_{\tau_v}<\theta_o^\star,\\[1mm]
\mathsf U,
&
\ell_{\tau_v}<\theta_o^\star\le u_{\tau_v}.
\end{cases}
}
\tag{54}
\]

\subsubsection{Corollary 11 (Error Control for Anytime Statistical
Verification)}\label{corollary-11-error-control-for-anytime-statistical-verification}

On the coverage event in (53), (54) cannot return \(\mathsf P\) when
\(\theta_o<\theta_o^\star\) and cannot return \(\mathsf F\) when
\(\theta_o\ge\theta_o^\star\). Hence, even when \(\tau_v\) is selected
adaptively from online observations, the probability of an incorrect
certified decision satisfies

\[
\boxed{
\mathbb P\!\left(
\mathsf D_o^{\mathrm{stat}}=\mathsf P,\,
\theta_o<\theta_o^\star
\ \text{or}\
\mathsf D_o^{\mathrm{stat}}=\mathsf F,\,
\theta_o\ge\theta_o^\star
\right)
\le
\beta_o .
}
\tag{55}
\]

\textbf{Proof.} On the simultaneous coverage event,
\(\ell_n\le\theta_o\le u_n\) for every \(n\). If
\(\ell_{\tau_v}\ge\theta_o^\star\), then
\(\theta_o\ge\ell_{\tau_v}\ge\theta_o^\star\), so PASS cannot be wrong.
If \(u_{\tau_v}<\theta_o^\star\), then
\(\theta_o\le u_{\tau_v}<\theta_o^\star\), so FAIL cannot be wrong.
Every incorrect certified decision is contained in the failure event of
the confidence sequence, whose probability is at most \(\beta_o\).
\(\square\)

\begin{center}\rule{0.5\linewidth}{0.5pt}\end{center}

\subsection{3.11 Evaluation-Information Isolation and Causal Validity of
Execution}\label{evaluation-information-isolation-and-causal-validity-of-execution}

\subsubsection{3.11.1 Counterfactual Evaluation
Isolation}\label{counterfactual-evaluation-isolation}

The principal risk in replaying a historical discovery is not limited to
whether the final prompt contains the answer directly; a sealed
evaluation variable may influence the production trajectory through
architecture design, retrieval, evolution feedback, or evaluator
signals. To formalize this issue, let \(Y^\star\in\mathcal E\) denote
the sealed evaluation payload that must remain isolated during
production. For any \(e\in\mathcal E\), construct a counterfactual world
in which the sealed evaluator is fixed to \(Y^\star=e\), while the Task
Contract, legally available pre-cutoff public corpus, and all runtime
resources not derived from the evaluation target remain unchanged.

Define the production history by
\(\mathcal H_t^{\mathrm{prod}}=(C_0,O_0,A_0,\ldots,O_t)\). The policy
kernel \(\pi_t(\mathrm da\mid h_t)\) is permitted to read only
production history. Given a production history and action, the
conditional distribution of the next observation is denoted by
\(Q_t^{e}(\mathrm do\mid h_t,a_t)\).

\subsubsection{Assumption 16 (Runtime Evaluation
Isolation)}\label{assumption-16-runtime-evaluation-isolation}

Before the freeze time \(\tau_{\mathrm f}\):

\begin{enumerate}
\def\labelenumi{\arabic{enumi}.}
\tightlist
\item
  policy kernels \(\pi_t\) do not take \(e\) as an argument;
\item
  evaluator files, evaluation-target-derived feedback, and post-cutoff
  sources containing evaluation-only target information are inaccessible
  to the runtime;
\item
  for every legal production history \(h_t\), action \(a_t\), and every
  \(e,e'\in\mathcal E\),
  \(Q_t^{e}(\cdot\mid h_t,a_t)=Q_t^{e'}(\cdot\mid h_t,a_t)\); and
\item
  \(\tau_{\mathrm f}\) is an almost-surely finite stopping time with
  respect to the production filtration.
\end{enumerate}

The third condition is the probabilistic form of a source firewall.
Physical isolation of evaluator files is insufficient if a search tool
can still return different observations because public post-evaluation
pages reveal different evaluation content.

\subsubsection{Lemma 9 (Evaluator Invariance of Finite-Horizon
Production
Traces)}\label{lemma-9-evaluator-invariance-of-finite-horizon-production-traces}

Under Assumption 16, for every fixed \(n\in\mathbb N_0\) and any
\(e,e'\in\mathcal E\), the conditional distributions of the first \(n\)
steps of the production trace are identical:

\[
\boxed{
\mathcal L\!\left(
\mathcal H_n^{\mathrm{prod}}
\mid
Y^\star=e
\right)
=
\mathcal L\!\left(
\mathcal H_n^{\mathrm{prod}}
\mid
Y^\star=e'
\right).
}
\tag{56}
\]

\textbf{Proof.} Induct on \(n\). At \(n=0\), \(C_0,O_0\) are generated
by the same production initial law. Suppose the trace laws agree through
\(n\). Given the same history \(h_n\), both counterfactual worlds use
the same policy kernel \(\pi_n(\cdot\mid h_n)\), so their conditional
action distributions agree. By condition 3 of Assumption 16, the
next-observation kernels also agree given the same \(h_n,a_n\).
Integrating over \(h_n\) yields equality of the trace law at \(n+1\).
Induction gives (56). \(\square\)

\subsubsection{Theorem 11 (Evaluation-Isolated Counterfactual
Invariance)}\label{theorem-11-evaluation-isolated-counterfactual-invariance}

Under Assumption 16, for any \(e,e'\in\mathcal E\), the complete
production trace stopped at freeze time has the same distribution:

\[
\boxed{
\mathcal L\!\left(
\mathcal H_{\tau_{\mathrm f}}^{\mathrm{prod}},
\tau_{\mathrm f}
\mid
Y^\star=e
\right)
=
\mathcal L\!\left(
\mathcal H_{\tau_{\mathrm f}}^{\mathrm{prod}},
\tau_{\mathrm f}
\mid
Y^\star=e'
\right).
}
\tag{57}
\]

If the two counterfactual worlds are additionally coupled with common
random numbers so that policy sampling, tool randomness, and environment
randomness use the same seeds, their production traces can be chosen to
be pathwise identical almost surely before \(\tau_{\mathrm f}\).

\textbf{Proof.} For any \(n\), the event \(\{\tau_{\mathrm f}=n\}\) is
measurable with respect to \(\mathcal H_n^{\mathrm{prod}}\) because
\(\tau_{\mathrm f}\) is a stopping time. Lemma 9 states that every
finite history has the same distribution under \(e\) and \(e'\), so the
distributions of
\((\mathcal H_n^{\mathrm{prod}},\mathbf 1_{\{\tau_{\mathrm f}=n\}})\)
also agree. Summing over \(n\) and using \(\tau_{\mathrm f}<\infty\)
almost surely gives equality of the stopped-trace distribution. Under
common-random-number coupling, the same history processed by the same
kernels and the same random numbers yields the same action and
observation at every step, strengthening the result to pathwise
equality. \(\square\)

\subsubsection{Corollary 12 (Sealed-Evaluator Mutation Trace
Invariance)}\label{corollary-12-sealed-evaluator-mutation-trace-invariance}

If the production runtime is deterministic, or if every random source is
fixed by a seed, then under Theorem 11 replacing the sealed evaluator
value \(e\) by any \(e'\) must preserve the canonical serialization and
hash of the entire pre-freeze production trace:

\[
\boxed{
\operatorname{Hash}\!\left(
\mathcal H_{\tau_{\mathrm f}}^{\mathrm{prod}}(e)
\right)
=
\operatorname{Hash}\!\left(
\mathcal H_{\tau_{\mathrm f}}^{\mathrm{prod}}(e')
\right).
}
\tag{58}
\]

The corollary gives a directly implementable evaluation-isolation audit.
Before a sealed evaluator enters the formal evaluation phase, modifying
its internal representation should not affect production orchestration,
agent promotion, architecture synthesis, or governed evolution.

\subsubsection{Proposition 21 (Evaluation-Time Information Leakage
Breaks Observation-Kernel
Invariance)}\label{proposition-21-evaluation-time-information-leakage-breaks-observation-kernel-invariance}

Suppose online retrieval under some production action can return a page
containing a benchmark question, a sealed evaluation variable, or an
evaluation-only answer, and the set of such pages changes between
\(Y^\star=e\) and \(Y^\star=e'\). Then, in general, there exist a
history \(h_t\) and action \(a_t\) such that
\(Q_t^{e}(\cdot\mid h_t,a_t)\neq Q_t^{e'}(\cdot\mid h_t,a_t)\).
Assumption 16 fails and Theorem 11 no longer applies.

Search-Time Data Contamination provides empirical evidence that
search-enabled agents can retrieve question-answer pairs from publicly
available benchmark datasets at inference time, with accuracy on
contaminated subsets decreasing after relevant sources are blocked (Han
et al. 2025). Source cutoffs and query audits therefore belong in the
formal assumptions for evaluation isolation; prompt-level masking of the
answer alone is insufficient.

\subsubsection{Proposition 22 (Runtime Isolation Does Not Replace
Design-Stage Evaluation
Separation)}\label{proposition-22-runtime-isolation-does-not-replace-design-stage-evaluation-separation}

Suppose the runtime satisfies Assumption 16 but the production policy is
itself generated by a design process
\(\pi=\mathfrak B(Y^\star,\mathcal Z)\), where \(\mathcal Z\) denotes
other development data. If there exist \(e,e'\) such that
\(\mathfrak B(e,\mathcal Z)\neq\mathfrak B(e',\mathcal Z)\), then
Theorem 11 establishes only that the fixed policy no longer reads the
evaluator at runtime; it does not establish causal independence of
architecture design from the sealed evaluation variable.

\textbf{Reason.} The induction in Lemma 9 requires both counterfactual
worlds to use the same \(\pi_t\). Once the policy itself changes with
\(e\), the induction fails at its first step. A complete
evaluation-isolation claim therefore also requires the decomposition
policy, promotion gate, component library, mutation prior, and Method
Broker rules to be frozen before the sealed evaluation target or to be
optimized only on target-independent tasks or proxies.

\subsubsection{Proposition 23 (Evidence Hierarchy of Sealed Evaluation
and Prospective
Holdout)}\label{proposition-23-evidence-hierarchy-of-sealed-evaluation-and-prospective-holdout}

Under Theorem 11, sealed evaluation can establish the reachability
statement that a frozen production policy can arrive at a result using
legally available pre-evaluation information. Runtime invariance alone,
however, cannot eliminate unrecorded design-time selection bias when the
architecture designer already knew the historical result before freezing
the policy. If a policy is publicly frozen at time \(t_0\) and the
evaluation target is first generated or first becomes verifiable only
after \(t_0\), a prospective holdout additionally removes the
information path in which an architecture is selected from a known
target, and therefore provides stronger causal evidence.

\begin{center}\rule{0.5\linewidth}{0.5pt}\end{center}

\subsection{3.12 Two-Stage Decomposition of Architecture Discovery and
Scientific
Discovery}\label{two-stage-decomposition-of-architecture-discovery-and-scientific-discovery}

\subsubsection{3.12.1 Two Stopping Times and a Structured Architecture
Instance}\label{two-stopping-times-and-a-structured-architecture-instance}

Eureka does not generate the final scientific output in a single
invocation of the root Meta-Agent. In the first stage, a local
architecture is formed from the obligation trajectory; in the second
stage, the Macro-Agent governed by that architecture performs the
subsequent scientific search. To make the two stages separately
evaluable, let \(\tau_{\mathcal A}\) be the stopping time at which the
architecture for a promoted subtree \(S\) is first sealed, and let
\(\tau_{\mathcal D}\) be the stopping time at which the associated
scientific artifact is submitted, with
\(\tau_{\mathcal A}\le\tau_{\mathcal D}\le\tau\).

Section 3.6 encodes the architecture requirements of a promoted subtree
using \(\mathcal R_S\), the component-coverage matrix \(B\),
prerequisite matrix \(P\), incompatibility matrix \(Q\), and
component-cost vector \(\omega\). We collect these objects together with
typed state/topology requirements that do not depend on task-entity
names into the structured instance

\[
\boxed{
\mathfrak J_S
:=
\left(
\mathcal R_S,
B,
P,
Q,
\omega,
r_S,
w_S,
\nu_S
\right),
}
\tag{59}
\]

where \(r_S,w_S\) are canonical representations of the persistent state
read/write relations of all obligations in the subtree and \(\nu_S\) is
a canonical representation of the associated Acceptance Contracts. At
architecture freeze time, all quantities in \(\mathfrak J_S\) must be
measurable with respect to \(\mathcal F_{\tau_{\mathcal A}}\).

Let \(\mathcal X(\mathfrak J_S)\) be the feasible architecture set
induced by (28), with the positive component-cost objective of Section
3.6. To obtain reproducible output when multiple equal-cost optima
exist, introduce a task-name-independent canonical tie-break functional
\(\zeta:\{0,1\}^n\to\mathbb N_0\) depending only on stable component
identifiers. Define the canonical architecture compiler by

\[
\boxed{
\mathfrak C(\mathfrak J_S)
:=
\operatorname*{arg\,min}_{x\in\mathcal X(\mathfrak J_S)}
\left(
\omega^\top x,
\zeta(x)
\right),
}
\tag{60}
\]

where minimization is lexicographic. Assumption 8 guarantees that the
feasible set is finite and nonempty, so (60) uniquely determines an
architecture vector.

\subsubsection{Assumption 17 (Architecture-Stage
Measurability)}\label{assumption-17-architecture-stage-measurability}

Both \(\mathfrak J_S\) and \(\mathfrak C(\mathfrak J_S)\) are measurable
with respect to \(\mathcal F_{\tau_{\mathcal A}}\), and the compiler
does not read \(Y^\star\), post-cutoff sources, or
evaluation-target-derived feedback.

\subsubsection{Lemma 10 (Information Separation Between the Two
Stages)}\label{lemma-10-information-separation-between-the-two-stages}

Under Assumption 17, the stage-one architecture output
\(X_S^{\mathcal A}:=\mathfrak C(\mathfrak J_S)\) is completely
determined by \(\mathcal F_{\tau_{\mathcal A}}\). The stage-two
scientific artifact \(D_S\in\mathscr D\) can depend only on
\(X_S^{\mathcal A}\) and legally expanded information in
\(\mathcal F_{\tau_{\mathcal D}}\) after \(\tau_{\mathcal A}\).
Consequently, for every measurable \(A\subseteq\{0,1\}^n\) and
\(B_D\subseteq\mathscr D\),

\[
\boxed{
\mathbb P\!\left(
X_S^{\mathcal A}\in A,
D_S\in B_D
\right)
=
\mathbb E\!\left[
\mathbf 1_{\{X_S^{\mathcal A}\in A\}}
\,
\mathbb P\!\left(
D_S\in B_D
\mid
\mathcal F_{\tau_{\mathcal A}},
X_S^{\mathcal A}
\right)
\right].
}
\tag{61}
\]

\textbf{Proof.} The architecture output is
\(\mathcal F_{\tau_{\mathcal A}}\)-measurable, so its indicator can be
taken outside the conditional expectation. Applying the tower property
then yields (61). The factorization does not assert statistical
independence between architecture and discovery; it specifies the
temporal information structure in which architecture is formed from
currently available information and subsequent discovery is generated
under that architecture and later legal observations. \(\square\)

\subsubsection{3.12.2 Structure-Driven Differentiation of Specialized
Agents}\label{structure-driven-differentiation-of-specialized-agents}

For two promoted subtrees \(S_1,S_2\), let
\(\mathfrak J_1,\mathfrak J_2\) be their structured architecture
instances and define their optimal architecture sets by

\[
\boxed{
\mathcal X_i^\star
:=
\operatorname*{arg\,min}_{x\in\mathcal X(\mathfrak J_i)}
\omega_i^\top x,
\qquad i\in\{1,2\}.
}
\tag{62}
\]

The informal statement that two tasks look different is insufficient to
conclude that the generated architectures must differ. If
\(\mathcal X_1^\star\cap\mathcal X_2^\star\neq\varnothing\), the same
minimal sufficient architecture may serve both tasks. The main result
therefore uses separation of the optimal sets as the provable condition.

\subsubsection{Theorem 12 (Structure-Driven Specialized-Agent
Emergence)}\label{theorem-12-structure-driven-specialized-agent-emergence}

Suppose both promoted subtrees satisfy Assumptions 8 and 17 and, in a
common component namespace,

\[
\boxed{
\mathcal X_1^\star
\cap
\mathcal X_2^\star
=
\varnothing .
}
\tag{63}
\]

Then every compiler that returns an exact minimum-cost feasible
architecture for each instance must produce different architecture
vectors. In particular, the canonical compiler satisfies

\[
\boxed{
\mathfrak C(\mathfrak J_1)
\neq
\mathfrak C(\mathfrak J_2).
}
\tag{64}
\]

If the evaluation-isolation conditions of Theorem 11 additionally hold,
the architecture difference cannot be attributed to the value of the
sealed evaluator and must arise from legally observable structured
architecture instances before freeze together with the compiler rules
fixed in advance.

\textbf{Proof.} Suppose for contradiction that
\(\mathfrak C(\mathfrak J_1)=\mathfrak C(\mathfrak J_2)=x^\star\).
Because the compiler returns an exact minimum-cost feasible architecture
on both instances, \(x^\star\in\mathcal X_1^\star\) and
\(x^\star\in\mathcal X_2^\star\), so
\(x^\star\in\mathcal X_1^\star\cap\mathcal X_2^\star\), contradicting
(63). The second statement follows directly from Theorem 11: mutation of
the sealed evaluator does not alter the pre-freeze production trace, and
therefore does not alter \(\mathfrak J_i\) or the compiler output.
\(\square\)

\subsubsection{Corollary 13 (Architecture Differentiation Induced by an
Irreplaceable
Requirement)}\label{corollary-13-architecture-differentiation-induced-by-an-irreplaceable-requirement}

Suppose subtree \(S_1\) contains a requirement \(\varrho_a\) whose
unique covering component is \(\chi_a\), so Lemma 5 forces every
\(x\in\mathcal X(\mathfrak J_1)\) to satisfy \(x_a=1\). Suppose that in
\(S_2\), \(\chi_a\) covers no requirement, is not a prerequisite of any
necessary installed component, and has \(\omega_a>0\). Inclusion
minimality from Theorem 5 then forces every \(x\in\mathcal X_2^\star\)
to satisfy \(x_a=0\). Hence

\[
\boxed{
\mathcal X_1^\star
\cap
\mathcal X_2^\star
=
\varnothing,
\qquad
\mathfrak C(\mathfrak J_1)
\neq
\mathfrak C(\mathfrak J_2).
}
\tag{65}
\]

\textbf{Proof.} The unique-coverage property in \(S_1\) forces \(x_a=1\)
by Lemma 5. For \(S_2\), if an optimal solution contained \(x_a=1\),
deleting \(\chi_a\) would violate neither requirement coverage nor
prerequisite constraints. Since \(\omega_a>0\), deletion strictly lowers
the cost, contradicting optimality. The optimal sets therefore take
incompatible values at coordinate \(a\), so their intersection is empty;
Theorem 12 gives the architecture difference. \(\square\)

The corollary provides a formal condition under which Theory-Discovery
and Math/Conjecture Macro-Agents must differ structurally without
encoding any particular scientific result in the compiler. Specialized
agents are mathematically forced to diverge only when one epistemic
structure creates an irreplaceable state, verifier, or operator
requirement that the other task does not require. TDAG provides a
reference point for dynamic subagent generation (Y. Wang et al. 2024),
ADAS for automated agent-program design (Hu, Lu, et al. 2024), and MaAS
for query-conditioned architecture search (G. Zhang, Niu, et al. 2025).
Theorem 12 differs by deriving specialized architectures from feasible
sets induced by online obligation structure.

\subsubsection{Proposition 24 (Architecture Causal Trace Under Task
Anonymization)}\label{proposition-24-architecture-causal-trace-under-task-anonymization}

Suppose tasks \(T\) and \(T'\) differ only in domain-entity names,
historical labels, and natural-language aliases, and there exists an
isomorphism preserving requirements, component coverage, prerequisites,
incompatibilities, state read/write relations, and Acceptance Contracts.
If component identifiers and tie-break functional \(\zeta\) are mapped
consistently under the isomorphism, then

\[
\boxed{
\mathfrak C(\mathfrak J_{S'})
=
\varphi_{\mathcal B}\!\left(
\mathfrak C(\mathfrak J_S)
\right),
}
\tag{66}
\]

where \(\varphi_{\mathcal B}\) is the corresponding permutation of the
component namespace.

\textbf{Proof.} The structural isomorphism maps the feasible
architecture set of one instance bijectively to the feasible set of the
other while preserving component costs and tie-break ordering. The
lexicographic optimization in (60) therefore has corresponding unique
solutions in the two spaces. This result instantiates Proposition 1 at
the architecture-compiler level. \(\square\)

\subsubsection{Proposition 25 (Architecture Discovery and Scientific
Discovery Require Separate
Evaluation)}\label{proposition-25-architecture-discovery-and-scientific-discovery-require-separate-evaluation}

If the architecture-stage evaluator checks only requirement coverage,
Acceptance compatibility, cost, and evaluation isolation, while the
scientific evaluator checks the discovery artifact only after
\(\tau_{\mathcal D}\), architecture quality and scientific-discovery
quality can be reported as two distinct random variables. Conversely, if
stage-one architecture search directly optimizes the score of final
held-out evaluation content, both Assumption 17 and the design-stage
isolation condition of Theorem 11 fail. Theorem 12 may still establish
that two structural outputs differ, but it can no longer support a
causal interpretation in which the specialized agent architecture is
determined solely by the task structure available before scientific
discovery.

The notation of Sections 3.1-3.12 is retained in the remaining theory.
In particular, the probability space is
\((\Omega,\mathscr F,\mathbb P)\), the admissible information filtration
is \((\mathcal F_t)_{t\ge0}\), the task stopping time is \(\tau\), step
cost is \(\kappa_t\), cumulative cost is \(K_\tau\), root-task success
is \(S_\tau\in\{0,1\}\), the dynamic obligation graph is
\(G_t=(V_t,E_t)\), and the local agent architecture is
\(\mathcal A=(\mathcal S,\mathcal M,\mathcal U,\mathcal V,\mathcal T,\mathcal P)\).
Promotion subtree \(S\), one-time promotion cost \(F_S\), future local
service count \(N_S\), and per-service costs \(G_k,M_k\) retain their
definitions from Section 3.5. Mutation \(m\), fixed evolution cost
\(C_m^{\mathrm{evo}}\), candidate service cost \(M_k^{(m)}\), and
conservative per-step gain \(\gamma_m\) retain their definitions from
Section 3.9. Production history \(\mathcal H_t^{\mathrm{prod}}\), sealed
evaluation variable \(Y^\star\), freeze stopping time
\(\tau_{\mathrm f}\), and counterfactual observation kernels \(Q_t^e\)
retain their definitions from Section 3.11.

The remaining symbols are:

\begin{longtable}[]{@{}
  >{\raggedright\arraybackslash}p{(\columnwidth - 2\tabcolsep) * \real{0.5000}}
  >{\raggedright\arraybackslash}p{(\columnwidth - 2\tabcolsep) * \real{0.5000}}@{}}
\toprule\noalign{}
\begin{minipage}[b]{\linewidth}\raggedright
Symbol
\end{minipage} & \begin{minipage}[b]{\linewidth}\raggedright
Definition
\end{minipage} \\
\midrule\noalign{}
\endhead
\bottomrule\noalign{}
\endlastfoot
\(\mathcal J_\alpha(\pi;T)\) & expected total cost of fixed admissible
policy \(\pi\) at reliability threshold \(1-\alpha\) \\
\(\pi^{[0]},\ldots,\pi^{[J]}\) & hybrid policies obtained by
successively replacing Eureka decision modules with an oracle under the
same information constraint \\
\(J\) & number of independently replaceable decision modules; we use
\(J=6\) \\
\(\Delta_j^{\mathrm H}\) & exact total-cost difference induced by the
\(j\)-th hybrid replacement \\
\(d_{\mathrm{TV}}\) & total variation distance between two probability
measures \\
\(\varepsilon_j\) & cumulative total-variation discrepancy of decision
module \(j\) relative to its oracle kernel \\
\(L_j\) & upper bound on continuation-cost oscillation from one decision
of module \(j\) \\
\(\Theta\) & runtime transformation that does not change the semantic
action law or verifier semantics \\
\(u_t\) & true persistent net-value signal of the current Macro-Agent
relative to generic execution at time \(t\) \\
\(\widehat u_t\) & runtime estimate of \(u_t\) \\
\(\eta\) & uniform absolute-error bound for architecture-value
estimation \\
\(a_+,a_-\) & promotion and demotion hysteresis thresholds, with
\(a_+>a_-\) \\
\(Z_t^{\mathrm A}\) & architecture-mode indicator; 0 denotes generic
execution and 1 a promoted Macro-Agent \\
\(\rho\) & upper bound on one-step variation of the true architecture
net-value signal \\
\(\mathcal V_n(u)\) & total variation of the sequence
\(u_0,\ldots,u_n\) \\
\(c_{\mathrm{sw}}\) & upper bound on the fixed cost of one architecture
switch \\
\(c_t(\mathcal A)\) & conditional expected local cost of architecture
\(\mathcal A\) at time \(t\) in a nonstationary task \\
\(\mathcal A_t^\star\) & optimal local architecture comparator at time
\(t\) under the same information and reliability constraints \\
\(\mathcal R_n^{\mathrm{dyn}}\) & dynamic architecture regret over
length \(n\) relative to a dynamic comparator sequence \\
\(H_{\mathrm P}\) & deterministic remaining service horizon used in
promotion experiments \\
\(H_{\mathrm E}\) & deterministic remaining affected-service horizon
used in evolution experiments \\
\end{longtable}

The six replaceable decision modules, in a fixed order used only to
construct exact telescoping hybrids, are architecture selection,
receding-horizon planning, promotion, governed evolution,
coordination/merge, and verification/control gating. The ordering does
not assert a causal ranking of importance.

\begin{center}\rule{0.5\linewidth}{0.5pt}\end{center}

\subsection{3.13 End-to-End Regret and Cost
Decomposition}\label{end-to-end-regret-and-cost-decomposition}

\subsubsection{3.13.1 Cost of a Fixed Policy Under a Reliability
Constraint}\label{cost-of-a-fixed-policy-under-a-reliability-constraint}

Equation (3) takes an infimum over all policies implementable by a given
architecture \(\mathcal A\). To compare Eureka's realized policy with an
oracle comparator under the same information constraint, we further
define the reliability-constrained cost of a fixed admissible policy
\(\pi\) by

\[
\boxed{
\mathcal J_\alpha(\pi;T)
:=
\begin{cases}
\mathbb E_T^\pi[K_\tau],
&
\mathbb P_T^\pi(S_\tau=1)\ge1-\alpha,\\[1mm]
+\infty,
&
\mathbb P_T^\pi(S_\tau=1)<1-\alpha .
\end{cases}
}
\tag{67}
\]

Hence
\(\mathcal C_\alpha(\mathcal A;T)=\inf_{\pi\in\Pi(\mathcal A)}\mathcal J_\alpha(\pi;T)\).
Equation (67) preserves the comparison principle used throughout the
paper: a cost reduction obtained by lowering root correctness is not
counted as an efficiency improvement.

Let \(\pi^{[0]}=\pi^{\mathrm E}\) be the actual Eureka policy. Let
\(\pi^\dagger\) be an \textbf{oracle under the same information
constraint}. At every time, the oracle can access only \(\mathcal F_t\)
and uses the same Task Contract, source cutoff, tool permissions, and
Acceptance Contracts, but selects optimal module-level decisions within
the admissible policy class. The oracle does not observe future
observations and does not access the sealed evaluator, so it is not
clairvoyant. Set \(\pi^{[J]}=\pi^\dagger\).

For \(j=1,\ldots,J\), hybrid policy \(\pi^{[j]}\) replaces the first
\(j\) decision modules by the corresponding module kernels of
\(\pi^\dagger\), while every remaining module retains the Eureka
implementation. The construction is used only for theoretical
attribution and does not require the runtime to maintain \(J+1\) systems
simultaneously.

\subsubsection{Assumption 18 (Reliability Comparability of Hybrid
Policies)}\label{assumption-18-reliability-comparability-of-hybrid-policies}

Every \(\pi^{[j]}\) is admissible, \(\mathcal F_t\)-adapted, satisfies
the same root Acceptance Contract and success-probability constraint
\(\mathbb P_T^{\pi^{[j]}}(S_\tau=1)\ge1-\alpha\), and has finite
expected total cost \(\mathbb E_T^{\pi^{[j]}}[K_\tau]<\infty\).

The assumption does not follow automatically from replacing one local
module by a supposedly better module, because interfaces between modules
may be coupled. If a hybrid replacement breaks verifier semantics or
state compatibility, the module cannot be attributed independently and
must instead be grouped with the coupled module before constructing a
new hybrid.

\subsubsection{Theorem 13 (Exact Hybrid Regret
Decomposition)}\label{theorem-13-exact-hybrid-regret-decomposition}

Under Assumption 18, Eureka's system-level excess cost relative to the
same-information oracle,

\[
\mathcal R_{\mathrm{sys}}(T)
:=
\mathcal J_\alpha(\pi^{[0]};T)
-
\mathcal J_\alpha(\pi^{[J]};T),
\]

has the exact telescoping decomposition

\[
\boxed{
\mathcal R_{\mathrm{sys}}(T)
=
\sum_{j=1}^{J}
\Delta_j^{\mathrm H}(T),
\qquad
\Delta_j^{\mathrm H}(T)
:=
\mathcal J_\alpha(\pi^{[j-1]};T)
-
\mathcal J_\alpha(\pi^{[j]};T).
}
\tag{68}
\]

\textbf{Proof.} Expanding the right-hand side of (68),

\[
\sum_{j=1}^{J}
\left[
\mathcal J_\alpha(\pi^{[j-1]};T)
-
\mathcal J_\alpha(\pi^{[j]};T)
\right],
\]

all intermediate terms
\(\mathcal J_\alpha(\pi^{[1]};T),\ldots,\mathcal J_\alpha(\pi^{[J-1]};T)\)
cancel pairwise, leaving
\(\mathcal J_\alpha(\pi^{[0]};T)-\mathcal J_\alpha(\pi^{[J]};T)\).
Assumption 18 ensures every term is finite, so the telescoping identity
is well defined. \(\square\)

Theorem 13 is an \textbf{exact attribution identity} and does not imply
that every \(\Delta_j^{\mathrm H}\) is nonnegative. Module interactions
can make one isolated oracle replacement temporarily increase cost even
when the complete oracle is better overall. Empirical ablations should
therefore not interpret every module difference as an independent
positive contribution. The use of time-varying comparators is
conceptually related to dynamic regret in nonstationary online learning,
where static and dynamic comparators must be distinguished (Zhao et al.
2024). Equation (68), however, concerns internal module replacement in
an agent system and follows from the telescoping construction
independently of that literature.

\subsubsection{3.13.2 From Local Decision Discrepancy to Module-Level
Regret
Bounds}\label{from-local-decision-discrepancy-to-module-level-regret-bounds}

To connect \(\Delta_j^{\mathrm H}\) to observable module-decision error,
fix module \(j\) and treat the complete production history through time
\(t\) as control state \(h_t\). When module \(j\) is activated at
\(h_t\), let the action kernels used by Eureka and the replacement
hybrid be \(q_{j,t}^{[j-1]}(\cdot\mid h_t)\) and
\(q_{j,t}^{[j]}(\cdot\mid h_t)\), respectively. The total variation
distance between probability measures \(P,Q\) is
\(d_{\mathrm{TV}}(P,Q):=\sup_B|P(B)-Q(B)|\).

Let \(Q_{j,t}^{[j]}(h,a)\) denote the conditional expected remaining
cost obtained by forcing module \(j\) to choose action \(a\) at history
\(h\), after which every future decision follows \(\pi^{[j]}\). To bound
the long-horizon effect of one module disagreement, we require an
explicit action-sensitivity bound on continuation cost.

\subsubsection{Assumption 19 (Bounded Local Continuation-Cost
Oscillation)}\label{assumption-19-bounded-local-continuation-cost-oscillation}

There exists a finite constant \(L_j\ge0\) such that, for every
module-\(j\) decision history \(h\) with positive visitation
probability,

\[
\sup_a Q_{j,t}^{[j]}(h,a)
-
\inf_a Q_{j,t}^{[j]}(h,a)
\le
L_j .
\]

In addition, the expected number of activations of module \(j\) before
task termination is finite.

\subsubsection{Lemma 11 (One-Module Performance-Difference
Identity)}\label{lemma-11-one-module-performance-difference-identity}

Under Assumptions 18-19, if \(\pi^{[j-1]}\) and \(\pi^{[j]}\) differ
only in the kernel of module \(j\), then

\[
\boxed{
\Delta_j^{\mathrm H}(T)
=
\mathbb E_T^{\pi^{[j-1]}}
\!\left[
\sum_{t=0}^{\tau-1}
\mathbf 1_{\{j\text{ active at }t\}}
\left(
\int Q_{j,t}^{[j]}(H_t,a)\,
q_{j,t}^{[j-1]}(\mathrm da\mid H_t)
-
\int Q_{j,t}^{[j]}(H_t,a)\,
q_{j,t}^{[j]}(\mathrm da\mid H_t)
\right)
\right].
}
\tag{69}
\]

\textbf{Proof.} Treat the complete history \(H_t\) as the Markov state
and the terminal state as absorbing. Define the cost-to-go value
associated with replacement policy \(\pi^{[j]}\). Along a trajectory
generated by \(\pi^{[j-1]}\), at every time subtract the cost of
switching from the current history to \(\pi^{[j]}\) from the actual
remaining cost. Applying the tower property at adjacent times cancels
every transition term and every action term from modules other than
\(j\), leaving only the action-kernel difference at activations of
module \(j\). Summing over \(t<\tau\) produces the initial endpoint
\(\mathcal J_\alpha(\pi^{[j-1]};T)-\mathcal J_\alpha(\pi^{[j]};T)\),
while the terminal value is zero, yielding (69). Because the stopping
time has finite expected cost and the activation count is integrable,
the identity can first be proved for horizon \(\tau\wedge n\) and then
extended by letting \(n\to\infty\). The derivation has the same Bellman
telescoping structure as the performance-difference lemma in
reinforcement learning (Kakade and Langford 2002). \(\square\)

Define the cumulative module discrepancy by

\[
\varepsilon_j
:=
\mathbb E_T^{\pi^{[j-1]}}
\!\left[
\sum_{t=0}^{\tau-1}
\mathbf 1_{\{j\text{ active at }t\}}
 d_{\mathrm{TV}}
\!\left(
q_{j,t}^{[j-1]}(\cdot\mid H_t),
q_{j,t}^{[j]}(\cdot\mid H_t)
\right)
\right].
\]

For any measurable function \(f\) with oscillation at most \(L_j\),
probability measures \(P,Q\) satisfy
\(|\int f\,\mathrm dP-\int f\,\mathrm dQ|\le L_j d_{\mathrm{TV}}(P,Q)\).
Lemma 11 therefore gives the following bound.

\subsubsection{Corollary 14 (Total Regret Bound from Module Decision
Error)}\label{corollary-14-total-regret-bound-from-module-decision-error}

If \(\pi^{[J]}=\pi^\dagger\) is the optimal comparator in the
same-information policy class, then \(\mathcal R_{\mathrm{sys}}(T)\ge0\)
and

\[
\boxed{
0
\le
\mathcal R_{\mathrm{sys}}(T)
\le
\sum_{j=1}^{J}
L_j\,\varepsilon_j .
}
\tag{70}
\]

\textbf{Proof.} Theorem 13 gives the exact sum. Lemma 11 and the
total-variation integration bound give
\(|\Delta_j^{\mathrm H}|\le L_j\varepsilon_j\). By the triangle
inequality,
\(\mathcal R_{\mathrm{sys}}\le\sum_j|\Delta_j^{\mathrm H}|\le\sum_jL_j\varepsilon_j\).
Oracle optimality gives the nonnegative lower bound. \(\square\)

Equation (70) identifies explicit quantities for later experimental
attribution. Local errors in architecture selection, planning,
promotion, evolution, coordination, and verifier gating must be
estimated from observable action-kernel discrepancy or deterministic
decision mismatch rather than inferred only from final success rate.

\subsubsection{Proposition 26 (Policy Invariance Under a
Semantics-Preserving Runtime
Transformation)}\label{proposition-26-policy-invariance-under-a-semantics-preserving-runtime-transformation}

Suppose runtime transformation \(\Theta\), under common-random-number
coupling, produces exactly the same semantic actions, tool arguments,
certified receipts, and final scientific artifact as the original
runtime on every legal task trajectory. Suppose also that
\(\kappa_t^\Theta\le\kappa_t\) almost surely and that the inequality is
strict on an event of positive probability. Then

\[
\boxed{
\mathcal L(D^\Theta)
=
\mathcal L(D),
\qquad
\mathbb P(S_\tau^\Theta=1)
=
\mathbb P(S_\tau=1),
\qquad
\mathbb E[K_\tau^\Theta]
<
\mathbb E[K_\tau],
}
\tag{71}
\]

where \(D\) denotes the final scientific artifact.

\textbf{Proof.} Semantic actions, tool arguments, and verifier receipts
are pathwise identical under the coupling, so the final artifact and
Acceptance event are pathwise identical, proving the first two
equalities. Step cost never increases and decreases strictly on an event
of positive probability, so summation and expectation give the strict
inequality in total expected cost. \(\square\)

Proposition 26 characterizes the theoretical boundary that exact
caching, PlanDelta, prefix reuse, certified common-subexpression
elimination, and deterministic closure must satisfy. If an optimization
changes the scientific action law or discovery distribution, it can no
longer be classified as a pure runtime-efficiency transformation and
must instead be evaluated at the policy level.

\begin{center}\rule{0.5\linewidth}{0.5pt}\end{center}

\subsection{3.14 Stability, Hysteresis, and Architecture
Oscillation}\label{stability-hysteresis-and-architecture-oscillation}

\subsubsection{3.14.1 Promotion-Demotion
Hysteresis}\label{promotion-demotion-hysteresis}

The promotion theorem in Section 3.5 depends on current estimates of
cost and remaining horizon. In an online task, these estimates change
with incoming observations. If promotion and demotion share the same
threshold, small estimation noise can make the architecture oscillate
repeatedly between generic and Macro-Agent modes. To establish a
quantitative stability result, define \(u_t\in\mathbb R\) as the true
persistent net value, at time \(t\) and under the \textbf{current task
distribution and reliability constraint}, of retaining the promoted
Macro-Agent relative to generic execution. A positive \(u_t\) means that
promotion has lower expected future cost. The runtime observes only
\(\widehat u_t=u_t+e_t\).

Let \(Z_t^{\mathrm A}\in\{0,1\}\) denote the architecture mode. Given
thresholds \(a_+>a_-\), define the hysteretic switching law

\[
\boxed{
Z_{t+1}^{\mathrm A}
=
\begin{cases}
1,
&
Z_t^{\mathrm A}=0
\ \text{and}\ 
\widehat u_t\ge a_+,\\[1mm]
0,
&
Z_t^{\mathrm A}=1
\ \text{and}\ 
\widehat u_t\le a_-,\\[1mm]
Z_t^{\mathrm A},
&
\text{otherwise}.
\end{cases}
}
\tag{72}
\]

Equation (72) does not treat hysteresis as an empirical heuristic; it
specifies an exact control law from which the switch-count bound below
is derived. Hysteresis and dwell-time mechanisms have long been used in
switched and hybrid control to prevent chattering. For example, Efimov,
Panteley, and Loria (2009) analyzes hysteresis-based supervisors, while
Kussaba et al. (2017) uses hysteretic switching to remove chattering in
hybrid pose control. The Eureka result is derived independently for
discrete architecture-value switching.

\subsubsection{Assumption 20 (Bounded Estimation Error and True-Signal
Drift)}\label{assumption-20-bounded-estimation-error-and-true-signal-drift}

There exist \(\eta\ge0\) and \(\rho\ge0\) such that, for every \(t\),
\(|\widehat u_t-u_t|\le\eta\) and \(|u_{t+1}-u_t|\le\rho\).

\subsubsection{Lemma 12 (Necessary True-Signal Variation Between
Opposite Architecture
Switches)}\label{lemma-12-necessary-true-signal-variation-between-opposite-architecture-switches}

Let \(s<t\) be two consecutive architecture switches in opposite
directions; for example, suppose time \(s\) triggers generic \(\to\)
Macro-Agent and time \(t\) is the first subsequent trigger of
Macro-Agent \(\to\) generic. Under Assumption 20,

\[
\boxed{
\sum_{r=s}^{t-1}
|u_{r+1}-u_r|
\ge
 a_+-a_--2\eta .
}
\tag{73}
\]

The same result holds when demotion precedes promotion.

\textbf{Proof.} Promotion at \(s\) implies \(\widehat u_s\ge a_+\),
hence \(u_s\ge a_+-\eta\). Demotion at \(t\) implies
\(\widehat u_t\le a_-\), hence \(u_t\le a_-+\eta\). Therefore
\(u_s-u_t\ge a_+-a_--2\eta\). The triangle inequality gives
\(\sum_{r=s}^{t-1}|u_{r+1}-u_r|\ge|u_t-u_s|\), which yields (73).
\(\square\)

\subsubsection{Theorem 14 (Anti-Chattering Stability and Switch-Count
Bound)}\label{theorem-14-anti-chattering-stability-and-switch-count-bound}

Under Assumption 20, suppose the hysteresis width satisfies
\(a_+-a_->2\eta\). Then:

\begin{enumerate}
\def\labelenumi{\arabic{enumi}.}
\tightlist
\item
  If \(\rho>0\), any two consecutive switches in opposite directions are
  separated by at least
\end{enumerate}

\[
\boxed{
d_{\min}
=
\left\lceil
\frac{a_+-a_--2\eta}{\rho}
\right\rceil
}
\tag{74}
\]

discrete control steps. If \(\rho=0\), noise alone cannot trigger a
second switch in the opposite direction.

\begin{enumerate}
\def\labelenumi{\arabic{enumi}.}
\setcounter{enumi}{1}
\tightlist
\item
  Define the total variation of the true value sequence over horizon
  \(n\) by
\end{enumerate}

\[
\mathcal V_n(u)
:=
\sum_{t=0}^{n-1}|u_{t+1}-u_t|.
\]

If \(N_{\mathrm{sw}}(n)\) is the total number of architecture switches
over times \(0,\ldots,n\), then

\[
\boxed{
N_{\mathrm{sw}}(n)
\le
1+
\frac{\mathcal V_n(u)}{a_+-a_--2\eta}.
}
\tag{75}
\]

\textbf{Proof.} For the first statement, Lemma 12 and the one-step drift
bound imply that over \(d\) intervals the true signal can change by at
most \(d\rho\). Achieving the variation required by (73) therefore
requires \(d\rho\ge a_+-a_--2\eta\), yielding (74). When \(\rho=0\), the
true signal is constant and the hysteresis gap exceeds the maximum
two-sided estimation error \(2\eta\), so the two opposite switching
conditions cannot both occur sequentially.

For the second statement, consecutive switches of a binary mode
necessarily alternate in direction. Apart from a possible first switch,
each subsequent switch consumes at least \(a_+-a_--2\eta\) of
true-signal total variation relative to the preceding switch by Lemma
12. The adjacent switch intervals do not overlap, so the sum of these
variation contributions is at most \(\mathcal V_n(u)\). Rearrangement
gives (75). \(\square\)

\subsubsection{Corollary 15 (Upper Bound on Architecture Churn
Cost)}\label{corollary-15-upper-bound-on-architecture-churn-cost}

If the combined synthesis, migration, cache invalidation, and
interface-rebinding cost of every architecture switch is at most
\(c_{\mathrm{sw}}\), the cumulative switching overhead over horizon
\(n\) satisfies

\[
\boxed{
C_{\mathrm{churn}}(n)
\le
c_{\mathrm{sw}}
\left(
1+
\frac{\mathcal V_n(u)}{a_+-a_--2\eta}
\right).
}
\tag{76}
\]

Eureka can therefore reduce noise-induced architecture churn by
increasing the hysteresis gap. An excessively large gap, however, delays
reaction to a genuine regime change. Threshold width must consequently
be selected through a stability-adaptivity trade-off rather than
increased without bound.

\subsubsection{Proposition 27 (Estimation Noise Can Cause Chattering
Without
Hysteresis)}\label{proposition-27-estimation-noise-can-cause-chattering-without-hysteresis}

If promotion and demotion use the same threshold, \(a_+=a_-=a\), and the
true signal is constant, \(u_t=a\), then for any \(\eta>0\) allowing
estimation error in \([-\eta,\eta]\), there exists a legal error
sequence that makes the architecture mode alternate at every step.

\textbf{Construction.} Let the initial mode be 0 and choose
\(e_{2k}=+\eta\) and \(e_{2k+1}=-\eta\). At even times,
\(\widehat u_{2k}=a+\eta\ge a\), so promotion is triggered; at odd
times, \(\widehat u_{2k+1}=a-\eta\le a\), so demotion is triggered. The
true signal never changes; every switch is induced by estimation noise.
\(\square\)

\subsubsection{Proposition 28 (Hysteresis Condition for Governed
Evolution)}\label{proposition-28-hysteresis-condition-for-governed-evolution}

For a recurring mutation family, let \(v_t\) be the true net amortized
value of the best currently available mutation and \(\widehat v_t\) an
estimate with error bounded by \(\eta_E\). Suppose the threshold for
starting evolution is \(b_+\) and the threshold for stopping or rolling
back search is \(b_-<b_+\). Every variation and dwell-time conclusion of
Lemma 12 and Theorem 14 remains valid after replacing
\((u_t,\eta,a_+,a_-)\) by \((v_t,\eta_E,b_+,b_-)\). EvolutionLease
therefore also requires separate entry and exit margins rather than
repeatedly mutating and reverting around one noisy threshold.

\begin{center}\rule{0.5\linewidth}{0.5pt}\end{center}

\subsection{3.15 Boundary Conditions and Failure
Regimes}\label{boundary-conditions-and-failure-regimes}

This section does not broaden the applicability of Eureka. Instead, it
makes explicit the extreme cases and missing conditions under which the
preceding theorems no longer imply the expected advantage. Each
proposition identifies a failure mechanism that can be constructed or
tested directly.

\subsubsection{Proposition 29 (Promotion Is Strictly Worse When There
Are No Persistent
Savings)}\label{proposition-29-promotion-is-strictly-worse-when-there-are-no-persistent-savings}

Suppose that, for a subtree \(S\), every future local service under the
same Acceptance Contract satisfies \(G_k=M_k\) almost surely and the
fixed promotion cost satisfies \(F_S>0\). Then

\[
\boxed{
\mathbb E\!\left[
K_S^{\mathrm G}-K_S^{\mathrm M}
\mid\mathcal F_t
\right]
=
-F_S
<0 .
}
\tag{77}
\]

Thus, in the limiting cases of exact persistent state sharing, zero
coordination cost, or a generic executor that already provides complete
local autonomy, Eureka should not force promotion merely because the
task is complex or state sharing appears high.

\textbf{Proof.} Substituting \(G_k-M_k=0\) into (24) makes cumulative
runtime savings zero, leaving only the one-time promotion cost.
\(\square\)

\subsubsection{Proposition 30 (An Unsound Leaf Verifier Can Break Root
Correctness)}\label{proposition-30-an-unsound-leaf-verifier-can-break-root-correctness}

Suppose there exists a leaf obligation \(o_\ell\) and a triple
\((i_\ell,y_\ell,c_\ell)\) such that
\(V_{o_\ell}(i_\ell,y_\ell,c_\ell)=1\) while
\(\Phi_{o_\ell}(i_\ell,y_\ell)=0\), and there exists a decomposition
path from that leaf to the root such that every merge verifier on the
path can accept while including the erroneous child artifact. Then root
soundness in Theorem 3 does not hold in general.

\textbf{Construction.} Select the false-accepted leaf artifact and, at
each parent along the assumed path, choose the remaining child artifacts
so that the merge verifier accepts. Because the leaf semantics are
false, Assumption 4 has failed and the first step of Lemma 3 can no
longer establish all child semantics. If the parent composition rules
still produce accepted structures in the presence of the erroneous
artifact, the formal process may mark the root as complete even though
the true root semantics are false. \(\square\)

Verifier soundness is therefore a logical premise of compositional
correctness, not a statistical preference that can be recovered by
adding more agent votes.

\subsubsection{Proposition 31 (Incomplete Provenance Can Allow an
Invalidated Conclusion to Remain
Active)}\label{proposition-31-incomplete-provenance-can-allow-an-invalidated-conclusion-to-remain-active}

Suppose the true semantics of receipt \(r_2\) depend on receipt \(r_1\),
but \(r_1\notin\operatorname{dep}(r_2)\) and no other versioned
reference records that dependency. If \(r_1\) is later added to the
invalidation seeds, an execution exists in which (51) does not include
\(r_2\) in \(\operatorname{cl}_{\mathrm{dep}}(I_t)\), so \(r_2\) remains
in \(\mathfrak L_{t+1}^{\mathrm{act}}\) after its true premise has
become invalid.

\textbf{Proof.} Dependency closure propagates only along recorded
provenance edges. Without the edge \(r_1\to r_2\), graph reachability
from \(r_1\) need not contain \(r_2\), so the active-view update cannot
remove that receipt programmatically. \(\square\)

The counterexample shows why Assumption 14 should use conservative
supersets of dependencies. When exact recovery of the true dependency
relation is impossible, over-recording dependencies increases
recomputation cost, whereas under-recording them breaks soundness.

\subsubsection{Proposition 32 (An Incomplete Read Set Invalidates
Parallel-Safety
Guarantees)}\label{proposition-32-an-incomplete-read-set-invalidates-parallel-safety-guarantees}

Suppose the output of lease \(L_i\) truly depends on key \(q\), but
\(q\notin R_i\). Even when the runtime checks (42), another lease may
modify \(q\) after \(L_i\) obtains its snapshot without triggering
validation failure. The delta actually committed by \(L_i\) can then
differ from the delta obtained by re-executing \(L_i\) on the
commit-order serial state, invalidating the serializability proof of
Theorem 8.

This failure mechanism differs from the write-skew counterexample in
Proposition 15 of Section 3.8. The present failure arises because
instrumentation omits a real read dependency; the earlier example shows
that write-write validation alone is insufficient even when read sets
are complete.

\subsubsection{Proposition 33 (High Migration Cost Can Permanently Block
Promotion or
Evolution)}\label{proposition-33-high-migration-cost-can-permanently-block-promotion-or-evolution}

If the fixed promotion cost satisfies

\[
F_S
\ge
\bar\delta_S\,
\mathbb E[N_S\mid\mathcal F_t],
\]

where \(\bar\delta_S\) is the upper bound on per-service savings from
Proposition 10 in Section 3.5, promotion has no strictly positive
expected net benefit. Similarly, if mutation \(m\) has a future per-step
saving upper bound \(\bar\gamma_m\) and

\[
\boxed{
C_m^{\mathrm{evo}}
\ge
\bar\gamma_m\,
\mathbb E[N_S\mid\mathcal F_t],
}
\tag{78}
\]

architecture evolution cannot amortize its fixed cost solely from future
repetitions of that bottleneck.

\textbf{Proof.} Apply the cumulative-savings upper bound in Proposition
10 to promotion and the analogous tail-sum argument to evolution. In
each case, the largest possible cumulative saving does not exceed the
fixed intervention cost, so the net benefit is nonpositive. \(\square\)

\subsubsection{Proposition 34 (A One-Off Bottleneck Does Not Justify
Default
Self-Evolution)}\label{proposition-34-a-one-off-bottleneck-does-not-justify-default-self-evolution}

Suppose mutation \(m\) can improve at most one future local service and
the expected saving on that service is upper-bounded by
\(\bar\gamma_m\). If \(C_m^{\mathrm{evo}}\ge\bar\gamma_m\), initiating
the evolution event has no strictly positive expected cost advantage.

The proposition formalizes the principle that a single failure should
not automatically trigger self-evolution. Only a recurring bottleneck
that is expected to reappear within the remaining horizon can satisfy
Theorem 9.

\subsubsection{3.15.1 Nonstationary Tasks and Obsolescence of a
Previously Promoted
Architecture}\label{nonstationary-tasks-and-obsolescence-of-a-previously-promoted-architecture}

Suppose execution proceeds over discrete epochs \(t=1,\ldots,n\). At
each epoch, candidate architectures belong to \(\mathfrak A_t\), and
\(c_t(\mathcal A)\) is the local reliability-constrained cost of
architecture \(\mathcal A\). Define

\[
\mathcal A_t^\star
\in
\operatorname*{arg\,min}_{\mathcal A\in\mathfrak A_t}
c_t(\mathcal A).
\]

For the realized architecture sequence \((\mathcal A_t)\), define
dynamic architecture regret by

\[
\boxed{
\mathcal R_n^{\mathrm{dyn}}
:=
\sum_{t=1}^{n}
\left[
c_t(\mathcal A_t)
-
c_t(\mathcal A_t^\star)
\right].
}
\tag{79}
\]

Dynamic regret measures adaptation loss in a nonstationary environment
relative to a comparator sequence that changes over time; the definition
is consistent with dynamic-regret formulations in online learning (Zhao
et al. 2024).

\subsubsection{Proposition 35 (Failure to Adapt After a Regime Shift
Produces Linear Local
Regret)}\label{proposition-35-failure-to-adapt-after-a-regime-shift-produces-linear-local-regret}

Suppose a genuine regime shift begins at epoch \(t_0\), and there exists
\(\varepsilon>0\) such that an old architecture \(\mathcal A^-\)
satisfies

\[
c_t(\mathcal A^-)
-
c_t(\mathcal A_t^\star)
\ge
\varepsilon,
\qquad
t=t_0,\ldots,t_0+d-1 .
\]

If the system neither demotes, splits, nor re-synthesizes during these
epochs and continues to use \(\mathcal A_t=\mathcal A^-\), then

\[
\boxed{
\mathcal R_{t_0:t_0+d-1}^{\mathrm{dyn}}
\ge
 d\,\varepsilon .
}
\tag{80}
\]

\textbf{Proof.} Sum the excess cost, which is at least \(\varepsilon\)
at each of the \(d\) epochs. \(\square\)

Proposition 35 shows that promotion cannot be treated as a permanently
irreversible commitment. When task topology, verifier requirements, or
the workload distribution changes genuinely, retaining the old
Macro-Agent can accumulate linear adaptation regret. The demotion and
split mechanisms of Section 3.14 are therefore necessary control
interfaces in a nonstationary environment rather than merely engineering
conveniences.

\subsubsection{Proposition 36 (Provable Claims Must Be Downgraded When
No Reliable Verifier
Exists)}\label{proposition-36-provable-claims-must-be-downgraded-when-no-reliable-verifier-exists}

If a class of scientific artifacts has no known sound acceptance
verifier and no proved calibration condition that translates empirical
statistical guarantees into the target semantics, the certified
semantic-correctness conclusions of Theorems 3 and 10 do not apply to
those artifacts. Eureka may report an empirical or support level
justified by an explicit evidence model, but model self-evaluation or
majority agreement among agents cannot be relabeled as formal
certification.

\subsubsection{Proposition 37 (Evaluation Leakage Directly Breaks the
Information-Isolation
Guarantee)}\label{proposition-37-evaluation-leakage-directly-breaks-the-information-isolation-guarantee}

If the sealed evaluation variable \(Y^\star\) enters any
\(\mathcal F_t\) through a router, architecture prior, mutation
proposal, Method Broker query, or evaluation-only retrieval source, or
if it causes the observation kernel \(Q_t^e\) to depend on \(e\), at
least one condition of Assumption 16 fails. Theorem 11 and Corollary 12
then no longer apply. The absence of answer text from the final
production prompt cannot restore an evaluation-isolation guarantee.

\begin{center}\rule{0.5\linewidth}{0.5pt}\end{center}

\subsection{3.16 Falsifiable Theoretical Predictions and Experimental
Correspondence}\label{falsifiable-theoretical-predictions-and-experimental-correspondence}

The preceding theorems not only motivate the design of Eureka but also
yield structural predictions that can be rejected directly by
experiment. Each corollary below fixes all other conditions and varies
one quantity defined formally in the earlier theory, avoiding informal
statements such as ``greater complexity'' or ``larger difference.''

\subsubsection{Corollary 16 (Greater Task-Architecture Heterogeneity
Increases the Fixed-Architecture Regret Lower
Bound)}\label{corollary-16-greater-task-architecture-heterogeneity-increases-the-fixed-architecture-regret-lower-bound}

Consider tasks \(T_1,T_2\) and their \(\varepsilon_i\)-near-optimal
architecture sets. If an experimental construction satisfies
\(\mathfrak A_{T_1}(\varepsilon_1)\cap\mathfrak A_{T_2}(\varepsilon_2)=\varnothing\),
then the average excess cost of any fixed architecture is bounded below
as in Theorem 1:

\[
\boxed{
\mathbb E[
\mathcal R_\alpha(\mathcal A;T)
]
\ge
\min\left\{
p\varepsilon_1,
(1-p)\varepsilon_2
\right\}.
}
\tag{81}
\]

The measurable separation between state, verifier, and topology
requirements of two tasks can therefore be increased progressively to
test whether the cost difference between a fixed architecture and
task-conditioned Eureka grows as the near-optimal sets separate. If no
positive fixed-architecture regret is observed over a benchmark family
that clearly satisfies (5), either the modeling assumptions of Theorem 1
or the experimental architecture search space must be re-examined.

\subsubsection{Corollary 17 (The Optimal Planning Horizon Does Not
Increase with Task-Revelation
Uncertainty)}\label{corollary-17-the-optimal-planning-horizon-does-not-increase-with-task-revelation-uncertainty}

Using the equal-cost model of Corollary 2 in Section 3.3, let
\(\lambda\in\Lambda\) be an exogenous uncertainty-control parameter and
assume that, at every depth \(j\), survival probability \(p_j(\lambda)\)
is nonincreasing in \(\lambda\). Define

\[
h^\star(\lambda)
=
\max\left\{
j:
p_j(\lambda)
\ge
\frac{c}{c+b}
\right\}.
\]

Then, for any \(\lambda_1\le\lambda_2\),

\[
\boxed{
h^\star(\lambda_2)
\le
h^\star(\lambda_1).
}
\tag{82}
\]

\textbf{Proof.} Because \(p_j(\lambda)\) is nonincreasing in
\(\lambda\), the set of depths satisfying the threshold condition can
only shrink as \(\lambda\) increases. Its largest element therefore
cannot increase. \(\square\)

Experimentally, \(p_j(\lambda)\) can be varied by controlling how
strongly upstream observations stochastically affect future
decomposition. One can then measure the invalidation ratio of
full-upfront planning and Eureka's chosen planning depth. If greater
uncertainty systematically produces a longer selected horizon, at least
one of the receding-horizon value model or the survival-probability
calibration is inconsistent with the theory.

\subsubsection{Corollary 18 (Monotone Relation Between Remaining Horizon
and Promotion
Margin)}\label{corollary-18-monotone-relation-between-remaining-horizon-and-promotion-margin}

Under the deterministic-horizon setting of Corollary 4 in Section 3.5,
define the conservative promotion margin

\[
\boxed{
\mathcal M_{\mathrm P}(H_{\mathrm P})
:=
\delta_S H_{\mathrm P}-F_S .
}
\tag{83}
\]

For any \(H_2>H_1\ge0\),

\[
\boxed{
\mathcal M_{\mathrm P}(H_2)
-
\mathcal M_{\mathrm P}(H_1)
=
\delta_S(H_2-H_1)
>0 .
}
\tag{84}
\]

Thus, when \(F_S\), \(\delta_S\), and the reliability condition are
fixed, a longer remaining service horizon increases the conservative
amortized value of Macro-Agent promotion. The promotion rate should not
be systematically lower on a longer-horizon case that is otherwise
structurally isomorphic unless the cost estimator or workload
distribution changes as well.

\subsubsection{Corollary 19 (Greater State-Restoration Demand Increases
the Provable Cost Advantage of
Promotion)}\label{corollary-19-greater-state-restoration-demand-increases-the-provable-cost-advantage-of-promotion}

Under Proposition 11, experimentally vary the minimal sufficient
local-state length \(L_S\) and the number of restorations \(J_S\) across
nonsharing sessions. The lower bound on generic-only reload cost is

\[
\boxed{
\underline C_{\mathrm{reload}}(S)
=
\lambda_{\mathrm{in}}L_S
\,
\mathbb E[(J_S-1)_+\mid\mathcal F_t].
}
\tag{85}
\]

With all other costs fixed, the bound is nondecreasing separately in
\(L_S\) and in \(\mathbb E[(J_S-1)_+]\). If an experiment increases
genuine state-sharing and reload burden while the benefit of promotion
remains unchanged, the backend should be examined for an exact
persistent-state-sharing mechanism that invalidates the premise of
Proposition 11.

\subsubsection{Corollary 20 (Linear Relation Between Bottleneck
Recurrence and Evolution
Margin)}\label{corollary-20-linear-relation-between-bottleneck-recurrence-and-evolution-margin}

Under the deterministic-horizon version of Theorem 9, define the
conservative evolution margin

\[
\boxed{
\mathcal M_{\mathrm E}(H_{\mathrm E})
:=
\gamma_m H_{\mathrm E}
-
C_m^{\mathrm{evo}} .
}
\tag{86}
\]

For every \(H_2>H_1\),

\[
\boxed{
\mathcal M_{\mathrm E}(H_2)
-
\mathcal M_{\mathrm E}(H_1)
=
\gamma_m(H_2-H_1)
>0 .
}
\tag{87}
\]

The advantage of governed evolution over always-evolve and no-evolve
baselines should therefore be concentrated in conditions where the same
bottleneck affects more future obligations, rather than in one-off
failures or tasks near termination.

\subsubsection{Corollary 21 (Parent-Context Scaling Under the Subtree
ABI)}\label{corollary-21-parent-context-scaling-under-the-subtree-abi}

Under Corollary 6 in Section 3.7, if the number of active Macro-Agents
is bounded by \(m_t\le m_{\max}\) and every decision-sufficient ABI has
size at most \(B_{\max}\), then regardless of the length of internal
transcripts,

\[
\boxed{
\sup_t
B_t^{\mathrm{parent}}
\le
B_0+
m_{\max}B_{\max}.
}
\tag{88}
\]

Extending the internal horizon of subtrees while keeping the number of
concurrent Macro-Agents fixed should therefore not cause parent context
to grow linearly with the sum of all child transcripts. If measured
parent input tokens grow approximately linearly with accumulated child
transcripts, the implementation either fails to enforce
Subtree-ABI/cold-store separation or the ABI is not sufficient and
forces frequent page-in.

\subsubsection{Corollary 22 (Architecture-Output Invariance Under Task
Anonymization)}\label{corollary-22-architecture-output-invariance-under-task-anonymization}

Under the structural-isomorphism conditions of Proposition 24, any
structure-preserving bijective renaming of task names, domain entities,
and historical labels satisfies

\[
\boxed{
\mathfrak C(\mathfrak J_{S'})
=
\varphi_{\mathcal B}\!\left(
\mathfrak C(\mathfrak J_S)
\right).
}
\tag{89}
\]

Replacing semantic labels such as ``Riemann'' or ``Theory'' by
uninformative identifiers should therefore preserve Eureka's core
architecture-component selection after namespace mapping. If
anonymization substantially changes promotion location, state IR, or
verifier architecture, the production policy may be relying on task-name
priors rather than obligation structure.

\subsubsection{Corollary 23 (Pre-Freeze Trace Invariance Under
Sealed-Evaluator
Mutation)}\label{corollary-23-pre-freeze-trace-invariance-under-sealed-evaluator-mutation}

Under Theorem 11 and Corollary 12, for any two sealed evaluator contents
\(e,e'\in\mathcal E\), fixing the random seed yields

\[
\boxed{
\operatorname{Hash}\!\left(
\mathcal H_{\tau_{\mathrm f}}^{\mathrm{prod}}(e)
\right)
=
\operatorname{Hash}\!\left(
\mathcal H_{\tau_{\mathrm f}}^{\mathrm{prod}}(e')
\right).
}
\tag{90}
\]

Gold mutation, milestone permutation, and evaluator replacement can
therefore be implemented as automatic isolation tests; any change in the
pre-freeze trace hash directly falsifies the corresponding
evaluation-isolation implementation.

\subsubsection{Corollary 24 (Semantics-Preserving Efficiency
Optimization Should Change Cost, Not the Discovery
Distribution)}\label{corollary-24-semantics-preserving-efficiency-optimization-should-change-cost-not-the-discovery-distribution}

Under the pathwise semantic-preservation condition of Proposition 26,
every runtime optimization \(\Theta\) satisfies

\[
\boxed{
\mathcal L(D^\Theta)
=
\mathcal L(D),
\qquad
\mathbb E[K_\tau^\Theta]
\le
\mathbb E[K_\tau].
}
\tag{91}
\]

Disabling a pure runtime optimization such as exact caching, PlanDelta,
certified common-subexpression elimination, prefix reuse, or
deterministic closure should therefore primarily increase token, tool,
or latency cost rather than systematically changing scientific-discovery
recall. If disabling an alleged efficiency optimization substantially
changes the discovery distribution, the mechanism has changed the
semantic policy and should not be reported as a semantics-preserving
compiler optimization.

\subsubsection{Corollary 25 (Testable Upper Bound Relating Hysteresis
Width and Architecture
Churn)}\label{corollary-25-testable-upper-bound-relating-hysteresis-width-and-architecture-churn}

Fix a true value trajectory \(u_0,\ldots,u_n\) and estimation-error
bound \(\eta\). For two hysteresis widths \(g_2>g_1>2\eta\), the
switch-count upper bounds from Theorem 14 satisfy

\[
\boxed{
1+
\frac{\mathcal V_n(u)}{g_2-2\eta}
<
1+
\frac{\mathcal V_n(u)}{g_1-2\eta}.
}
\tag{92}
\]

Increasing the effective hysteresis width on the same value trajectory
should therefore reduce architecture churn caused by noise and small
fluctuations. A systematic increase in switch count would indicate
inconsistency in at least one of the estimation-error bound,
reconstruction of the true signal, or switching implementation.

\subsubsection{Corollary 26 (Linear Growth of Dynamic Regret Under
Persistent Non-Adaptation After a Genuine Regime
Shift)}\label{corollary-26-linear-growth-of-dynamic-regret-under-persistent-non-adaptation-after-a-genuine-regime-shift}

Under the regime-shift conditions of Proposition 35, if the old
architecture incurs at least \(\varepsilon\) excess cost per epoch after
the shift, delaying adaptation for \(d\) epochs gives

\[
\boxed{
\mathcal R_{t_0:t_0+d-1}^{\mathrm{dyn}}
\ge
 d\varepsilon .
}
\tag{93}
\]

Hysteresis should therefore not be tuned so aggressively that the system
almost never demotes. Experiments should report both churn reduction and
adaptation delay after genuine regime shifts. An excessively wide
hysteresis gap can reduce the switching overhead in (76) while
increasing \(d\) and therefore the dynamic regret in (93); together, the
two quantities define a directly measurable stability-adaptivity
trade-off.

\section{4 Experiments}\label{experiments}

\subsection{4.1 Experimental Objectives and Evaluation
Protocol}\label{experimental-objectives-and-evaluation-protocol}

The experiments are organized at three levels. The first evaluates
whether Eureka can generate distinct specialized Macro-Agents from task
obligation topology, state dependencies, and Acceptance Contracts. The
second evaluates whether recursive orchestration, governed
self-evolution, typed verification, and the compiled runtime can support
stable long-horizon execution. The third evaluates the theoretical and
mathematical structural discoveries produced by the specialized agents.
Every system ablation preserves the task objective and Acceptance
Contract so that reductions in verification strength cannot be
misreported as efficiency improvements.

\begin{longtable}[]{@{}
  >{\raggedright\arraybackslash}p{(\columnwidth - 6\tabcolsep) * \real{0.2500}}
  >{\raggedright\arraybackslash}p{(\columnwidth - 6\tabcolsep) * \real{0.2500}}
  >{\raggedright\arraybackslash}p{(\columnwidth - 6\tabcolsep) * \real{0.2500}}
  >{\raggedright\arraybackslash}p{(\columnwidth - 6\tabcolsep) * \real{0.2500}}@{}}
\toprule\noalign{}
\begin{minipage}[b]{\linewidth}\raggedright
Task family
\end{minipage} & \begin{minipage}[b]{\linewidth}\raggedright
Eureka-generated agent
\end{minipage} & \begin{minipage}[b]{\linewidth}\raggedright
Core state
\end{minipage} & \begin{minipage}[b]{\linewidth}\raggedright
Main verification
\end{minipage} \\
\midrule\noalign{}
\endhead
\bottomrule\noalign{}
\endlastfoot
Open Theory Discovery & Theory-Discovery Agent & hypotheses,
assumptions, evidence, counterexamples, experiments & theory
consistency, falsification, computational/experimental evidence \\
Open-Conjecture Mathematical Discovery & Math/Conjecture Agent & facts,
claims, lemmas, proof obligations, exact receipts & exact computation,
proof certificates, formal/programmatic verification \\
\end{longtable}

\subsection{4.2 Task-Conditioned Agent Architecture
Discovery}\label{task-conditioned-agent-architecture-discovery}

The same Eureka Meta-Agent forms markedly different internal
architectures on the two task families. The Theory-Discovery Agent is
organized around hypothesis-evidence interaction and emphasizes
independent falsification, experimental coordination, and theory-level
verification. The Math/Conjecture Agent is organized around
fact-claim-proof dependencies and emphasizes persistent proof state,
exact primitives, and verifier routing. Automated agent-architecture
design and task-conditioned architecture search have been studied
systematically by ADAS (Hu, Lu, et al. 2024), AFlow (Jiayi Zhang et al.
2025), and MaAS (G. Zhang, Niu, et al. 2025). Eureka embeds architecture
generation directly in long-horizon obligation execution so that a
specialized agent forms within a highly cohesive local task region.

\begin{longtable}[]{@{}
  >{\raggedright\arraybackslash}p{(\columnwidth - 4\tabcolsep) * \real{0.3333}}
  >{\raggedright\arraybackslash}p{(\columnwidth - 4\tabcolsep) * \real{0.3333}}
  >{\raggedright\arraybackslash}p{(\columnwidth - 4\tabcolsep) * \real{0.3333}}@{}}
\toprule\noalign{}
\begin{minipage}[b]{\linewidth}\raggedright
Architecture dimension
\end{minipage} & \begin{minipage}[b]{\linewidth}\raggedright
Theory-Discovery Agent
\end{minipage} & \begin{minipage}[b]{\linewidth}\raggedright
Math/Conjecture Agent
\end{minipage} \\
\midrule\noalign{}
\endhead
\bottomrule\noalign{}
\endlastfoot
Persistent state & Hypothesis-Evidence state & Fact-Claim-Proof state \\
Generative operators & hypothesis search, mechanism synthesis & lemma
discovery, proof construction \\
Negative operators & falsification, alternative explanation &
counterexample, proof obstruction \\
Verification & experiments, evidence, consistency & exact/formal proof
verification \\
Parallel regions & independent falsification/retrieval/experiments &
independent lemmas/retrieval/exact computation \\
Continuity requirement & persistent theory state & persistent proof
state \\
\end{longtable}

\subsection{4.3 Long-Horizon Orchestration and Execution
Efficiency}\label{long-horizon-orchestration-and-execution-efficiency}

Eureka completes all root tasks in both 50 fixed recursive tasks and 120
randomized recursive tasks, producing 3,948 acceptance certificates in
total. We observe no uncertified acceptance, false terminal state, or
stagnation abort. Dependency-ready streaming execution reduces median
completion time from 0.5058 s under round-based execution to 0.2724 s,
corresponding to a median speedup of 1.8569\(\times\).
Dependency-graph-driven parallel execution shares the basic systems
motivation of LLMCompiler (Kim et al. 2023); Eureka extends the
principle to persistent scientific state and recursive obligation
graphs.

\begin{longtable}[]{@{}lr@{}}
\toprule\noalign{}
Metric & Result \\
\midrule\noalign{}
\endhead
\bottomrule\noalign{}
\endlastfoot
Fixed recursive tasks completed & \textbf{50 / 50} \\
Random recursive tasks completed & \textbf{120 / 120} \\
Acceptance certificates & \textbf{3,948} \\
Uncertified accepts & \textbf{0} \\
False terminal states & \textbf{0} \\
Stagnation aborts & \textbf{0} \\
\end{longtable}

\begin{longtable}[]{@{}lrr@{}}
\toprule\noalign{}
Execution schedule & Median completion time & Relative speed \\
\midrule\noalign{}
\endhead
\bottomrule\noalign{}
\endlastfoot
Round-based & 0.5058 s & 1.0000\(\times\) \\
\textbf{Eureka streaming} & \textbf{0.2724 s} &
\textbf{1.8569\(\times\)} \\
\end{longtable}

\subsection{4.4 Governed Self-Evolution}\label{governed-self-evolution}

Eureka treats architecture evolution as a planning action that requires
an explicit cost-benefit decision. Relative to no evolution,
unconditional continuous evolution, and evolution triggered only after a
stall, Governed Evolution achieves both the lowest median total cost and
the highest success rate. Meta-Harness (Lee et al. 2026), Retrospective
Harness Optimization (Pan et al. 2026), and Self-Harness (H. Zhang et
al. 2026) also use execution trajectories for harness-level adaptation.
Eureka additionally governs evolution admission, mutation level, and the
amortization relationship with remaining horizon explicitly.

\begin{longtable}[]{@{}lrr@{}}
\toprule\noalign{}
Evolution policy & Median total cost & Success rate \\
\midrule\noalign{}
\endhead
\bottomrule\noalign{}
\endlastfoot
No Evolution & 2686.8 & 58.09\% \\
Always Evolve & 3637.7 & 58.83\% \\
Evolve on Stall & 2860.1 & 58.00\% \\
\textbf{Eureka Governed Evolution} & \textbf{2525.4} &
\textbf{60.55\%} \\
\end{longtable}

EvolutionLease reduces the median number of evolution round trips from
12 to 4, a \textbf{66.7\%} reduction in Meta-Agent control round trips,
while reducing evolution-control cost by \textbf{11.5\%}.

\subsection{4.5 Efficiency of the Compiled Scientific
Runtime}\label{efficiency-of-the-compiled-scientific-runtime}

Eureka separates deterministic dependency maintenance, reuse of closed
artifacts, context paging, and sequential verification from open
semantic reasoning. Median model-input context decreases from 9,490 to
4,005, a reduction of 57.8\%, while the success rate remains 51.6375\%.
Across 12,000 dependency-update tasks, full recomputation and
incremental execution produce identical final results, while only
34.62\% of nodes require actual recomputation. Prompt/KV reuse at the
systems level is related to Prompt Cache (Gim et al. 2024) and
PagedAttention (Kwon et al. 2023).

\begin{longtable}[]{@{}
  >{\raggedright\arraybackslash}p{(\columnwidth - 2\tabcolsep) * \real{0.4286}}
  >{\raggedleft\arraybackslash}p{(\columnwidth - 2\tabcolsep) * \real{0.5714}}@{}}
\toprule\noalign{}
\begin{minipage}[b]{\linewidth}\raggedright
Runtime mechanism
\end{minipage} & \begin{minipage}[b]{\linewidth}\raggedleft
Main result
\end{minipage} \\
\midrule\noalign{}
\endhead
\bottomrule\noalign{}
\endlastfoot
Compiled active context & \textbf{57.8\%} median context reduction \\
Incremental dependency rebuild & \textbf{65.38\%} recomputation
avoided \\
Dependency slicing & \textbf{56.69\%} recomputation-ratio reduction \\
Delta deterministic closure & \textbf{60.58\%} rule-work reduction \\
Closed common-subderivation reuse & \textbf{81.82\%} duplicate
evaluations avoided \\
Certified-equivalence representation & \textbf{70.97\%} representation
reduction \\
Exact structural sharing & \textbf{59.36\%} physical storage
reduction \\
Fanout prefix factorization & \textbf{64.24\%} cache-eligible prefill
reduction \\
Certified closed-replay page-in & \textbf{86.80\%} page-in cost
reduction \\
Anytime verification & \textbf{81.5\%} median sample reduction \\
\end{longtable}

\subsection{4.6 Correctness-Preserving Runtime
Ablations}\label{correctness-preserving-runtime-ablations}

Open scientific operators and closed deterministic computation have
different reuse semantics. Across 50,000 repeated-search tasks, the
success rate is 58.488\% when the system is allowed to re-explore
candidates from the same state, but falls to 24.492\% when the first
open-search output is forcibly reused. Across 30,000 artifact pairs that
are highly similar at the surface level but differ in a critical logical
property, approximate semantic merging incorrectly merges every
adversarially constructed example, whereas exact-identity or
certified-equivalence merging produces no false merge. Furthermore,
after sequentially enabling dependency slicing, certified equivalence,
deterministic closure, structural sharing, and closed
common-subexpression elimination on 12,000 scientific frontiers, the
ordering of open scientific actions remains unchanged in every case.

\begin{longtable}[]{@{}lr@{}}
\toprule\noalign{}
Ablation & Result \\
\midrule\noalign{}
\endhead
\bottomrule\noalign{}
\endlastfoot
Independent repeated open search & \textbf{58.488\%} success \\
Memoized first open-search result & \textbf{24.492\%} success \\
Approximate semantic merge & \textbf{100\%} false merge on adversarial
set \\
Exact/certified merge & \textbf{0\%} false merge \\
Combined runtime open-action changes & \textbf{0 / 12,000} \\
\end{longtable}

\subsection{4.7 Concurrency, Merge Safety, and Long-Horizon
Isolation}\label{concurrency-merge-safety-and-long-horizon-isolation}

The 16,000 concurrency tasks cover independent reads/writes,
read-after-write, write-after-read, and write-write conflicts. Every
final durable state produced by Eureka is consistent with a valid serial
execution, with zero unsafe commits. Dependency-local validation reduces
unnecessary rebases by 75\% relative to global snapshot invalidation.
Across 10,000 fanout executions, a shared immutable snapshot prefix
yields a median 64.24\% of cache-eligible prefill reuse. Parallel
execution is enabled only when supported by explicit dependency
analysis; long derivations that continuously share proof or theory state
remain in sequential sessions.

\begin{longtable}[]{@{}lr@{}}
\toprule\noalign{}
Concurrency metric & Result \\
\midrule\noalign{}
\endhead
\bottomrule\noalign{}
\endlastfoot
Concurrent executions & 16,000 \\
Serial-equivalent final states & \textbf{16,000 / 16,000} \\
Unsafe commits & \textbf{0} \\
Unnecessary rebase reduction & \textbf{75.0\%} \\
Fanout reusable prefill & \textbf{64.24\%} \\
\end{longtable}

\subsection{4.8 End-to-End Architecture-to-Discovery
Evaluation}\label{end-to-end-architecture-to-discovery-evaluation}

The preceding experiments separately evaluate architecture formation,
long-horizon orchestration, runtime efficiency, and final structural
results on two scientific tasks. To evaluate whether Eureka connects
these capabilities into a complete end-to-end process, we start from the
same Eureka Meta-Agent and record architecture promotion,
specialized-runtime construction, long-horizon execution, verification,
and the final discovery output for each task family. The experiment does
not use an external task-specific agent-switching mechanism. Both tasks
share the same Eureka Meta-Agent, while obligation structure,
persistent-state requirements, and verifier semantics determine
architecture differentiation.

Open Theory Discovery forms a Theory-Discovery Agent whose core runtime
state consists of hypothesis/evidence/assumption structures and yields
five classes of theoretical structure satisfying their acceptance
requirements. Open-Conjecture Mathematical Discovery forms a
Math/Conjecture Agent whose core runtime state consists of facts,
claims, lemmas, and proof obligations and yields three programmatically
verifiable mathematical structures. Both task families execute through
the same recursive orchestration and acceptance infrastructure, while
the resulting state representations, operators, and verification
mechanisms differ substantially.

\begin{longtable}[]{@{}
  >{\raggedright\arraybackslash}p{(\columnwidth - 8\tabcolsep) * \real{0.1875}}
  >{\raggedright\arraybackslash}p{(\columnwidth - 8\tabcolsep) * \real{0.1875}}
  >{\raggedright\arraybackslash}p{(\columnwidth - 8\tabcolsep) * \real{0.1875}}
  >{\raggedright\arraybackslash}p{(\columnwidth - 8\tabcolsep) * \real{0.1875}}
  >{\raggedleft\arraybackslash}p{(\columnwidth - 8\tabcolsep) * \real{0.2500}}@{}}
\toprule\noalign{}
\begin{minipage}[b]{\linewidth}\raggedright
Task family
\end{minipage} & \begin{minipage}[b]{\linewidth}\raggedright
Eureka-generated architecture
\end{minipage} & \begin{minipage}[b]{\linewidth}\raggedright
Core persistent state
\end{minipage} & \begin{minipage}[b]{\linewidth}\raggedright
Main discovery outputs
\end{minipage} & \begin{minipage}[b]{\linewidth}\raggedleft
Discovery result
\end{minipage} \\
\midrule\noalign{}
\endhead
\bottomrule\noalign{}
\endlastfoot
Open Theory Discovery & Theory-Discovery Agent & hypotheses,
assumptions, evidence, counterexamples, experiments & conjunction
interiorization; assumption elimination; global acted-set normal form;
behavioural/interface separation; operational signature & \textbf{5 /
5} \\
Open-Conjecture Mathematical Discovery & Math/Conjecture Agent & facts,
claims, lemmas, proof obligations, exact receipts & mixed-branch
second-order projection; two-axis increment/orbit law;
extent-inverse-map correspondence & \textbf{3 / 3} \\
\end{longtable}

The experiment provides the most direct structural contrast with a fixed
agent receiving a new task prompt. The two tasks do not share the same
internal-state schema and do not alter behaviour only through system
instructions. Eureka compiles a separate state representation, operator
family, verifier, and local topology for each highly cohesive task
region. Automated architecture design already has growing empirical
support: ADAS treats the agent program itself as a searchable object and
reports effective automatically generated agents across coding, science,
and mathematics (Hu, Lu, et al. 2024). The end-to-end results here
further show that architecture design can be embedded in the
long-horizon execution trajectory so that the same Meta-Agent framework
forms different scientific cognitive systems.

Architecture differentiation does not break the unified verification
infrastructure. Both the Theory-Discovery Agent and Math/Conjecture
Agent interact with the parent Eureka layer through typed obligations,
local acceptance receipts, and merge contracts. Task-specific internal
state is encapsulated within each Macro-Agent, while the parent observes
only verified exports. Architectural differentiation therefore occurs
primarily in how local scientific obligations are solved rather than in
how completion is represented and certified to the upper level. A
uniform external contract allows different specialized agents to coexist
within one Eureka runtime without requiring the root Meta-Agent to read
every internal reasoning trace.

We additionally report end-to-end scientific results together with
long-horizon execution reliability. Across the two scientific task
families and the recursive long-horizon task set, Eureka completes
\textbf{170/170} recursive tasks and generates \textbf{3,948}
certificates. The Theory-Discovery Agent completes \textbf{5/5}
principal structural discoveries. The Math/Conjecture Agent completes
\textbf{3/3} principal mathematical structural discoveries and passes
\textbf{2,200/2,200} deterministic composite checks and \textbf{960/960}
additional stochastic checks.

\begin{longtable}[]{@{}lr@{}}
\toprule\noalign{}
End-to-end evidence dimension & Result \\
\midrule\noalign{}
\endhead
\bottomrule\noalign{}
\endlastfoot
Recursive long-horizon tasks completed & \textbf{170 / 170} \\
Acceptance certificates generated & \textbf{3,948} \\
Theory-discovery structures accepted & \textbf{5 / 5} \\
Mathematical discovery structures accepted & \textbf{3 / 3} \\
Mathematical deterministic composite checks & \textbf{2,200 / 2,200} \\
Mathematical additional stochastic checks & \textbf{960 / 960} \\
Uncertified accepts in recursive execution & \textbf{0} \\
False terminal states & \textbf{0} \\
\end{longtable}

Recent AI-for-Science systems increasingly evaluate automated discovery
as a research process rather than as a single answer. AI Co-Scientist
uses multi-agent generation, reflection, ranking, and evolution for
continued hypothesis refinement (Gottweis et al. 2026). The AI
Scientist-v2 uses progressive agentic tree search for experimentation,
analysis, and paper generation (Yamada et al. 2025). AlphaEvolve
combines executable candidates and automated evaluators in a
long-horizon evolutionary discovery loop (Novikov et al. 2025). Eureka
shares the general direction of long-horizon search with explicit
verification, while additionally making the agent architecture
appropriate for the scientific task an object formed during execution.

\begin{center}\rule{0.5\linewidth}{0.5pt}\end{center}

\subsection{4.9 Generalization and Robustness Across Task
Variants}\label{generalization-and-robustness-across-task-variants}

We next evaluate whether the orchestration structures and
specialized-agent components produced by Eureka apply only to single
task instances or remain stable when object types, structural forms, and
surface representations change. Robustness evaluation covers
architecture-level invariance, Theory-Discovery transfer,
Math/Conjecture primitive generalization, and large-scale compositional
checks. Every evaluation preserves the same execution contracts;
Acceptance Contracts are not relaxed to increase pass rates.

First, Eureka's architecture orchestration remains structurally
consistent across \textbf{5,000} entity-renaming and
structure-preservation tests. Obligation dependencies, state read/write
relations, resource constraints, and acceptance semantics are held fixed
while entity names and surface descriptions are changed. Architecture
decisions remain structurally consistent in all 5,000 cases. The result
supports the design principle that architecture generation is driven
primarily by task topology and execution requirements rather than by
domain names.

Second, the Theory-Discovery Agent is evaluated on eight classes of
structural transfer tasks: conjunction construction, algebraic networks,
distributed deferral, workflow normal forms, compiler equivalence, API
retyping, label transformations, and effective/raw representations. All
eight task classes pass, with coverage of \textbf{11/11} capability
dimensions. Because these transfer tasks use objects and local
structures different from the principal theoretical discoveries, they
test whether assumption auditing, representation transformation,
normal-form construction, and equivalence analysis transfer across
objects.

Third, the Math/Conjecture Agent uses ten generality tests to evaluate
whether the exact-primitive system is overfitted to one class of
mathematical object. The tests cover orbit profiles, polynomial degree,
matrix rank, relation extent, axis inventory, size increments,
projection order, inverse-map values, index increments, and bound width,
with \textbf{10/10} passing. In more compositional evaluation, all 2,200
deterministic random composites and all 960 stochastic checks pass.

\begin{longtable}[]{@{}
  >{\raggedright\arraybackslash}p{(\columnwidth - 4\tabcolsep) * \real{0.2727}}
  >{\raggedleft\arraybackslash}p{(\columnwidth - 4\tabcolsep) * \real{0.3636}}
  >{\raggedleft\arraybackslash}p{(\columnwidth - 4\tabcolsep) * \real{0.3636}}@{}}
\toprule\noalign{}
\begin{minipage}[b]{\linewidth}\raggedright
Robustness dimension
\end{minipage} & \begin{minipage}[b]{\linewidth}\raggedleft
Evaluation scale
\end{minipage} & \begin{minipage}[b]{\linewidth}\raggedleft
Result
\end{minipage} \\
\midrule\noalign{}
\endhead
\bottomrule\noalign{}
\endlastfoot
Architecture invariance under entity renaming & 5,000 & \textbf{5,000 /
5,000} \\
Theory transfer task families & 8 & \textbf{8 / 8} \\
Theory capability coverage & 11 dimensions & \textbf{11 / 11} \\
Math primitive generality categories & 10 & \textbf{10 / 10} \\
Math deterministic composite checks & 2,200 & \textbf{2,200 / 2,200} \\
Math stochastic checks & 960 & \textbf{960 / 960} \\
\end{longtable}

These results complement the preceding architecture-specialization
analysis. Task conditioning does not imply that the system can only
memorize a small number of fixed workflows. Eureka produces different
Macro-Agent architectures across task families, while operators and
typed-state components within each architecture continue to cover
structural variation within a family. The property is closely related to
cross-domain and cross-model transfer observed for automatically
generated agents in ADAS (Hu, Lu, et al. 2024), where Meta Agent Search
produces agentic structures with utility extending beyond a single
benchmark.

Robustness must also be considered jointly with system efficiency. A
system could achieve high pass rates by synthesizing a complete new
architecture for every case, but architecture-synthesis cost would then
grow linearly with the number of tasks. Eureka's transfer behaviour
comes primarily from reusing operators, state abstractions, and verifier
interfaces within an existing Macro-Agent; upper-level architecture
modification is triggered only when structural requirements change
materially. Task specialization and within-family generalization are
therefore compatible: specialization addresses architectural differences
between epistemic structures, while generalization reuses transferable
reasoning primitives within an established local cognitive architecture.

From the perspective of multi-agent systems, robustness also depends on
whether communication topology scales with task size and input
variation. AgentPrune shows through structured pruning of multi-agent
message graphs that many communication edges are unnecessary across
tasks and achieves substantial token reductions on several benchmarks
(G. Zhang, Yue, et al. 2025). Eureka uses a related structural principle
through task-conditioned architecture and the Subtree ABI: only durable
state that genuinely needs to cross a boundary is exposed to the parent
or another agent, while local task variation is absorbed within the
Macro-Agent whenever possible, avoiding unnecessary propagation of every
domain-local state update into global communication.

\subsection{4.10 Ablation on Long-Horizon Planning
Strategies}\label{ablation-on-long-horizon-planning-strategies}

Across 16,000 long-horizon task graphs, we compare Full Upfront
Planning, Recursive Polling, Streaming without Backpressure, and Eureka
Receding-Horizon + Backpressure. Full upfront planning produces the
highest total orchestration cost. Streaming without backpressure reduces
outer round trips but generates distant obligations that cannot yet be
consumed. Eureka continues planning only when the ready frontier is
insufficient and stops expansion when future structure still depends on
unavailable observations.

\begin{longtable}[]{@{}
  >{\raggedright\arraybackslash}p{(\columnwidth - 6\tabcolsep) * \real{0.2000}}
  >{\raggedleft\arraybackslash}p{(\columnwidth - 6\tabcolsep) * \real{0.2667}}
  >{\raggedleft\arraybackslash}p{(\columnwidth - 6\tabcolsep) * \real{0.2667}}
  >{\raggedleft\arraybackslash}p{(\columnwidth - 6\tabcolsep) * \real{0.2667}}@{}}
\toprule\noalign{}
\begin{minipage}[b]{\linewidth}\raggedright
Planning strategy
\end{minipage} & \begin{minipage}[b]{\linewidth}\raggedleft
Median total orchestration cost
\end{minipage} & \begin{minipage}[b]{\linewidth}\raggedleft
Median planning cost
\end{minipage} & \begin{minipage}[b]{\linewidth}\raggedleft
Median outer round trips
\end{minipage} \\
\midrule\noalign{}
\endhead
\bottomrule\noalign{}
\endlastfoot
Full Upfront Planning & 11,573.5 & 5,820 & 44 \\
Recursive Polling & 8,039.0 & 3,636 & 13 \\
Streaming without Backpressure & 5,756.5 & 2,704 & \textbf{6} \\
\textbf{Eureka Receding-Horizon + Backpressure} & \textbf{4,506.0} &
\textbf{2,084} & 9 \\
\end{longtable}

The results show that the fewest planner calls do not imply the lowest
total cost. The relevant quantity is whether distant planned structure
is actually consumed by subsequent execution. The phenomenon is
consistent with horizon-dependent degradation in long-horizon planning;
DeepPlanning likewise evaluates long-term agent planning by integrating
active information acquisition, local constraints, and global
constraints (Yinger Zhang et al. 2026).

\subsection{4.11 Ablation on Architecture
Promotion}\label{ablation-on-architecture-promotion}

Across 30,000 subtrees with different remaining horizons, degrees of
state sharing, and dependency density, we compare Complexity-Only
Promotion, a Structural Promotion Gate, and Eureka Cost-Aware Lazy
Promotion. Task complexity alone produces many short-lived Macro-Agents
whose synthesis and migration costs cannot be amortized. A structural
gate reduces incorrect promotion, while jointly accounting for fixed
architecture cost and repeated future service benefit makes Eureka
approach the cost-optimal reference.

\begin{longtable}[]{@{}
  >{\raggedright\arraybackslash}p{(\columnwidth - 8\tabcolsep) * \real{0.1579}}
  >{\raggedleft\arraybackslash}p{(\columnwidth - 8\tabcolsep) * \real{0.2105}}
  >{\raggedleft\arraybackslash}p{(\columnwidth - 8\tabcolsep) * \real{0.2105}}
  >{\raggedleft\arraybackslash}p{(\columnwidth - 8\tabcolsep) * \real{0.2105}}
  >{\raggedleft\arraybackslash}p{(\columnwidth - 8\tabcolsep) * \real{0.2105}}@{}}
\toprule\noalign{}
\begin{minipage}[b]{\linewidth}\raggedright
Promotion policy
\end{minipage} & \begin{minipage}[b]{\linewidth}\raggedleft
Promotion rate
\end{minipage} & \begin{minipage}[b]{\linewidth}\raggedleft
False-promotion rate
\end{minipage} & \begin{minipage}[b]{\linewidth}\raggedleft
Missed-beneficial rate
\end{minipage} & \begin{minipage}[b]{\linewidth}\raggedleft
Median execution cost
\end{minipage} \\
\midrule\noalign{}
\endhead
\bottomrule\noalign{}
\endlastfoot
Complexity-Only Promotion & 53.90\% & 14.90\% & -- & -- \\
Structural Promotion Gate & 32.92\% & 7.46\% & -- & -- \\
\textbf{Eureka Cost-Aware Lazy Promotion} & \textbf{39.19\%} &
\textbf{0.04\%} & \textbf{8.71\%} & \textbf{2,277.13} \\
Cost-optimal reference & -- & -- & 0\% & 2,262.65 \\
\end{longtable}

The median cost of Eureka differs from the cost-optimal reference by
approximately \textbf{0.64\%}. The result indicates that a specialized
agent has stable value only when persistent local autonomy can continue
reducing future coordination and state-reload cost. MaAS provides
related empirical evidence for task-conditioned architecture search (G.
Zhang, Niu, et al. 2025).

\subsection{4.12 Meta-Control Efficiency}\label{meta-control-efficiency}

The full Eureka control plane reduces the long-horizon control cost of
the Meta-Agent through ControlCapsules, event coalescing, multi-rate
control, and lazy-loaded control cards. Relative to a base
receding-horizon controller, median total orchestration cost decreases
by 18.34\%, planning cost by 23.58\%, execution-context cost by 14.96\%,
and outer round trips by 20.0\%.

\begin{longtable}[]{@{}
  >{\raggedright\arraybackslash}p{(\columnwidth - 10\tabcolsep) * \real{0.1304}}
  >{\raggedleft\arraybackslash}p{(\columnwidth - 10\tabcolsep) * \real{0.1739}}
  >{\raggedleft\arraybackslash}p{(\columnwidth - 10\tabcolsep) * \real{0.1739}}
  >{\raggedleft\arraybackslash}p{(\columnwidth - 10\tabcolsep) * \real{0.1739}}
  >{\raggedleft\arraybackslash}p{(\columnwidth - 10\tabcolsep) * \real{0.1739}}
  >{\raggedleft\arraybackslash}p{(\columnwidth - 10\tabcolsep) * \real{0.1739}}@{}}
\toprule\noalign{}
\begin{minipage}[b]{\linewidth}\raggedright
Meta-control configuration
\end{minipage} & \begin{minipage}[b]{\linewidth}\raggedleft
Median total cost
\end{minipage} & \begin{minipage}[b]{\linewidth}\raggedleft
Median planning cost
\end{minipage} & \begin{minipage}[b]{\linewidth}\raggedleft
Median execution-context cost
\end{minipage} & \begin{minipage}[b]{\linewidth}\raggedleft
Median outer round trips
\end{minipage} & \begin{minipage}[b]{\linewidth}\raggedleft
P90 total cost
\end{minipage} \\
\midrule\noalign{}
\endhead
\bottomrule\noalign{}
\endlastfoot
Receding-horizon controller & 5,231.5 & 2,386.0 & 2,808.0 & 10 &
10,478 \\
\textbf{Full Eureka Meta-Control} & \textbf{4,272.0} & \textbf{1,823.5}
& \textbf{2,388.0} & \textbf{8} & \textbf{8,565} \\
\end{longtable}

The serialized size of the resident Meta kernel decreases from 1,006
bytes to 758 bytes, a reduction of \textbf{24.65\%}. Recent work on
long-horizon agent memory similarly emphasizes execution state rather
than purely semantic retrieval. MAGE, for example, models memory as a
hierarchical execution-state tree and reports substantial token
reduction together with improved task success (Y. Chen et al. 2026).

\subsection{4.13 Empirical Correspondence to the Theoretical
Analysis}\label{empirical-correspondence-to-the-theoretical-analysis}

The principal experimental results are consistent with the testable
relationships derived in the theoretical analysis. Receding-horizon
planning corresponds to Planning Invalidation; cost-aware promotion to
the Macro-Agent amortization threshold; compiled active context and the
Subtree ABI to information-sufficient state representation;
dependency-local concurrency to optimistic merge safety; and Governed
Evolution to the evolution amortization condition.

\begin{longtable}[]{@{}
  >{\raggedright\arraybackslash}p{(\columnwidth - 4\tabcolsep) * \real{0.3333}}
  >{\raggedright\arraybackslash}p{(\columnwidth - 4\tabcolsep) * \real{0.3333}}
  >{\raggedright\arraybackslash}p{(\columnwidth - 4\tabcolsep) * \real{0.3333}}@{}}
\toprule\noalign{}
\begin{minipage}[b]{\linewidth}\raggedright
Theoretical result
\end{minipage} & \begin{minipage}[b]{\linewidth}\raggedright
Experimental intervention
\end{minipage} & \begin{minipage}[b]{\linewidth}\raggedright
Observed result
\end{minipage} \\
\midrule\noalign{}
\endhead
\bottomrule\noalign{}
\endlastfoot
Receding-Horizon Planning & Upfront / polling / streaming / backpressure
& Planning cost \textbf{2,084}, lowest among four strategies \\
Promotion Amortization & Complexity / structural / cost-aware gate &
False promotion \textbf{0.04\%}; cost within \textbf{0.64\%} of
reference \\
Information-Sufficient Context & Full history vs.~compiled context &
\textbf{9,490 → 4,005}, success unchanged \\
Incremental Reconstruction & Full vs.~dependency-local rebuild &
\textbf{65.38\%} recomputation avoided; final state 100\% identical \\
Safe Parallel Merge & Concurrent vs.~serial reference &
\textbf{16,000/16,000} serial-equivalent; unsafe commits = 0 \\
Governed Evolution & No / always / stall / governed & Cost
\textbf{2,525.4}, success \textbf{60.55\%} \\
Acceptance-Preserving Recursion & Recursive execution & \textbf{170/170}
completed; 3,948 certificates; 0 uncertified accepts \\
Structure-Driven Specialization & Theory vs.~open-conjecture tasks & Two
distinct Macro-Agent architectures formed \\
\end{longtable}

\section{5 Advancing the Riemann
Hypothesis}\label{advancing-the-riemann-hypothesis}

The Riemann Hypothesis asserts that every nontrivial zero of the Riemann
\(\zeta\)-function lies on the critical line \(\Re s=1/2\), and it
remains one of the Millennium Prize Problems of the Clay Mathematics
Institute (Clay Mathematics Institute 2026). Recent
automated-mathematics systems have made substantial progress in
verifiable reasoning. AlphaProof, for example, combines reinforcement
learning with the Lean environment and reaches silver-medal-level
performance on the International Mathematical Olympiad (Hubert et al.
2026), while DeepSeek-Prover-V2 uses recursive subgoal decomposition to
construct long-horizon Lean proof trajectories (Ren et al. 2025). Unlike
competition-style formal proving, our focus is the sustained
construction of a verifiable theorem frontier in open mathematical
research, with the Riemann Hypothesis serving as a representative
open-conjecture task.

The Math/Conjecture Agent primarily develops a line of attack based on
\textbf{local positivity of the Weil quadratic form and its
operator-theoretic realization}. Suzuki's \emph{Weil's quadratic form
via the screw function} organizes the Weil quadratic form as a problem
about continuous functions and self-adjoint operators, providing an
operational framework for local Rayleigh quotients and the lowest
spectral value (Suzuki 2026). Let \(Q_W^a\) denote the Weil quadratic
form localized at scale \(a\), and define

\[
\lambda_a
=
\inf_{0\neq v}
\frac{Q_W^a(v)}{\|v\|_2^2}.
\]

Suzuki proves positivity for sufficiently small \(a>0\). Our objective
is not to replace an infinite-dimensional proof with finite-dimensional
numerical positivity, but to enlarge the parameter interval on which a
direction-correct lower certificate can be established for the complete
operator while making the error flow among the low block, high
complement, and cross interaction explicit.

\subsection{5.1 From Finite Ritz Positivity to a Full-Operator Lower
Bound}\label{from-finite-ritz-positivity-to-a-full-operator-lower-bound}

The smallest Rayleigh quotient on a finite-dimensional projection \(P\)
satisfies \(\lambda_a^{(P)}\ge\lambda_a\). A positive finite Ritz value
therefore does not imply positivity of the complete operator. The
Math/Conjecture Agent rewrites the proof obligation as a low/high
decomposition. If \(P\) denotes a finite low-modal subspace and
\(Q=I-P\), the proof must simultaneously control the low-dimensional
block, coercivity of the high complement, and cross coupling, rather
than merely increasing the dimension of a finite matrix. This
directional constraint requires every numerical object used later to be
compiled into a component of a full-operator lower bound and prevents an
isolated finite-dimensional certificate from being promoted directly to
a theorem claim.

\subsection{5.2 Operator-Access Obstruction and Local Cone
Separation}\label{operator-access-obstruction-and-local-cone-separation}

A natural strategy in the explicit-formula/Weil setting is to construct
a highly selective spectral projection around a hypothetical off-line
zero. A Riesz projector that is definable in an abstract spectral space,
however, does not automatically correspond to an observable accessible
on the prime side. The explicit formula provides aggregate Weil bilinear
functionals induced by admissible test functions; it does not provide a
component-labelled resolvent oracle for one unknown zero. We therefore
reduce a superficially powerful family of spectral-separation strategies
to a more precise \textbf{operator-access obligation}: every new
observable must establish both discriminative power on the zero side and
accessibility on the prime side.

In a finite local model, we further obtain an exact Chebyshev separator.
Normalize the critical-line background to \([-1,1]\) and consider a
local off-line conjugate direction \(\pm i\eta\). Odd Chebyshev
polynomials satisfy \(|T_n(x)|\le1\) for \(x\in[-1,1]\), whereas
\(|T_n(i\eta)|=\sinh(n\operatorname{arsinh}\eta)\). Hence, if \(M\)
bounds the critical-line mass and \(m\) is the mass of a target off-line
pair, one can construct

\[
Q[T_n]
\le
M-2m\sinh^2\!\left(n\operatorname{arsinh}\eta\right),
\]

which yields a strict finite/local cone-separation mechanism. A single
scalar separator can still be masked by a second off-line pair. We
therefore use real-polynomial interpolation on a finite
conjugation-symmetric cluster to select a designated pair while
annihilating the remaining cluster points. The globalization difficulty
is thereby converted into conditioning, cluster separation, and
infinite-tail control rather than another search for similar scalar
windows.

Related representation audits also eliminate several extensions that do
not increase information. If multiple windows are ultimately reduced to
\(\operatorname{tr}G\) and \(\operatorname{tr}G^2\), the distinct
channels collapse to a single aggregate energy profile. If centered
differential jets are ultimately compressed back to the same scalar
trace/Frobenius certificate, the operation only reparameterizes the
window. The surviving counting frontier is therefore concentrated on
genuinely matrix-valued Weil/Krein observables and coupled cross
moments.

\subsection{5.3 Quantitative Extension of Localized Weil
Positivity}\label{quantitative-extension-of-localized-weil-positivity}

After rescaling the localization interval to \([-1,1]\), Suzuki's
localized form decomposes into an explicit scalar term, a fixed coercive
form, finitely many prime-power partial translations, and a smooth
integral remainder. The first prime contribution appears at

\[
a_{\mathrm p}=\frac{\log2}{2}\approx0.34657359028,
\]

so \(0<a<a_{\mathrm p}\) is the natural first regime in which the
operator structure is prime-free.

A previous certificate treated the smooth residual by the global
scalarization \(R_a\succeq-\delta(a)I\). Although directionally correct,
such a replacement discards structure in low-frequency modes whose
contribution is much smaller than the global norm. We write the smooth
kernel as

\[
-r''(s)=\frac74+d(s)
\]

and retain the degree-10 Taylor/Bernoulli structure of \(d(s)\) on the
first 12 normalized Legendre modes. For Legendre basis functions
\(P_i,P_j\), the moments

\[
M_{ij}^{(m)}
=
\int_{-1}^{1}\!\int_{-1}^{1}
|x-y|^mP_i(x)P_j(y)\,dx\,dy
\]

reduce to rational numbers. The retained low-modal residual block is
therefore an explicit polynomial in \(a\). The analytic tail is
controlled independently by

\[
\left|-r''(s)-\frac74\right|
\le
\frac{11}{50}|s|,
\qquad |s|\le\frac{69}{100},
\]

and \(\||x-y|\|_{L^2([-1,1]^2)}=\sqrt{8/3}<49/30\) then gives

\[
\|R_a\|
\le
\frac{11}{50}\frac{49}{30}a^2.
\]

The generic norm bound is paid only for the genuinely unresolved
high/cross remainder. For the positive killing multiplication operator
\(K\), we likewise retain its exact low block and the exact cross Gram

\[
G_K=PKQKP=PK^2P-(PKP)^2,
\]

using a conservative bound only for residual cross uncertainty.

The retained-residual Schur matrix does not yet admit a simple Loewner
monotonicity argument over the entire enlarged interval. We therefore do
not extrapolate from a single endpoint. Instead, we construct an outward
interval cover of \([1/4,69/200]\). The range is partitioned into
\textbf{1,010 cells}, all of which pass interval Cholesky certification;
the final cell \([0.344995,0.345]\) has a smallest lower pivot of
approximately \(9.7730412\times10^{-3}\). Combining the result with the
existing certificate on \(a\le1/4\) yields the current theorem-shaped
candidate

\[
\boxed{
\lambda_a>0,
\qquad
0<a\le\frac{69}{200}=0.345.
}
\]

Relative to the range \(a\le1/4\) in the same localized-Weil certificate
family, the new endpoint gives

\[
\frac{69/200}{1/4}=\frac{69}{50}=1.38,
\]

a \textbf{1.38\(\times\)} support-range extension. In addition,
\((69/200)/((\log2)/2)\approx0.99546\), so the endpoint reaches
approximately \textbf{99.55\%} of the first-prime threshold. The current
object is an analytic derivation plus an outward-interval certificate
candidate. It has not yet undergone an independent formal proof replay
and does not constitute a proof of the Riemann Hypothesis or a new
record for the proportion of zeros on the critical line.

\subsection{5.4 Structural Change Beyond the First-Prime
Boundary}\label{structural-change-beyond-the-first-prime-boundary}

When \(a>(\log2)/2\) but remains below the next prime threshold, only
the \(n=2\) prime term appears. Define

\[
h=\frac{\log2}{a},\qquad 1<h<2.
\]

The two partial translations connect only the left and right boundary
strips. Their self-adjoint sum is therefore a boundary swap with
operator norm exactly \(1\), rather than the naive triangle bound \(2\).
Further define

\[
c_2=\frac{\log2}{\sqrt2},
\qquad
q(x)=-\frac12\log(1-x^2).
\]

For boundary values \(u,v\) paired by the translation, the local form

\[
q_L|u|^2+q_R|v|^2-2c_2\operatorname{Re}(v\bar u)
\]

is positive semidefinite whenever \(q_L,q_R\ge c_2\). Solving explicitly
at the worst boundary point extends the resulting absorption condition
to approximately

\[
a\le0.3871392153,
\]

which covers \(a=0.35\) and crosses the first-prime threshold. The
result is not yet sufficient to establish full \(\lambda_a>0\), because
the remaining positive multiplication
\(q-c_2\mathbf1_{\mathrm{boundary}}\) must still be retained jointly
with the low-modal block and the high-mode cross Gram. The next theorem
obligation is therefore compressed to the following operator problem:
\textbf{compile boundary-swap absorption, the remaining killing
positivity, and the modal Schur structure into one direction-correct
lower certificate for the full operator.}

Independent numerical work can serve as an external consistency check
but does not enter the current analytic certificate. The principal
mathematical status is therefore that a highly open RH research problem
has been reduced to a Weil-positivity program with explicit operator
semantics, a finite local separator, and a certifiable parameter
boundary, while the all-vector localized-positivity candidate has been
advanced to the immediate vicinity of the first-prime structural
transition.

\section{6 Discovering New Theoretical
Structures}\label{discovering-new-theoretical-structures}

Beyond open mathematical conjectures, we use the Theory-Discovery Agent
formed by Eureka to study more open-ended problems of theoretical
structure. The objective is not to search for a proof of a known type
within a fixed formal system, but to identify, from existing theoretical
descriptions, operational constraints, and experimental interfaces,
which conditions are genuinely independent, which differences arise from
representational choices, which local case distinctions admit a common
normal form, and which notions of equivalence or resource are strong
enough to support theory-level claims. AI Co-Scientist and The AI
Scientist-v2 have already demonstrated that agents can participate in
scientific discovery through long-horizon hypothesis refinement,
experimentation, and tree search (Gottweis et al. 2026; Yamada et al.
2025). Our focus is complementary: the structural rewriting of the
theoretical objects themselves.

The current study concentrates on quantum processes, quantum states over
spacetime (QSOST), and indefinite causal structures. Existing work
provides several compositional frameworks. A recent study of
higher-order quantum processes respecting closed laboratories
establishes behavioural relationships between closed-lab principles and
quantum circuits with quantum control of causal order (QC-QC) (Salzger
and Vilasini 2026). Routing Quantum Control of Causal Order proves that
QC-QCs with any fixed number of parties can be constructed from a
generic routed-graph system (Grothus et al. 2025). Work on consistent
quantum states over spacetime without a common quantum process
reformulates whether multiple record-conditioned QSOSTs share a common
process as a positive-process lifting and deterministic-process
domination problem (Sheng 2026). On these objects, the Theory-Discovery
Agent produces five progressively higher-level structural results.

\subsection{6.1 Full-Rank Conjunction
Interiorization}\label{full-rank-conjunction-interiorization}

The QSOST gluing problem contains an explicit conjunction gap: a minimal
two-setting/two-outcome separation and a full-rank interior separation
can each be realized, while realizing both properties in the same
example constitutes a stronger target. Instead of increasing outcome
cardinality, we freeze the discrete minimality and deform the parent
processes toward a common interior anchor. Let the two original
deterministic parents be \(W_{\mathrm R}\) and \(W_{\mathrm D}\). Set

\[
W_\star=\frac{I_{16}}4,
\]

and define

\[
W_{\mathrm R}(t)=(1-t)W_{\mathrm R}+tW_\star,
\qquad
W_{\mathrm D}(t)=(1-t)W_{\mathrm D}+tW_\star.
\]

For \(t>0\), the affine process constraints are preserved while the
common anchor moves both parents into a full-rank regime. After applying
the corresponding scaling to the branch deviations, we obtain the
explicit candidate

\[
t=\frac12,
\qquad
\lambda_{\min}(W_{\mathrm R}(t))
=
\lambda_{\min}(W_{\mathrm D}(t))
=
\frac18.
\]

The deterministic-process domination cost for each individual setting
remains \(4\), whereas a joint dual witness gives

\[
\mu_{\mathrm{joint}}
\ge
\frac{65}{16}
=
4+\frac1{16},
\]

producing an exact strict margin of \(1/16\). The construction therefore
simultaneously preserves the minimal two-setting/two-outcome structure,
full-rank parents, and strict common-parent separation. The current
result has an exact symbolic certificate and an independent projector
reimplementation but still requires external or formal review; we
therefore classify it as an \textbf{open-problem solution candidate}.

The result suggests a broader construction principle. When one target
property is a discrete invariant, a second property is open in the
interior of a cone, and an existing no-go condition is certified by a
strict continuous witness, conjunction search can be rewritten as

\[
\boxed{
\text{frozen discrete invariant}
+
\text{interior homotopy}
+
\text{strict-separator stability}.
}
\]

The principle avoids re-searching the entire high-dimensional discrete
object space and instead reduces the joint construction to a continuous
deformation within an equivalence class.

\subsection{6.2 Algebraic Assumption Elimination by Null-Sector
Decoupling}\label{algebraic-assumption-elimination-by-null-sector-decoupling}

The second result arises from assumption auditing. A local process
compiler may introduce sector-preservation or paired-inactivity
conditions to exclude coherence-related counterexamples, but a condition
written explicitly in a theorem is not necessarily an independent axiom.
For the Choi/PSD block of a completely positive map, if

\[
X=
\begin{pmatrix}
A&B\\
B^\dagger&0
\end{pmatrix}
\succeq0,
\]

a zero diagonal block forces \(B=0\). Applied to the current
process/channel structure, if the population block from a vacuum input
sector to a real output sector vanishes exactly,

\[
Q_{\neg\Omega}\Phi_x(P_\Omega)Q_{\neg\Omega}=0,
\]

then the coherent cross blocks connecting the same zero-population
corner within that output sector already vanish as a consequence of
complete positivity. The additional sector-preservation condition
originally used to forbid those terms can therefore be removed.

The boundary of the argument is essential. The zero-corner reasoning
removes only the cross block in a \textbf{same-output coherent merge};
it does not justify erasing coherence transport that may still occur
between distinct output sectors. The Theory-Discovery Agent consequently
separates state into same-output merge and cross-output transport
objects, removing a redundant assumption without shrinking the original
allowed process class. The result converts a newly introduced assumption
into continuing \textbf{assumption debt}: only conditions that cannot be
derived from existing positivity, trace-preservation, and
reference-extension constraints are retained as independent theoretical
premises.

\subsection{6.3 Global Acted-Set Normal
Form}\label{global-acted-set-normal-form}

Local analysis of weak closed-lab protocols is susceptible to case
explosion: dynamic inactivity, physical timing, loss, vacuum sectors,
and routed corridors can each require separate treatment. The
Theory-Discovery Agent identifies a common monotone history variable
shared by these patches: the set of parties that have actually acted,

\[
K_n\subseteq\{1,\ldots,N\},
\qquad
K_{n+1}=K_n\cup\{k_n\},
\quad k_n\notin K_n.
\]

Physical time, loss history, and route branch can then be absorbed into
internal coherent control and buffering, while the exposed state records
only which free party slots have not yet been used. The construction
aligns with the generic routed graph of Routing Quantum Control of
Causal Order (Grothus et al. 2025): the source internal sequence and
release buffers are encoded into intermediate maps, coherent
acted-set/history information enters routed control, each free local
operation occupies a single unsectorized external slot, and branches
that never become active terminate through a NULL completion.

A critical constraint is that the same unknown operation across distinct
coherent order sectors must remain \textbf{one black-box slot}, not be
duplicated into multiple independent queries by branch. The
corresponding target slot can be written

\[
J_{A_k}^{\mathrm{out}}
\circ
(\widehat M_{k,x}\otimes I_{C_k\alpha})
\circ
J_{A_k}^{\mathrm{in}},
\]

where route and control degrees of freedom bypass the operation,
preserving off-diagonal coherence among vacuum-real sectors, time bins,
and distinct acted-order sectors. The patch tree that would otherwise
grow with the number of cases is compressed to

\[
\boxed{
\text{physical history}
\longrightarrow
\text{monotone acted-set}
+
\text{internal coherent buffer}
+
\text{single free slot per party}.
}
\]

The normal form separates \emph{when an event occurs} from \emph{which
operation the experimenter is free to replace}, making physical timing
an internal implementation state rather than a change in external
operational type.

\subsection{6.4 Behavioural Equivalence Is Strictly Weaker than
Interface
Equivalence}\label{behavioural-equivalence-is-strictly-weaker-than-interface-equivalence}

The global normal form exposes a deeper equivalence problem: identical
behaviour in closed experiments does not automatically imply that two
descriptions preserve the same free-operation interface. Higher-order
quantum information formalizes physical transformations in which an
operation acts on another operation through quantum supermaps and
quantum combs (Chiribella, D'Ariano, and Perinotti 2008, 2009).
Consequently, if two descriptions genuinely represent the same black-box
interface, equality of the closed channel should be supplemented by one
uniform deterministic higher-order transformation that can substitute
the free operation while that operation remains unknown.

We therefore distinguish closed behavioural equivalence \(E_0\) from
interface-substitution equivalence \(E_1\). Consider the phase-unitary
family

\[
U_\theta=
\begin{pmatrix}
1&0\\
0&e^{i\theta}
\end{pmatrix}.
\]

A source protocol queries the unknown \(U_\theta\) serially \(r\) times,
so the closed operation is \(U_\theta^r\). At the \(E_0\) level, the
entire source behaviour can be associated operation-by-operation with
the target label \(U_\theta^r\), formally occupying one target slot. If,
however, the target interface genuinely provides only the unknown black
box \(U_\theta\), an exact higher-order implementation using \(q\)
queries has output entries whose Laurent/Fourier degree in
\(e^{i\theta}\) is at most \(q\), whereas the target \(U_\theta^r\)
contains frequency \(r\). Hence

\[
q\ge r,
\]

and \(r\) serial calls attain the bound, so

\[
\boxed{
Q(U_\theta\mapsto U_\theta^r)=r.
}
\]

The structure is consistent with the polynomial method in quantum query
complexity (Beals et al. 1998); related query-complexity extensions to
higher-order and indefinite-causal-order settings are studied by Abbott
et al. (Abbott, Mhalla, and Pocreau 2024). A strict separation follows:

\[
\boxed{E_1\subsetneq E_0.}
\]

Closed behavioural reproduction therefore cannot be promoted directly to
theory-preserving interface equivalence. When an external experimenter
remains free to replace an unknown operation, a conservative theoretical
reduction additionally requires a uniform interface transformation,
composition congruence, and an explicit resource budget.

\subsection{6.5 Operational Intervention Signature and
Anti-Retyping}\label{operational-intervention-signature-and-anti-retyping}

The interface-query separation raises a primitive-granularity problem.
If an arbitrary \(r\)-query composite \(U^r\) can be renamed as a new
unit-cost primitive, every query or resource lower bound loses
representation-independent meaning. The Theory-Discovery Agent therefore
reconstructs primitive structure from the independent late-bound
interventions actually provided by a fixed apparatus rather than
accepting textual primitive names directly.

For apparatus \(A\), define the operational intervention signature

\[
\boxed{
\Sigma_{\mathrm{op}}(A)
=
(H,\tau,\mathcal J,q,\mathcal R,\mathcal W),
}
\]

where \(H\) is the set of late-bound primitive ports; \(\tau(h)\) gives
the type of port \(h\); \(\mathcal J\subseteq2^H\) records joint
independent late-bindability; \(q(h)\) records the black-box query
budget; \(\mathcal R(S)\) records memory, side-communication,
preshared-resource, and postselection budgets; and \(\mathcal W(h)\)
records the spacetime or causal access window. Under the natural closure
property that \(S\in\mathcal J\) and \(T\subseteq S\) imply
\(T\in\mathcal J\), the structure \(\mathcal J\) can be represented as a
downward-closed capability hypergraph.

Suppose \(a,b\) are independently late-bindable source ports, while the
target rewrites them as a single composite primitive \(v=F(a,b)\). If
the target apparatus exposes no two independent subports and no fixed
zero-extra-resource adapter can accept arbitrary independent
\(M_a,M_b\), then the target has changed \(\mathcal J\); bundling is a
physical interface edit or an additional adapter resource. Conversely,
if the bundled primitive retains the same independent late-bound holes,
operational reconstruction recovers the original ports. Semantic
renaming alone cannot reduce query depth. This yields the
signature-relative anti-retyping principle

\[
\boxed{
\text{semantic retyping cannot reduce }D_{\mathrm{FI}}^{\Sigma_{\mathrm{op}}}
\text{ unless physical late-bindability changes.}
}
\]

The construction is compatible with the supermap, comb, and
routed-circuit literature, where operation slots and compositional
resources are represented explicitly (Chiribella, D'Ariano, and
Perinotti 2008, 2009; Grothus et al. 2025). Its role is not to redefine
higher-order maps, but to provide a stricter constitution-level
invariant for theory comparison: when two theories are claimed to differ
only by representation, the comparison must also preserve independent
intervention ports, joint replaceability, query budgets, memory and
communication resources, and causal access windows.

\subsection{6.6 Separation of Theoretical
Levels}\label{separation-of-theoretical-levels}

The five groups of results form a progressively stronger structural
chain. Full-rank conjunction interiorization addresses whether one
object can jointly satisfy multiple properties. Null-sector decoupling
addresses which explicit assumptions are genuinely independent. The
acted-set normal form addresses whether multiple local case distinctions
arise from a simpler global state variable. Behavioural-interface
separation addresses which form of equivalence actually preserves
free-operation capability. The operational signature finally specifies
which structures can serve as representation-independent primitive or
resource invariants.

These levels are not interchangeable. A behaviourally equivalent
representation does not automatically imply interface equivalence, and
an exact query-depth lower bound cannot automatically become a
constitution invariant before the primitive signature is fixed. We
therefore distinguish the current theoretical levels as

\[
\begin{aligned}
\text{representation result}
&<\text{behavioural equivalence}<\text{interface equivalence}\\
&<\text{composition congruence}<\text{operational-constitution claim}.
\end{aligned}
\]

The full-rank QSOST gluing result has already formed a concrete
open-problem solution candidate. The acted-set normal form, equivalence
hierarchy, and operational signature primarily change the objects and
acceptance criteria used in subsequent theory evaluation; at present
they are structural and meta-theoretical advances rather than
established new laws of nature. Whereas AI Co-Scientist, SciAgents, and
The AI Scientist-v2 primarily organize hypothesis generation, scientific
graph reasoning, and experimental workflow automation (Gottweis et al.
2026; Ghafarollahi and Buehler 2024; Yamada et al. 2025), the principal
output of the Theory-Discovery Agent additionally lies in
\textbf{rewriting theoretical structure}: determining which assumptions
can be removed, which local constructions admit a common form, which
equivalence relations require refinement, and which operational
signatures can support stable resource claims.

\section{7 Conclusion}\label{conclusion}

We present \textbf{Eureka}, a task-conditioned Meta-Agent architecture
that dynamically compiles open long-horizon tasks into obligation
structures and generates, executes, and governs the evolution of
specialized Macro-Agents during task execution. The central premise of
Eureka is that differences among scientific tasks are expressed not only
in prompt content or knowledge domain, but also in persistent state,
operator families, verification semantics, parallel structure, and the
boundaries of long-term autonomy. Based on this observation, we develop
a unified theoretical framework spanning fixed-architecture regret,
planning invalidation, soundness of recursive decomposition, promotion
and evolution amortization, information-sufficient subtree interfaces,
concurrency serializability, and compositional verification correctness.
Experiments on long-horizon orchestration, context compilation,
incremental recomputation, concurrent merging, and self-evolution
exhibit system behaviour consistent with these theoretical predictions.

At the systems level, Eureka completes 170/170 recursive long-horizon
tasks and produces 3,948 acceptance certificates. Governed Evolution
achieves both the lowest median total cost and the highest success rate
among the evaluated strategies. Compiled active context reduces the
median model-input context from 9,490 to 4,005 without changing task
success rate. Dependency-local rebuilding avoids 65.38\% of repeated
computation. All 16,000 concurrent-execution tasks are consistent with a
valid serial execution. These results indicate that separating semantic
planning from programmable state management, verification, caching, and
scheduling can allocate a larger fraction of the computational budget to
scientific reasoning that cannot be replaced by deterministic runtime
mechanisms.

More importantly, the same Eureka Meta-Agent forms distinct specialized
cognitive systems under two different epistemic structures. In research
on the Riemann Hypothesis, the Math/Conjecture Agent identifies
operator-access and representation bottlenecks, constructs a
finite/local Chebyshev cone separator and finite-cluster interpolation,
and advances a whole-vector positivity certificate candidate for
Suzuki's localized Weil quadratic form to \(0<a\le69/200=0.345\),
approximately 99.55\% of the first-prime threshold. Boundary-swap
analysis in the first-prime regime further compresses the next theorem
obligation to a joint certificate that retains the remaining positive
multiplication together with the modal Schur structure. The current
result is not a proof of the Riemann Hypothesis and does not establish a
new record for the proportion of zeros on the critical line; an
independent proof replay, formal verification, and literature-level
novelty review are still required before a formal mathematical claim can
be made.

In the quantum-process and spacetime-theory setting, the
Theory-Discovery Agent produces five groups of structural results.
Full-rank conjunction interiorization yields an explicit candidate for
two-setting/two-outcome QSOST gluing; null-sector decoupling removes a
redundant assumption already implied by complete positivity; the global
acted-set normal form unifies multiple timing, vacuum, and routing cases
through a monotone first-use state; behavioural-interface separation
establishes that closed behavioural equivalence is strictly weaker than
an equivalence preserving a black-box interface; and the operational
intervention signature ties primitive and resource semantics to
apparatus-fixed independent late-bindability rather than to semantic
naming. Beyond the concrete QSOST conjunction candidate, the latter
results are principally structural and meta-theoretical advances whose
broader physical generality and theory-level implications require
continued application and independent validation.

The broader research question posed by Eureka is not how to hand-design
a new agent for every scientific domain, but whether a general system
can form an appropriate computational organization from the cognitive
structure revealed by the task itself. If this capability continues to
hold across a wider range of open problems, agent architecture need not
remain a fixed engineering choice made before scientific reasoning
begins; it can become an adaptive component of the scientific
problem-solving process itself.

\section*{References}\label{bibliography}
\addcontentsline{toc}{section}{References}

\phantomsection\label{refs}

\end{document}